\documentclass[10pt,twocolumn,letterpaper]{article}
\usepackage[pagenumbers]{cvpr}

\usepackage{graphicx}
\usepackage{booktabs}
\usepackage{array}
\usepackage{amsmath,amssymb,bm}
\usepackage{xcolor}
\usepackage{pifont}
\usepackage{tikz}
\providecommand{\eyebrow}[1]{{\fontsize{7.5}{9}\selectfont\scshape\color{black!68}#1}}
\providecommand{\figcap}[1]{{\fontsize{7}{8}\selectfont\itshape\color{black!72}#1}}

\definecolor{figbad}{RGB}{214,69,69}
\definecolor{figok}{RGB}{61,163,93}
\providecommand{\iconbad}{\begin{tikzpicture}[x=1in, y=1in, baseline=-0.035in]\draw[figbad, line width=0.6pt] (0,0) circle (0.05); \draw[figbad, line width=0.7pt] (-0.025,-0.025) -- (0.025,0.025) (-0.025,0.025) -- (0.025,-0.025);\end{tikzpicture}}
\providecommand{\iconok}{\begin{tikzpicture}[x=1in, y=1in, baseline=-0.035in]\draw[figok, line width=0.6pt] (0,0) circle (0.05); \draw[figok, line width=0.7pt] (-0.028,0.0) -- (-0.008,-0.024) -- (0.03,0.027);\end{tikzpicture}}

\definecolor{cvprblue}{rgb}{0.21,0.49,0.74}
\usepackage[pagebackref,breaklinks,colorlinks,allcolors=cvprblue]{hyperref}
\def\confName{CVPR}
\def\confYear{2027}
\title{SoftPAT3D: Physics-Augmented Text-to-3D Scene Generation with Deformables}
\author{Lei Shu \quad Alberta Longhini \quad Leonidas Guibas \quad Gordon Wetzstein \quad Jeannette Bohg\\Stanford University\\{\tt\small \{leishu, alberta, gordon.wetzstein, bohg\}@stanford.edu, guibas@cs.stanford.edu}}
\usepackage{caption}
\usepackage{placeins}
\newcommand{\ixEpisodes}{28}
\newcommand{\ixScenes}{23}

\newcommand{\ixMove}{23}
\newcommand{\ixReached}{23}
\newcommand{\ixGripErr}{5.7}

\newcommand{\ixSecPerStep}{1.3}

\newcommand{\heldoutN}{32}

\newcommand{\matCases}{26}
\newcommand{\matScenes}{11}

\newcommand{\matSec}{8}
\newcommand{\matOursMedPct}{3}
\newcommand{\matSearchMedPct}{16}

\newcommand{\bfScenes}{32}
\newcommand{\bfDiv}{0}
\newcommand{\bfCross}{66.2}
\newcommand{\bfPen}{2.5}
\newcommand{\bfContact}{25.2}
\newcommand{\bfFloat}{13/40}
\newcommand{\bfUnsup}{0/40}
\newcommand{\bfBodies}{40}
\newcommand{\bfPresent}{32/32}

\newcommand{\ctrlFullTime}{2.4\,min}

\newcommand{\gapMean}{2.7}

\newcommand{\studyRigidOurs}{10}
\newcommand{\studyRigidN}{13}

\newcommand{\gapHoMean}{3.0}

\newcommand{\ablDevN}{33}

\newcommand{\ablDevAComplete}{18}

\newcommand{\ablDevDSettled}{33}

\newcommand{\ablDevDComplete}{21}

\newcommand{\ablHoDSettled}{24}
\newcommand{\ablHoDGates}{23}
\newcommand{\ablHoDPrompt}{14}
\newcommand{\ablHoDComplete}{14}
\newcommand{\ablDevSheetN}{26}
\newcommand{\ablDevVolN}{7}

\newcommand{\ablDevDSheetComplete}{16}

\newcommand{\ablHoVolN}{2}

\newcommand{\ablHoDSheetComplete}{14}

\newcommand{\eEEAstraComplete}{7}

\newcommand{\eEEAstraGeom}{15}

\newcommand{\eEEBlComplete}{0}
\newcommand{\eEEBlFComplete}{8}

\newcommand{\eEEBlFGeom}{19}

\newcommand{\eEEBlGeom}{11}

\newcommand{\eEEBlQComplete}{7}

\newcommand{\eEEHoloComplete}{2}

\newcommand{\eEEHoloGeom}{2}

\newcommand{\eEEOursComplete}{21}

\newcommand{\eEEOursGeom}{30}

\newcommand{\eEERigidComplete}{0}

\newcommand{\eEERigidGeom}{15}

\newcommand{\vlmN}{60}

\newcommand{\vlmMajAgree}{49}

\newcommand{\vlmMiniAgree}{44}

\newcommand{\budN}{12}

\newcommand{\budGrid}{9}

\newcommand{\idN}{60}
\newcommand{\idCat}{48}
\newcommand{\idApp}{34}

\newcommand{\idExtMed}{0.80}
\newcommand{\idExtQa}{0.57}
\newcommand{\idExtQb}{1.10}
\newcommand{\idBodies}{74}

\newcommand{\hcRaters}{5}
\newcommand{\hcN}{60}

\newcommand{\hcUnanYes}{30}

\newcommand{\hcInter}{92}
\newcommand{\hcMeanAgree}{94}
\newcommand{\hcMeanAgreeN}{56.4}

\newcommand{\pairRaters}{5}
\newcommand{\pairN}{10}
\newcommand{\pairPhysOurs}{10}
\newcommand{\pairPromptOurs}{10}
\newcommand{\pairVotes}{50}
\newcommand{\pairPhysVotes}{50}
\newcommand{\pairPromptVotes}{46}

\newcommand{\pairInterPhys}{100}
\newcommand{\pairInterPrompt}{88}
\newcommand{\pairRigid}{4}
\newcommand{\pairAstra}{3}
\newcommand{\pairHolo}{3}

\newcommand{\hoCorrectable}{7}
\newcommand{\hoCorrected}{5}
\newcommand{\hoClosed}{19}

\newcommand{\devClosed}{25}
\newcommand{\hoClosedGrid}{19}

\newcommand{\relcorrPos}{12}

\newcommand{\tsAllN}{65}

\newcommand{\tsAllGates}{57}

\newcommand{\tsHoN}{32}

\newcommand{\tsHoYes}{17}

\newcommand{\tsHoImgYes}{15}

\newcommand{\repOneDevSettled}{33}

\newcommand{\repTwoDevSettled}{33}

\newcommand{\assetDevSettled}{22}
\newcommand{\assetDevGates}{19}

\newcommand{\assetDevComplete}{14}
\newcommand{\assetDevSheetComplete}{9}

\newcommand{\assetHoSettled}{16}
\newcommand{\assetHoGates}{13}

\newcommand{\assetHoComplete}{7}
\newcommand{\assetHoSheetComplete}{7}

\newcommand{\repDevList}{21, 21, 20}

\newcommand{\repHoList}{14, 14, 13}

\newcommand{\bfqScenes}{30}

\newcommand{\bfqDiv}{0}
\newcommand{\bfqCross}{24.8}
\newcommand{\bfqPen}{2.2}
\newcommand{\bfqContact}{16.3}
\newcommand{\bfqFloat}{13/38}

\newcommand{\plausrigid}{43}
\newcommand{\plausrigidN}{13}
\newcommand{\cliprigid}{23.9}
\newcommand{\vqarigid}{0.50}

\newcommand{\plausours}{90}

\newcommand{\clipours}{28.2}
\newcommand{\vqaours}{0.74}

\newcommand{\plausastra}{81}
\newcommand{\plausastraN}{11}
\newcommand{\clipastra}{24.9}
\newcommand{\vqaastra}{0.47}

\newcommand{\plausoursastra}{87}

\newcommand{\clipoursastra}{27.7}
\newcommand{\vqaoursastra}{0.65}

\newcommand{\hobN}{25}
\newcommand{\hobBodies}{27}
\newcommand{\hobBlCross}{24.4}
\newcommand{\hobBlFloat}{9}
\newcommand{\hobBlContact}{17}
\newcommand{\hobOursContact}{53}

\newcommand{\hdAbsent}{21}
\newcommand{\hdNoMesh}{3}
\newcommand{\hdPlaced}{8}
\newcommand{\hdNamed}{2}
\newcommand{\astraNone}{3}
\newcommand{\astraScored}{29}

\newcommand{\foldN}{9}
\newcommand{\foldRetainedMed}{100}
\newcommand{\foldHeld}{7}
\newcommand{\foldOpened}{2}


\begin{document}
\twocolumn[{%
\renewcommand\twocolumn[1][]{#1}%
\maketitle
\begin{center}
  \captionsetup{type=figure}
\setlength{\fboxsep}{0pt}%
\centering
\providecommand{\tzH}{2.0in}
\providecommand{\tzO}{2.667in}
\providecommand{\tzR}{1.66in}
\providecommand{\tzP}{1.88in}
\providecommand{\tzG}{0.16in}%
\providecommand{\tzS}{0.08in}%
\setlength{\tabcolsep}{0pt}%
\begin{tabular}{@{}p{\tzP}@{\hspace{\tzG}}p{\tzR}@{\hspace{\tzG}}p{\tzO}@{}}
\hspace{0.14in}\parbox[b][\tzH][c]{1.72in}{\raggedright
  {\fontsize{38}{38}\selectfont\color{black!22}``\par}\vspace{-13pt}
  {\fontsize{12.5}{17}\selectfont\itshape several clothes thrown messily over the back and seat of a fabric sofa\par}}
& \includegraphics[height=\tzH]{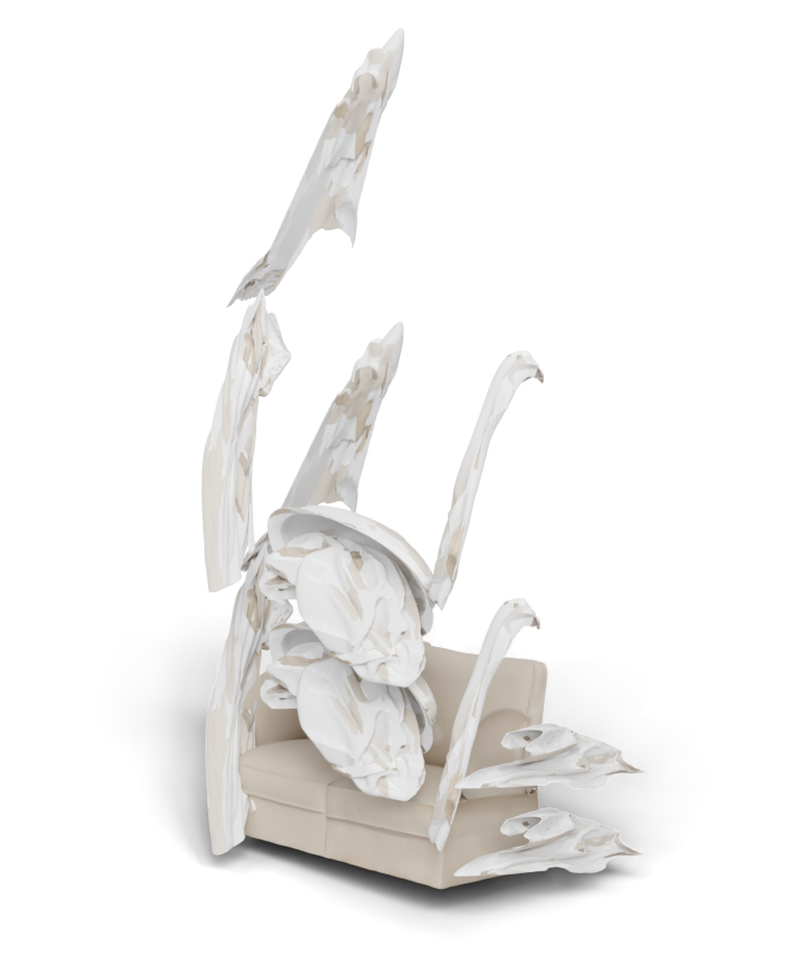}
& \includegraphics[height=\tzH]{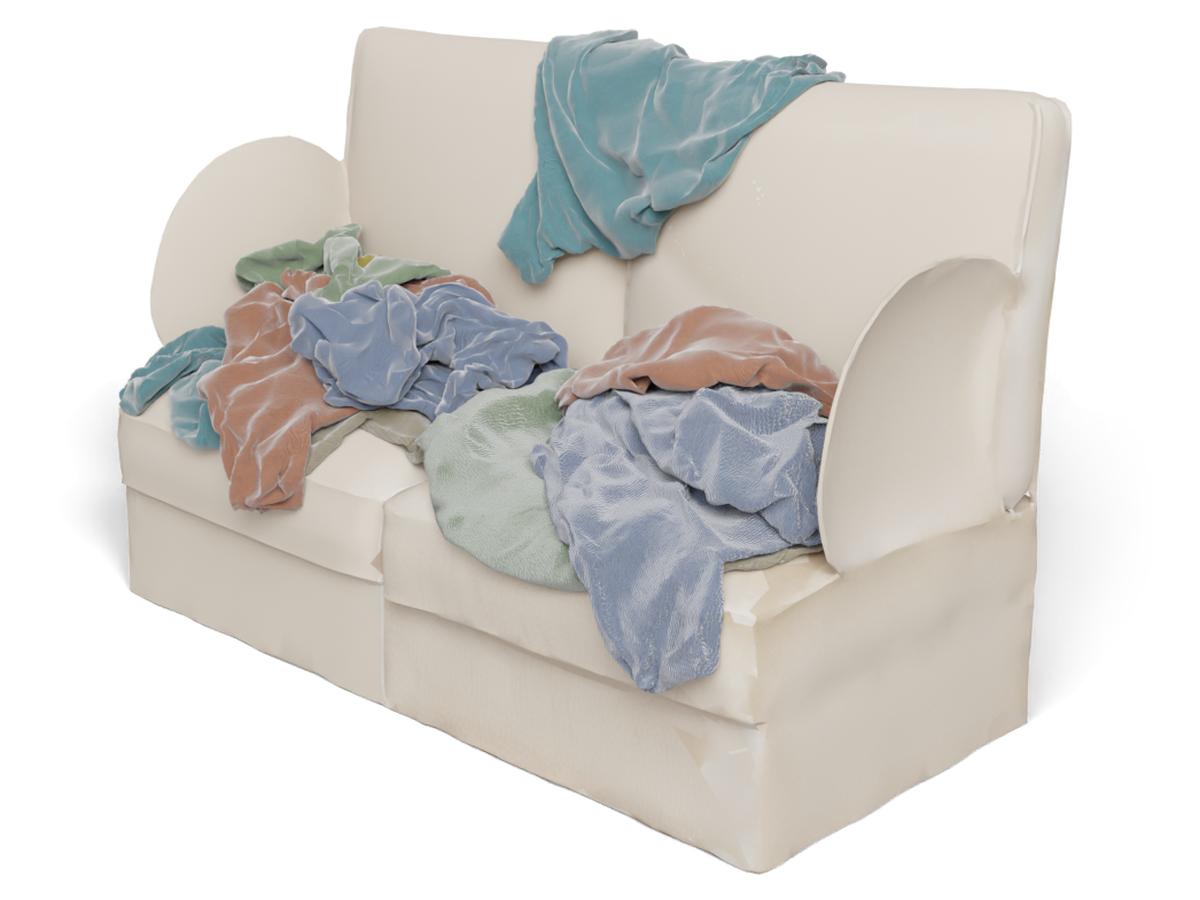} \\[3pt]
\multicolumn{1}{@{}l}{\hspace{0.14in}\parbox{1.72in}{\centering\eyebrow{text prompt}}} & \multicolumn{1}{@{\hspace{\tzG}}c}{\eyebrow{\iconbad\, rigid pipeline (previous work)}} & \multicolumn{1}{@{\hspace{\tzG}}c@{}}{\eyebrow{\iconok\, SoftPAT3D (ours), equilibrated}}
\end{tabular}\par\vspace{5pt}
\providecommand{\tzW}{1.021in}
\begin{tabular}{@{}c@{\hspace{\tzS}}c@{\hspace{\tzS}}c@{\hspace{\tzS}}c@{\hspace{\tzS}}c@{\hspace{\tzS}}c@{}}
\includegraphics[width=\tzW]{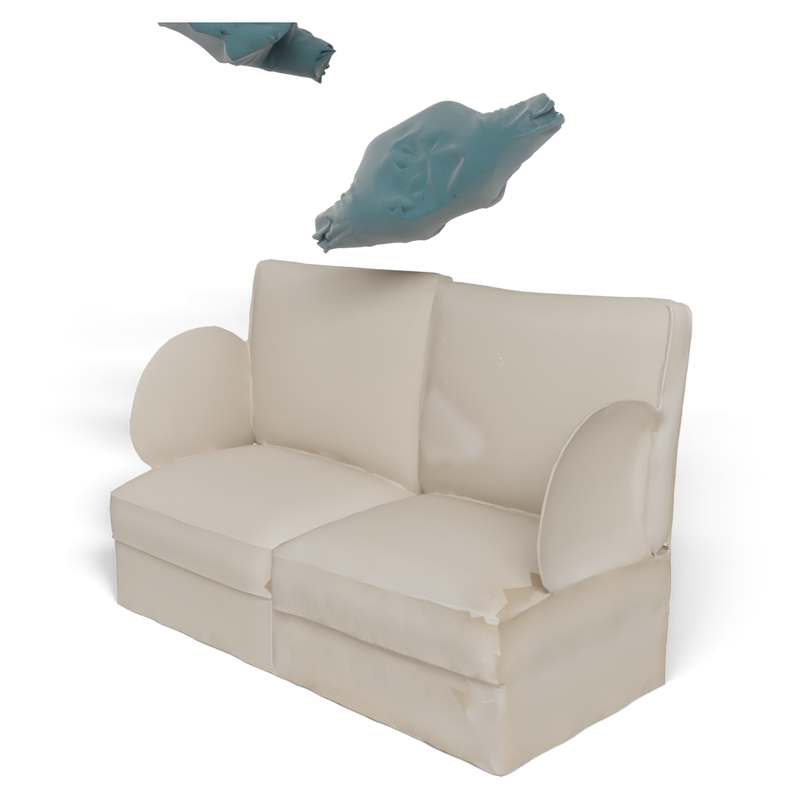}&
\includegraphics[width=\tzW]{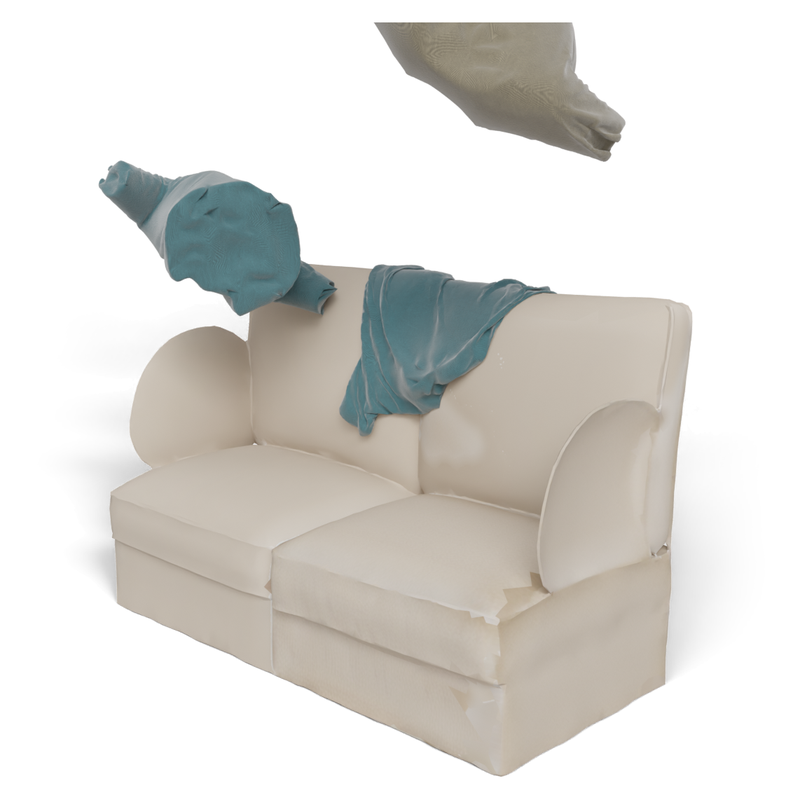}&
\includegraphics[width=\tzW]{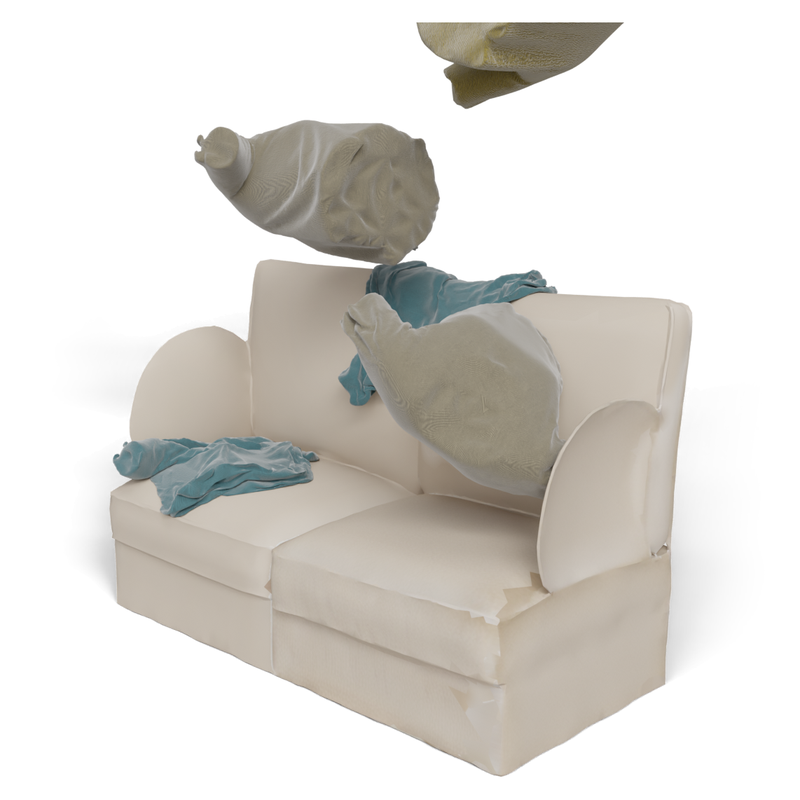}&
\includegraphics[width=\tzW]{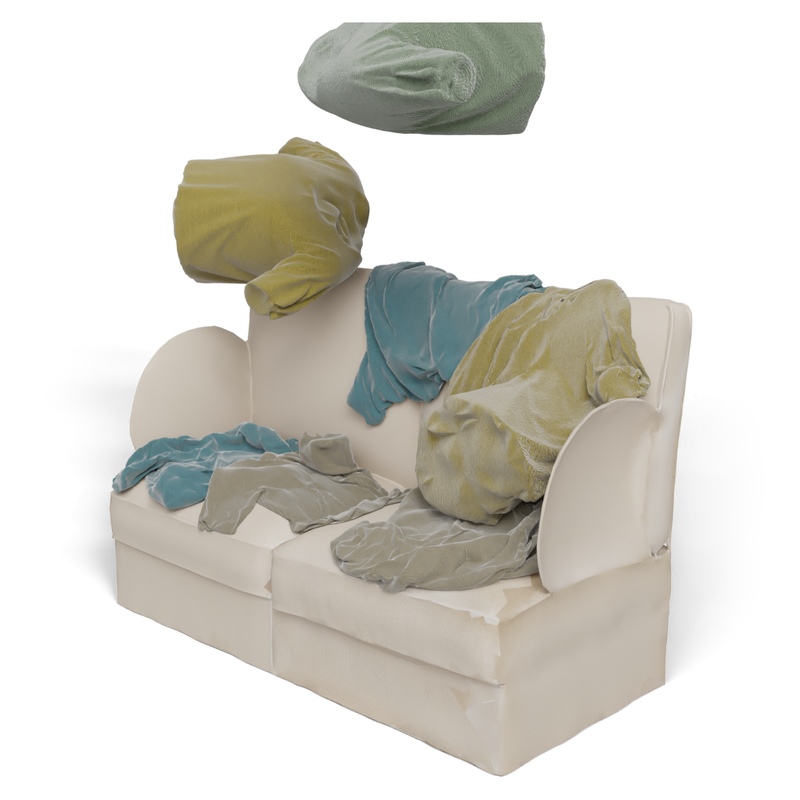}&
\includegraphics[width=\tzW]{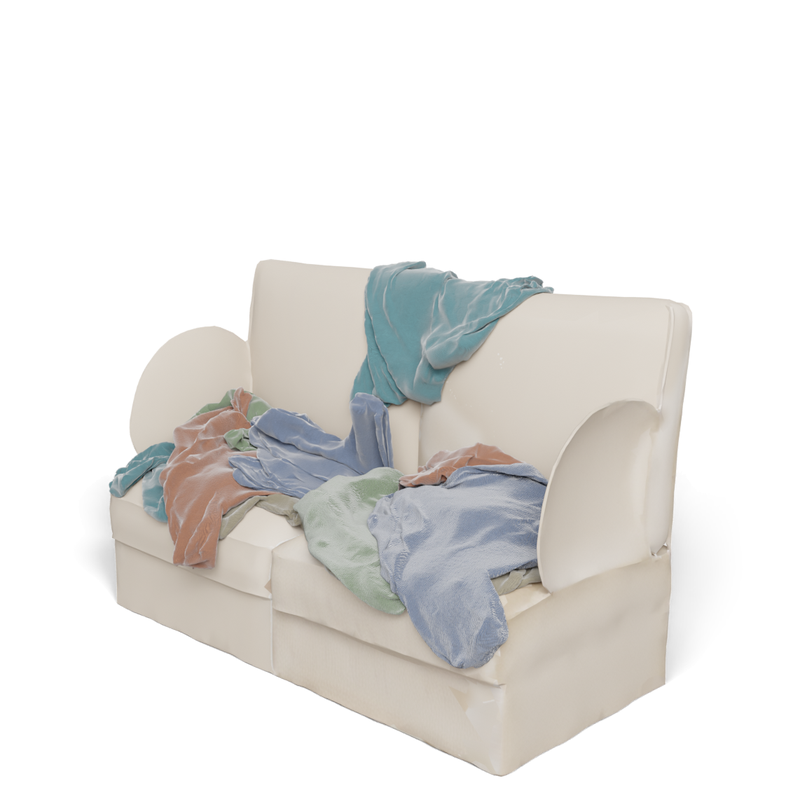}&
\includegraphics[width=\tzW]{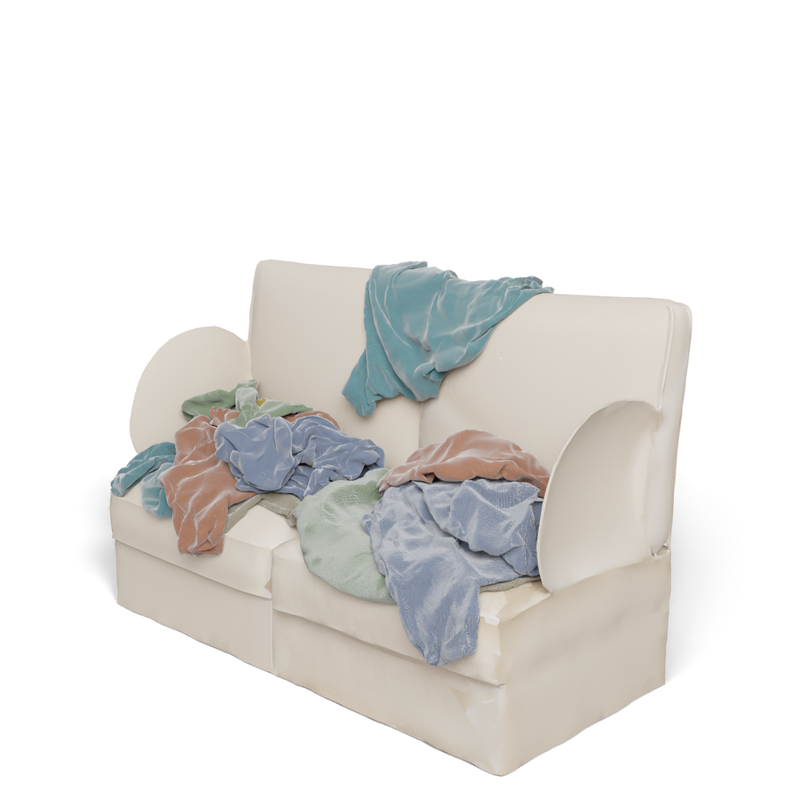}\\[2pt]
\eyebrow{release} & \eyebrow{first contact} & \eyebrow{landing} & \eyebrow{piling} & \eyebrow{relaxing} & \eyebrow{equilibrium}
\end{tabular}

  \captionsetup{width=0.9\textwidth}
  \captionof{figure}{\textbf{SoftPAT3D generates simulation-ready scenes with deformable objects from text.} Given a prompt, the pipeline generates the furniture and the deformable objects, constructs the inputs that a deformable simulation requires, a rest shape, a material and an intersection-free initial state, and drives the complete scene to static equilibrium under one contact model, verified by geometric checks and a vision--language model. Left: prior work simulates the generated clothes as rigid bodies, which float above or intersect the sofa. Right: the SoftPAT3D result, in which every garment rests on the sofa and none intersects it. Bottom: the garments falling and coming to rest in one simulation; the result is simulation-ready, and the same solver edits and manipulates it.}
  \label{fig:teaser}
\end{center}
}]

\begin{abstract}
We present SoftPAT3D, a physics-augmented framework for text-to-3D scene
generation in which deformable objects, such as cloth, stuffed toys and
ropes, are generated as physically valid, simulation-ready bodies and
placed in intersection-free contact with the furniture that supports
them, with one another and with themselves. Existing physics-augmented
pipelines simulate rigid bodies only and cannot accommodate a generated
deformable object, which text-to-3D generation returns as a
geometrically invalid mesh, sometimes self-intersecting, that has no
rest shape, no material model and no admissible initial state. These are
the inputs a deformable simulation requires, and we construct each from
information already in the pipeline. The rest shape comes from how the
object is manufactured rather than from the generated physically
inconsistent deformed configuration, and a material library assigns each
deformable object its initial elastic parameters, which can then be
optimized. The occluded support is regenerated and converted into a
watertight collision mesh, each object is released from an optimizable
placement above the supporting part that the prompt specifies, and the
complete scene is driven to static equilibrium under one unified contact
model, with completion verified by geometric checks and a
vision--language model. When a scene fails the prompt, a grid search
over placements corrects it, and the adjoint of a differentiable
simulation of the same contact energy reaches the targets the grid does
not, places several bodies jointly and recovers and optimizes material
parameters. Multi-body scenes equilibrate in a single simulation, and
the pipeline also runs from the text prompt alone. On all sixty-five
prompts, SoftPAT3D produces no undesired interpenetration of any kind,
whereas every other system does, and its scenes directly support scene
editing and robotic manipulation.
\end{abstract}

\section{Introduction}
\label{sec:intro}

Generating 3D scenes from natural language has broad applications in robotics, embodied AI and content creation, and recent pipelines have progressed from retrieving and arranging assets~\cite{holodeck,atiss} to generating or reconstructing the objects and settling the layout with a physics engine~\cite{pat3d,scenesmith,simuscene}. This physics stage yields a scene in which objects rest on one another without interpenetration and which a simulator can edit.

However, every object in such a physics stage is rigid. The rooms that users describe contain deformable objects, a tablecloth, a throw over a sofa, pillows and a quilt on a bed, and on these the pipeline fails. Text-to-3D generation returns the deformable object already draped, as a closed, self-intersecting surface that is plausible from the generation viewpoint but has no rest shape, no material and no admissible initial state. Current systems either simulate this surface as a rigid body, which leaves it floating above or intersecting the furniture (Fig.~\ref{fig:teaser}), or demote it to a non-simulated covering~\cite{scenesmith}.

Simulating a generated deformable object in this physics stage raises three challenges. First, the solver requires a rest shape, the geometry at which the elastic energy vanishes; the generated surface provides none, and thickening it would bake the generator's folds into the rest shape. Second, the simulation requires an intersection-free initial state, yet the support beneath the object is occluded in the reference image and generated merged with its occluder, and the object must start above its support in the relation the prompt states. Third, a converged simulation is not necessarily a complete scene, since the object may slide to the floor or settle on a support generated as a solid block, so the pipeline needs a completion criterion.

To address these challenges, we propose SoftPAT3D, which constructs all three inputs while leaving the solver of the rigid pipeline unchanged. Our key observation is that the rest shape has a source that exists for every object class without any observation: the way the object is manufactured. Garments are sewn from parametric sewing patterns, linens are cut as single panels and pillows are stuffed from two seamed panels, under class-specific size priors, so the rest shape carries the construction of its class and none of the generator's folds. SoftPAT3D then prepares the initial state from the prompt and the layout: the deformable object is inpainted out of its support, which is regenerated and converted into a watertight collision mesh, and a language model reads the supporting part off the prompt, above which placement rules position the object. All bodies are simulated to static equilibrium under one contact barrier, and two geometric checks and a vision--language model decide whether the scene is complete. Finally, when the scene fails the prompt, a grid search over placements corrects it, initialized by the adjoint of a differentiable simulation of the same energy, which also optimizes several bodies jointly and differentiates the material and the rest shape.

The result is a text-to-3D pipeline whose physics stage simulates deformable objects; the concurrent CoDimRecon~\cite{codimrecon} reconstructs such a scene from a multi-view capture instead (Sec.~\ref{sec:related}). We validate SoftPAT3D on sixty-five prompts, thirty-two of them written after the pipeline was frozen, against four baselines. Across every thin-shell and volumetric-solid scene in the paper, no equilibrated object intersects its support, whereas Holodeck~\cite{holodeck} crosses at $71$ edges per thousand, a cloth solver on the generated asset at $196$, and the rigid pipeline leaves its cloth $15$\,mm inside the furniture; under one completion criterion applied to every system, SoftPAT3D completes \eEEOursComplete{} of 34 scenes against at most \eEEBlFComplete{} for any baseline. Multi-body scenes equilibrate in one simulation, the pipeline also runs from the prompt alone, and the same solver edits and manipulates the result. We also compare the placement gradient with a grid search at equal compute and characterize the prompts outside the placement vocabulary (Sec.~\ref{sec:limits}). In summary, our main contributions are:
\begin{itemize}\setlength{\itemsep}{1pt}
\item We construct rest shapes from manufacturing priors: garments sewn from parametric sewing patterns, linens cut as panels and pillows stuffed from seamed panels, under class-specific size priors, which give a generated deformable object the rest shape that its surface lacks.
\item We propose a physics-aware initialization that regenerates the occluded support into a watertight collision mesh and maps the support relation in the prompt to a placement by one placement principle, so that thin shells, volumetric solids and rigid supports reach static equilibrium under a single contact barrier.
\item We define a completion criterion of geometric checks and a VLM-based prompt check, and a placement optimization by grid search, compared at equal compute with the adjoint of a differentiable simulation, which reaches targets past an obstruction, extends to several bodies and recovers material parameters.
\end{itemize}

\section{Related Work}
\label{sec:related}

\paragraph{Text-to-3D scene generation.} Layout-based methods compose retrieved or generated assets~\cite{holodeck,atiss,physcene,scenethesis}. PAT3D~\cite{pat3d} adds a rigid physics stage with differentiable layout optimization, concurrent systems equilibrate or verify rigid scenes~\cite{simuscene,scenesmith,sage,physcene3d,stable}, and procedural~\cite{infinigen}, program-authoring~\cite{simworlds,viga} and single-image~\cite{telephysics,wonderplay} systems build scenes by other routes. None of these simulates a deformable object at the scene level (Table~\ref{tab:scope}): SceneSmith~\cite{scenesmith} demotes tablecloths to non-simulated coverings, and systems that generate deformable assets place them by layout without scene-level simulation~\cite{embodiedgenv2,physxomni,embodiedsurvey}. Amodal completion recovers an object occluded in the scene image~\cite{pix2gestalt,reparo,amodal3r,gen3dsr,hiscene,scenemaker,scenecomplete} or its self-occluded parts~\cite{slice3d}; here it is the support, which we regenerate and decompose into convex cells~\cite{coacd}.

\begin{table}[t]
\centering\scriptsize
\setlength{\tabcolsep}{2pt}
\begin{tabular*}{\linewidth}{@{\extracolsep{\fill}}l cccccc@{}}
\toprule
 & prompt & rigid & soft sim. & soft equil. & soft grip & soft grad. \\
\midrule
Holodeck~\cite{holodeck} & \checkmark & -- & -- & -- & -- & -- \\
PAT3D~\cite{pat3d} & \checkmark & \checkmark & -- & -- & -- & -- \\
SimuScene~\cite{simuscene}$^{\ddagger}$ & -- & \checkmark & -- & -- & -- & -- \\
SceneSmith~\cite{scenesmith} & \checkmark & \checkmark & -- & -- & -- & -- \\
CoDimRecon~\cite{codimrecon}$^{\dagger\ddagger}$ & -- & \checkmark & \checkmark & \checkmark & \checkmark & -- \\
SoftPAT3D (ours) & \checkmark & \checkmark & \checkmark & \checkmark & \checkmark & \checkmark \\
\bottomrule
\end{tabular*}

\caption{\textbf{Scope.} Input (a text prompt; $^{\ddagger}$SimuScene takes a single image and CoDimRecon a multi-view capture), a rigid-body physics stage, and, for deformable objects, simulation, equilibrium on their supports, manipulation by a gripper and a gradient through contact. PAT3D's gradient and manipulation act on rigid bodies. $^{\dagger}$Concurrent work.}
\label{tab:scope}
\end{table}

\paragraph{Simulation-ready deformable assets and environments.} Physics-grounded dynamics animate a given image or captured shape~\cite{gid,physgen,physgaussian,physdreamer}, and simulation-ready asset generation~\cite{atlas3d,physx3d,sophy,physxomni} produces one object at a time; neither brings a scene to equilibrium under contact. Elsewhere, soft objects enter a scene per asset or from a library~\cite{infinigen,behavior1k,garmentlab,digitalcousins}, or as dynamic episodes~\cite{text2sim,robogen,cp4d}. Sewing patterns provide garment rest shapes~\cite{garmentcode,korosteleva2021,garmentcodedata,sewformer,dresscode,chatgarment,aipparel}.

\paragraph{Deformable simulation and differentiable physics.} Discrete membrane and bending models~\cite{grinspun} describe thin shells, incremental potential contact (IPC) keeps implicit time stepping intersection-free~\cite{ipc}, its codimensional extension covers shells and rods~\cite{cipc}, and vertex block descent minimizes the same potential by local block updates on the GPU~\cite{vbd}. Differentiable simulators propagate gradients through explicit~\cite{difftaichi} and implicit~\cite{diffcloth,diffvbd} steps and estimate materials from captures~\cite{diffavatar,bayescloth,phystwin,forceaware} or images~\cite{dress1to3}, after gradient-free search~\cite{bhat03}. CoDimRecon~\cite{codimrecon}, concurrent with this work, reconstructs a simulation-ready scene of rods, shells and solids from multi-view RGB under IPC contact; the capture supplies resting geometry, layout and scale, so each object starts near equilibrium and is settled, whereas a generative pipeline has no observation to fit. We thus address this gap by constructing the inputs that a capture would otherwise supply.

\section{Method}
\label{sec:method}

\begin{figure*}[t]
  \centering
  \includegraphics[width=\textwidth]{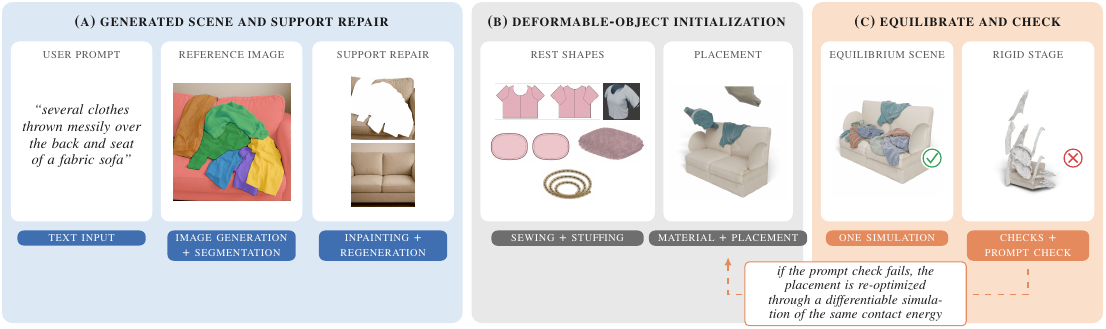}
  \caption{\textbf{Overview of SoftPAT3D.} Each card shows the state of one prompt at that point of the pipeline, and its label names the step that produced it. (a) The prompt is expanded into a reference image, segmented and lifted as in PAT3D, and the support hidden under each deformable object is inpainted, regenerated and converted into a collision mesh (Sec.~\ref{sec:init}). (b) Each deformable object receives a rest shape from its manufacturing process under a size prior (Sec.~\ref{sec:patterns}), a material from the material library and a placement above the part the prompt names (Sec.~\ref{sec:init}). (c) One simulation brings the scene to equilibrium under a single contact barrier (Sec.~\ref{sec:energies}), and the result passes the geometric checks and the prompt check (Sec.~\ref{sec:gates}); a scene that fails enters placement optimization (Sec.~\ref{sec:dvbd}).}
  \label{fig:pipeline}
\end{figure*}

Given a text prompt, SoftPAT3D runs the generation stages of PAT3D~\cite{pat3d} and then simulates the scene with its deformable objects to static equilibrium. Each stage constructs one input that the solver requires and generation does not supply (Fig.~\ref{fig:pipeline}): support repair (Sec.~\ref{sec:init}), deformable-object initialization (Secs.~\ref{sec:patterns} and~\ref{sec:energies}), equilibration with a completion check (Sec.~\ref{sec:gates}) and, when the prompt check fails, placement optimization (Sec.~\ref{sec:dvbd}).

\subsection{Preliminaries}
\label{sec:base}

PAT3D expands the prompt into a reference image, segments
it~\cite{sam}, lifts each crop to a textured mesh~\cite{hunyuan3d},
arranges the objects with a language model, converts them to
simulation meshes~\cite{ftetwild} and simulates them as rigid bodies to static equilibrium.
Following incremental potential contact~\cite{ipc}, each time step advances the scene state $\bm{x}$, the vector of all vertex positions, to the
minimizer of
\begin{equation}
E(\bm{x}) \;=\; \tfrac{1}{2\Delta t^{2}}\,
\lVert \bm{x}-\hat{\bm{x}} \rVert^{2}_{\mathbf{M}}
\;+\; \sum_{o}\Psi_{o}(\bm{x}_{o}) \;+\; B(\bm{x}),
\label{eq:ip}
\end{equation}
where $\hat{\bm{x}}=\bm{x}_n+\Delta t\,\bm{v}_n+\Delta
t^{2}\mathbf{M}^{-1}\bm{f}_{\mathrm{ext}}$ is the predicted position,
$\mathbf{M}$ the mass matrix, $\Psi_{o}$ the elastic energy of object
$o$ and $B$ the log-barrier contact potential,
\begin{equation}
B(\bm{x}) = \kappa \sum_{k} b\big(d_k(\bm{x})\big),\qquad
b(d) = -(d-\hat d)^{2}\ln\tfrac{d}{\hat d},
\label{eq:barrier}
\end{equation}
summed over every pair $k$ of surface primitives whose distance
$d_k$ is below the activation distance $\hat d$, and zero beyond it, with lagged
Coulomb friction. Each step is minimized by Newton's method with a continuous-collision line search, so a trajectory that starts intersection-free remains so~\cite{abd}. Rigid supports keep the
affine-body energy
$\Psi_{\mathrm{ABD}}(\mathbf{A})=\kappa_{a}V\lVert\mathbf{A}^{\!\top}\mathbf{A}-\mathbf{I}\rVert_F^{2}$
at $\kappa_a=100$\,MPa and are fixed in place. SoftPAT3D changes only the
energies $\Psi_o$ of the deformable objects and the state $\bm{x}$ at the
first step.

\subsection{Rest shapes}
\label{sec:patterns}

The elastic energy requires a rest shape, and the generated asset does
not provide one. A lifted towel arrives as a closed, already draped
surface whose two layers lie closer than $\hat d$, and thickening it would bake the generator's folds into the rest shape (Table~\ref{tab:control}, first row).

We therefore describe each deformable object by its manufacturing process, as
flat panels cut to an outline, sewn along seams and optionally
stuffed. Garments are sewn from parametric sewing
patterns~\cite{garmentcode,korosteleva2021}, relaxed once to equilibrium and
stored in a library of four patterns, T-shirt, trousers, skirt and dress, which every named garment in the paper draws on; plain linens are single panels discretized
on a simulation grid, a $128\times128$ mesh on
$[0,w]\times[0,\ell]$, cut to an outline where the object has one, such as a round cloth or an apron; pillows are two seamed panels inflated into a
superellipsoid $|x/a|^{p}+|y/b|^{p}+|z/c|^{p}=1$ whose interior
tetrahedralization serves as the stuffing. Since the scale of a monocular layout is ambiguous, every deformable object and support is clamped to the real-world size range of its class.

\subsection{Per-object energies}
\label{sec:energies}

Each object keeps or replaces its elastic energy according to its material
class. A keyword classifier on the object's name assigns it to \textsc{rigid}, \textsc{thin-shell}, \textsc{volumetric-soft} or \textsc{rod}, matching rigid compound names first. Cloth-class objects receive a Neo-Hookean
membrane on the rest shape of the panel and discrete bending on its dihedral
angles~\cite{grinspun},
\begin{equation}
\Psi_{\mathrm{shell}}(\bm{x}) = \sum_{t} \bar A_t\, h\, W(\mathbf{F}_t)
+ k_b \sum_{e} \frac{\lVert \bar{\bm{e}}\rVert}{\bar h_e}\,
(\theta_e-\bar\theta_e)^{2},
\label{eq:shell}
\end{equation}
where $\mathbf{F}_t\in\mathbb{R}^{3\times 2}$ is the deformation
gradient of triangle $t$ with rest area $\bar A_t$, $\theta_e$ the
dihedral angle at edge $e$ with rest value $\bar\theta_e$, rest
length $\lVert\bar{\bm{e}}\rVert$ and height $\bar h_e$, and the
membrane energy density is
\begin{equation}
W(\mathbf{F})=\tfrac{\mu}{2}\big(\operatorname{tr}\mathbf{F}^{\!\top}\mathbf{F}-2\big)
-\mu\ln J+\tfrac{\lambda}{2}\ln^{2}J,
\label{eq:membrane}
\end{equation}
with $J=\sqrt{\det\mathbf{F}^{\!\top}\mathbf{F}}$ and Lam\'e parameters $\mu=E/2(1+\nu)$ and
$\lambda=E\nu/(1+\nu)(1-2\nu)$. Pillow-class objects are
tetrahedralized and receive the rest-stabilized Neo-Hookean energy~\cite{snh}
\begin{equation}
W_{\mathrm{vol}}(\mathbf{F})=\tfrac{\mu}{2}\big(\operatorname{tr}\mathbf{F}^{\!\top}\mathbf{F}-3\big)
+\tfrac{\lambda}{2}\big(\det\mathbf{F}-1-\tfrac{\mu}{\lambda}\big)^{2}
\label{eq:snh}
\end{equation}
with foam parameters. Cables, cords and ropes are an exploratory extension, a line mesh with the stretching and bending model of the codimensional solver's rods~\cite{cipc} (supplementary material). A material library supplies the parameters of each class (Sec.~\ref{sec:setup}). All energies share the barrier of
Eq.~\eqref{eq:barrier}, so mixed scenes of thin shells, volumetric solids,
rods and rigid supports require no additional coupling.

\subsection{Physics-aware initialization}
\label{sec:init}

The barrier requires an intersection-free initial state and collision meshes with a well-defined interior, neither of which generated assets provide.

\paragraph{Support repair.} The support is partly occluded by the deformable
object in the reference image, and the generator fills the occluded
region with the occluder, merging a seat with the pile on it. We inpaint the deformable object out of the crop of the support before
lifting~\cite{pix2gestalt,gen3dsr,scenecomplete}, regenerate the
support from the completed image and align it to its slot in the layout.
Without this step, the generator fills the mask with cloth merged into the support in 32 of 33 scenes. The regenerated mesh is then converted into a watertight collision mesh~\cite{ftetwild}, since a single intersecting pair makes the solver reject the initial state. Contact is resolved against this closed surface, whose mean distance to the textured asset is \gapMean\,mm (development) and \gapHoMean\,mm (held-out).

\paragraph{Placement.} Objects visible in the reference image are
placed at their layout positions. For objects the prompt introduces beyond the image, a language model, given a geometric summary of the candidate surfaces, returns a part such as the seat, backrest, arm or rim and a drop height. The cloth is sized from the top surface of its support with a $0.1$\,m overhang, or to the size the language model estimates, and placed at the drop height above the part.
Every placement follows one principle: the object is placed above the part that the prompt names, at the lowest height at which its initial state is intersection-free, with its extent clamped to that part. The supplementary material lists the rules that apply this principle to the furniture classes of the scene set; a new support class needs no rule beyond the part vocabulary. Every placement is tested for intersection before the
first step, and an intersecting placement is raised,
$y_0\leftarrow y_0+\delta$, until $\min_k d_k(\bm{x}_0)>\hat d$.

\subsection{Completion checks and rendering}
\label{sec:gates}

A converged simulation is not necessarily a complete scene. We therefore apply two
geometric checks first. The containment check requires the centre of mass of the deformable object to project inside the ground projection $\mathcal{F}$ of its support. The support check rejects as a generation
failure a support that is nearly as solid as its bounding box or far
outside the size range of its class.
Scenes that pass both geometric checks are then checked against the prompt by a vision--language model, the prompt check, queried twice on one render, $J=[j_1=j_2=\mathrm{yes}]$; three further views and a second model agree with it on \vlmMajAgree{} and \vlmMiniAgree{} of \vlmN{} scenes, and \hcRaters{} human raters, who agree with one another on \hcInter\% of the renders, return its verdict on \hcMeanAgree\% of the scenes each on average (supplementary material). We call a scene \emph{complete} when
\begin{equation}
C \;=\; P\wedge S\wedge[X=0]\wedge G\wedge J,
\label{eq:complete}
\end{equation}
where $P$ holds when every requested deformable object is present, $S$ when the simulation converged
without the solver rejecting an intersecting state, $X$ is the number of deformable surface edges that intersect a collision mesh, and $G$ holds when both geometric checks pass. In the main run the prompt check is recorded but not acted upon (Sec.~\ref{sec:dvbd}).

\subsection{Placement optimization}
\label{sec:dvbd}

A scene that fails the prompt check has usually placed the object on a part of its support other than the one the prompt specifies. A language model maps the prompt to the intended part from a fixed vocabulary, or reports that no placement can satisfy it, and the part is resolved on the collision mesh to a target
$\bm{p}$, the area-weighted centroid of its faces,
$\bm{p}=\sum_{f\in\mathcal{P}}A_f\bm{c}_f/\sum_{f}A_f$. The placement
$\theta=(c_x,c_z)$ is then optimized so that the equilibrium state
$\bm{x}^{\star}(\theta)$, rather than the initial one, meets the
prompt,
\begin{equation}
\theta^{\star}=\arg\min_{\theta} L(\theta),\qquad
L(\theta)=\lVert \Pi(\bm{x}^{\star}(\theta))-\bm{p}\rVert^{2},
\label{eq:release}
\end{equation}
where $\Pi$ projects the centre of mass of the equilibrated cloth onto the
ground plane. We solve Eq.~\eqref{eq:release} with Adam using the
adjoint of a differentiable simulation of the same energy. The simulation consists of
$N$ implicit steps, each solved by Newton's method to stationarity
of Eq.~\eqref{eq:ip},
\begin{equation}
\bm{R}_n=\tfrac{1}{\Delta t^{2}}\mathbf{M}(\bm{x}_{n+1}-\hat{\bm{x}}_n)
+\nabla(\Psi+B+D)\big|_{\bm{x}_{n+1}}=\bm{0},
\label{eq:residual}
\end{equation}
where $\hat{\bm{x}}_n$ is the predicted position of Eq.~\eqref{eq:ip}, which carries the external force, $D$ is the lagged friction potential, and $\theta$ enters through
$\bm{x}_0$. For a loss on $\bm{x}_N$ the gradient is
$\mathrm{d}L/\mathrm{d}\theta{=}\sum_n\bm{w}_n^{\!\top}\partial\bm{R}_n/\partial\theta$,
with adjoint variables $\bm{w}_n$ obtained by a reverse sweep that solves one linear system per step with the Newton matrix $\partial\bm{R}_n/\partial\bm{x}_{n+1}$ of the converged step~\cite{diffipc}; a solver-level alternative replays the block sweep in reverse~\cite{diffvbd}. The adjoint is exact as long as the active contact set does
not change with $\theta$, where it matches finite differences. For a single body the loss has two parameters, which a grid search covers: the pipeline re-simulates the placements
\begin{equation}
\theta_j=\theta_0+f_j\,(\bm{p}_{xz}-\theta_0),\qquad
f_j\in\{\tfrac12,\tfrac34,1\},
\label{eq:grid}
\end{equation}
at the size the language model estimates and at the class size, subjects each candidate to the geometric checks and the prompt check, and keeps the first to pass. The adjoint serves where the search space grows: it optimizes the placements of several bodies jointly (Sec.~\ref{sec:gradient}) and, through the same sweep, the material parameters.

\begin{figure*}[t]
  \centering
\setlength{\tabcolsep}{2pt}%
\providecommand{\galh}{0.49in}
\providecommand{\galhb}{0.58in}
\providecommand{\gnmA}[1]{\parbox[t]{0.71in}{\centering\figcap{#1}}}
\providecommand{\gnmB}[1]{\parbox[t]{1.33in}{\centering\figcap{#1}}}
\noindent\begin{tabular}{@{}ccccccccc@{}}
\includegraphics[height=\galh]{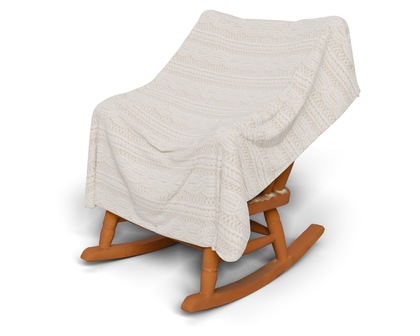} &
\includegraphics[height=\galh]{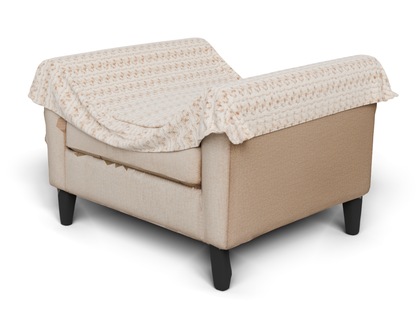} &
\includegraphics[height=\galh]{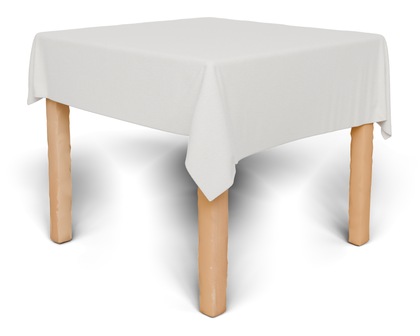} &
\includegraphics[height=\galh]{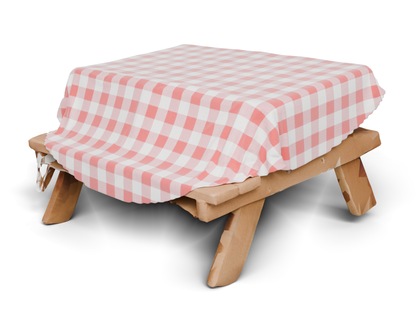} &
\includegraphics[height=\galh]{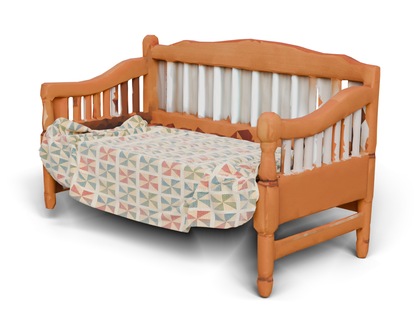} &
\includegraphics[height=\galh]{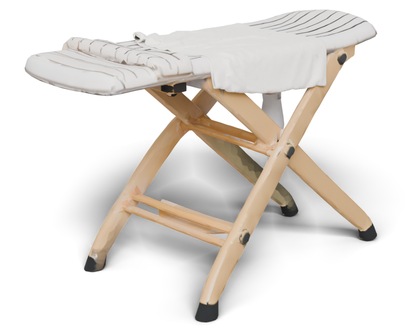} &
\includegraphics[height=\galh]{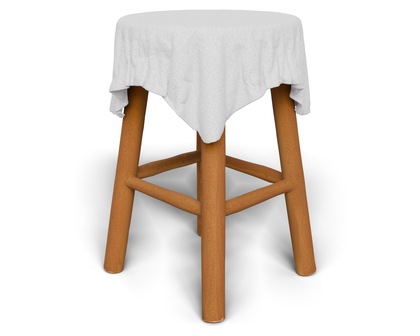} &
\includegraphics[height=\galh]{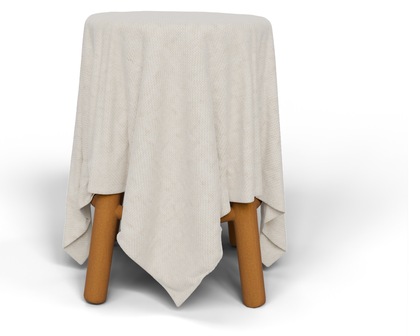} &
\includegraphics[height=\galh]{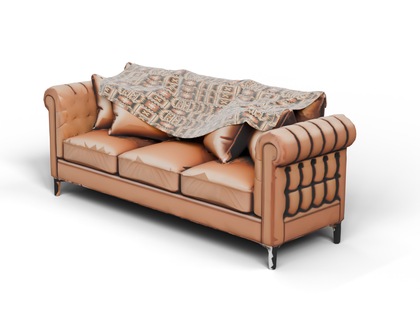} \\[0pt]
\gnmA{throw on a rocking chair} & \gnmA{blanket on an armchair} & \gnmA{tablecloth on a dining table} & \gnmA{cloth on a picnic table} & \gnmA{quilt on a daybed} & \gnmA{skirt on a chair} & \gnmA{cloth on a stool} & \gnmA{cloth on a round stool} & \gnmA{throw on a leather sofa}
\end{tabular}
\par\vspace{3pt}
\noindent\begin{tabular}{@{}cccccccc@{}}
\includegraphics[height=\galh]{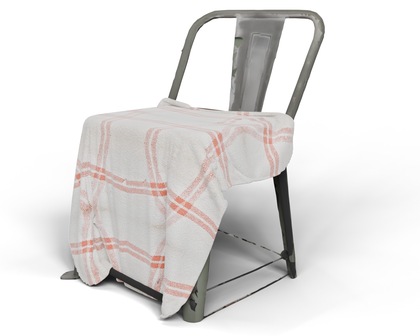} &
\includegraphics[height=\galh]{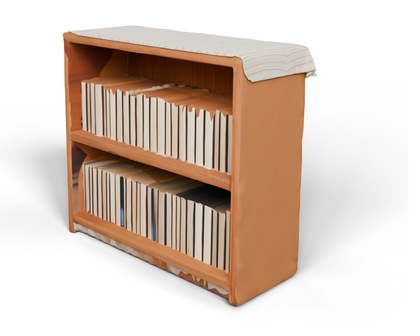} &
\includegraphics[height=\galh]{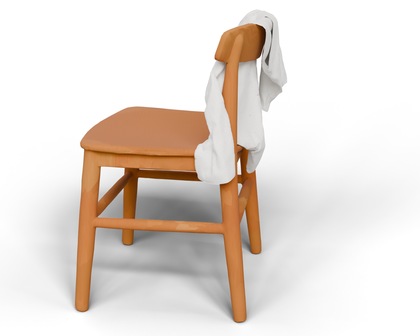} &
\includegraphics[height=\galh]{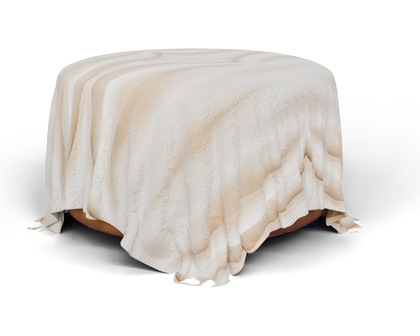} &
\includegraphics[height=\galh]{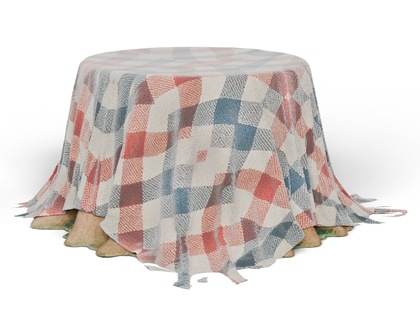} &
\includegraphics[height=\galh]{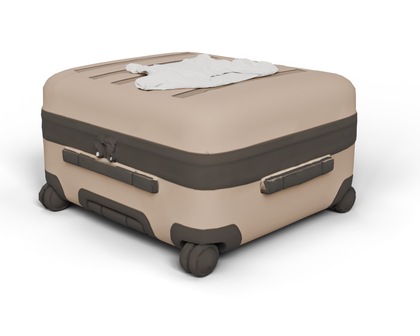} &
\includegraphics[height=\galh]{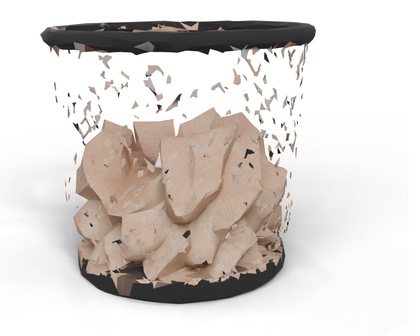} &
\includegraphics[height=\galh]{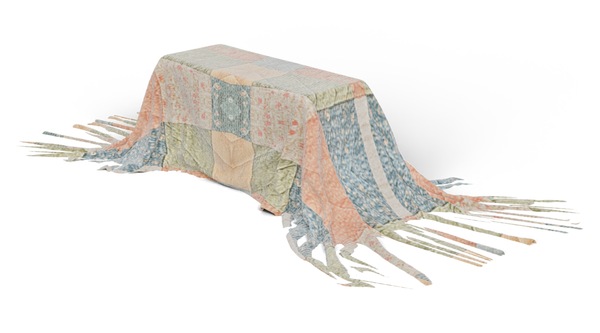} \\[0pt]
\gnmA{towel on a bar stool} & \gnmA{blanket on a bookshelf} & \gnmA{jacket on a desk chair} & \gnmA{blanket on an ottoman} & \gnmA{blanket on a tree stump} & \gnmA{sweater on a suitcase} & \gnmA{tissues in a trash bin} & \parbox[t]{0.95in}{\centering\figcap{quilt on a trunk}}
\end{tabular}
\par\vspace{3pt}
\noindent\begin{tabular}{@{}ccccc@{}}
\includegraphics[height=\galhb]{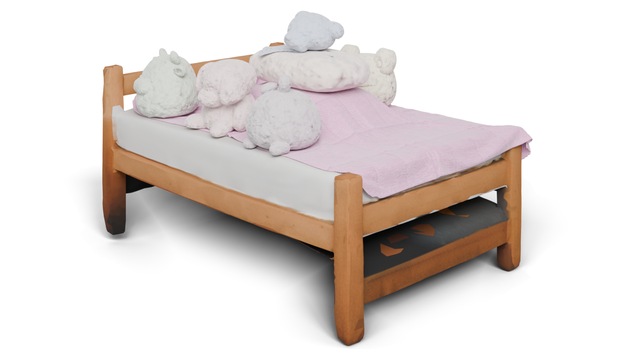} &
\includegraphics[height=\galhb]{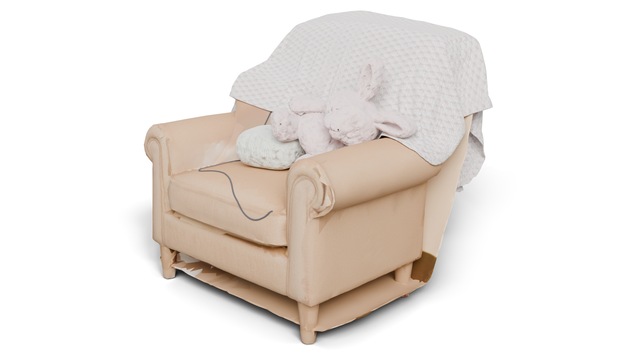} &
\includegraphics[height=\galhb]{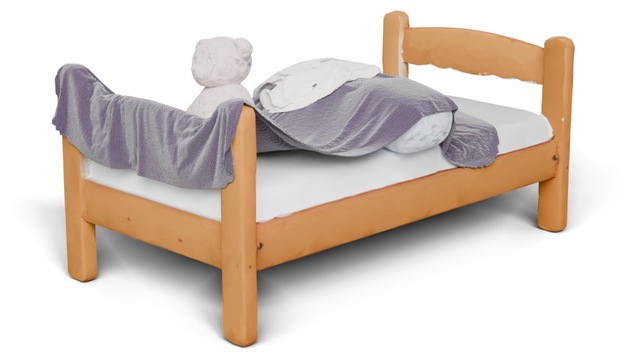} &
\includegraphics[height=\galhb]{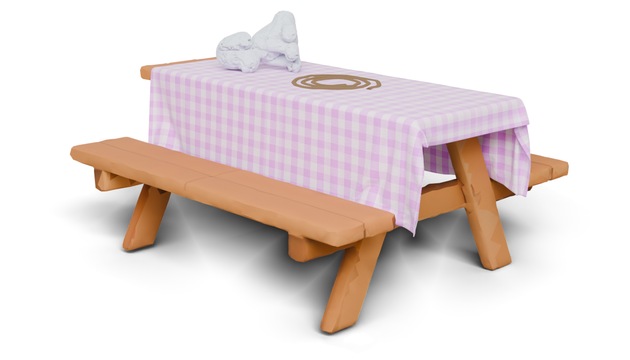} &
\includegraphics[height=\galhb]{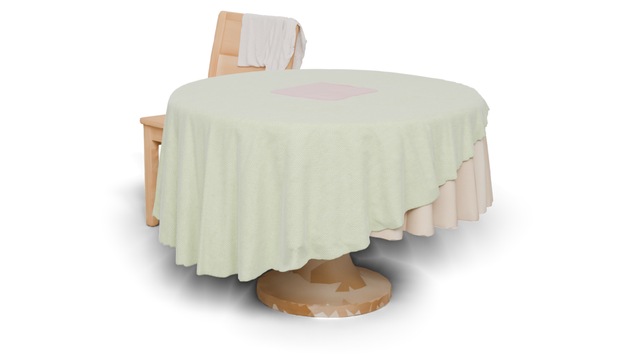} \\[0pt]
\gnmB{messy bed: quilt, pillows, five toys} & \gnmB{reading nook: blanket, pillows, rabbit, cable} & \gnmB{child's bed: quilt, hoodie, pillows, bear} & \gnmB{picnic: cloth, napkins, toy} & \gnmB{dining set: cloth, two napkins, sweater on a chair}
\end{tabular}

  \caption{\textbf{Generated scenes.} Top two rows: single-object scenes from the main run, development and held-out, each a generated support with a sewn cloth
  at equilibrium on the part the prompt specifies. Bottom: multi-body scenes
  with two to eight deformable objects, quilts, pillows, garments and
  stuffed toys, brought to equilibrium in one simulation. Every scene is rendered on the
  collision mesh of the solver.}
  \label{fig:gallery}
\end{figure*}

\section{Experimental Results}
\label{sec:exp}

\subsection{Experimental setup}
\label{sec:setup}

\paragraph{Scenes.} Since there is no standard benchmark for this task, we construct our own: thirty-two single-object scenes and three multi-body scenes generated end to end during development, and a second batch of thirty-two prompts written after the pipeline was frozen and run with one regeneration allowed and no other intervention. The main run is one run of the final code (scene set and prompts in the supplementary material).

\paragraph{Baselines.} The rigid pipeline runs the same prompt through the same stages and is measured after its own physics stage with the cloth asset kept rigid. Blender's cloth solver runs on the generated asset as
placed by the rigid pipeline (Blender) and on our inputs, the sewn
cloth at our placement over our collision meshes
(Blender$^{*}$), so that for cloth only the solver differs.
Holodeck~\cite{holodeck}, a retrieval system, and GPT-6 Astra, a frontier model prompted to write a Blender program, build their own scenes from the same prompts (supplementary material). The baselines of PAT3D, GraphDreamer, Blender-MCP and MIDI, generate no deformable object and are not rerun.

\begin{figure*}[t]
  \centering
\setlength{\tabcolsep}{1.5pt}%
\providecommand{\rowlab}[2]{\rotatebox{90}{\makebox[#1]{#2}}}
\providecommand{\cmpw}{0.87in}
\providecommand{\cmph}{0.58in}
\noindent\begin{tabular}{@{}c@{\hspace{3pt}}cccccc@{}}
 & \eyebrow{dining table} & \eyebrow{armchair} & \eyebrow{rocking chair} & \eyebrow{cloth, stool} & \eyebrow{daybed} & \eyebrow{picnic table} \\[1pt]
\rowlab{\cmph}{\iconbad\,\eyebrow{rigid}} &
\includegraphics[width=\cmpw]{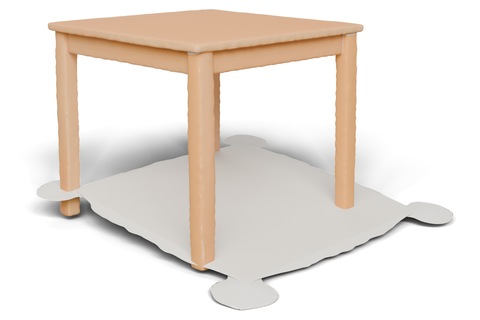} &
\includegraphics[width=\cmpw]{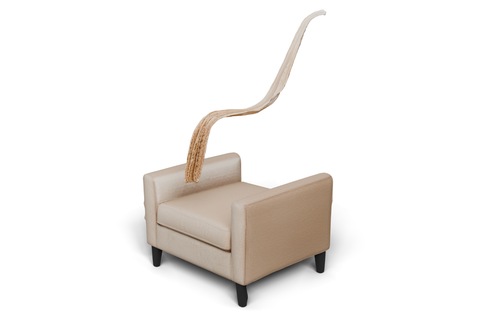} &
\includegraphics[width=\cmpw]{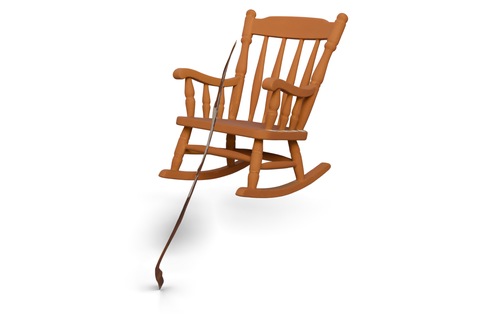} &
\includegraphics[width=\cmpw]{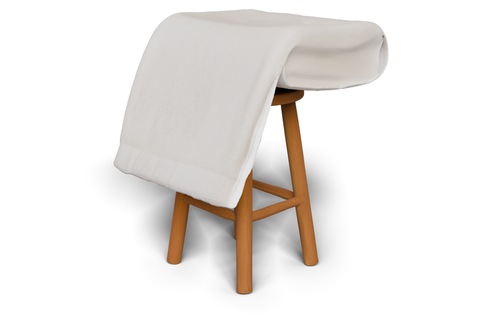} &
\includegraphics[width=\cmpw]{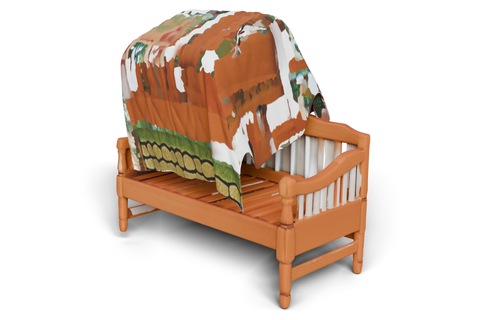} &
\includegraphics[width=\cmpw]{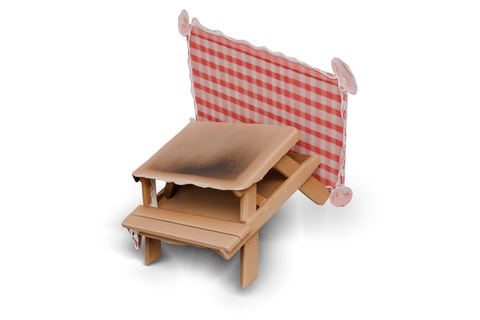} \\[1.5pt]
\rowlab{\cmph}{\iconok\,\eyebrow{ours}} &
\includegraphics[width=\cmpw]{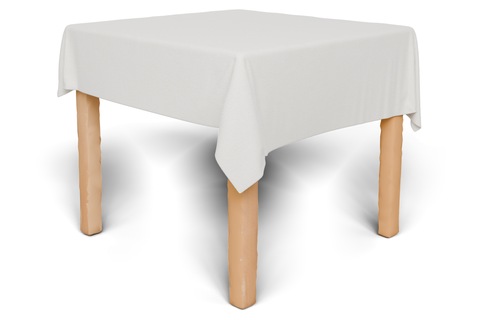} &
\includegraphics[width=\cmpw]{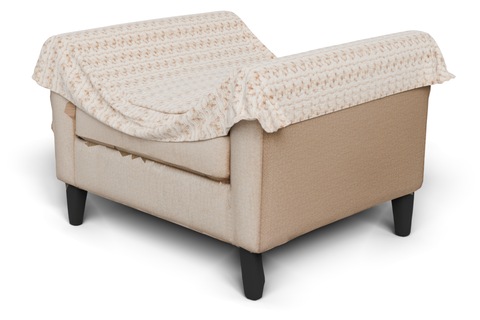} &
\includegraphics[width=\cmpw]{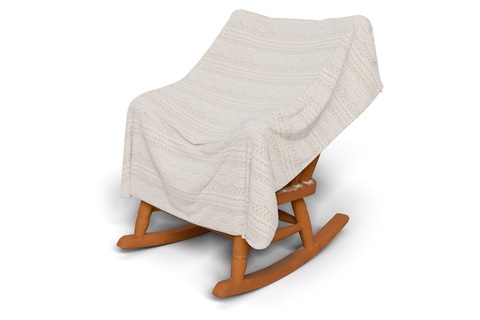} &
\includegraphics[width=\cmpw]{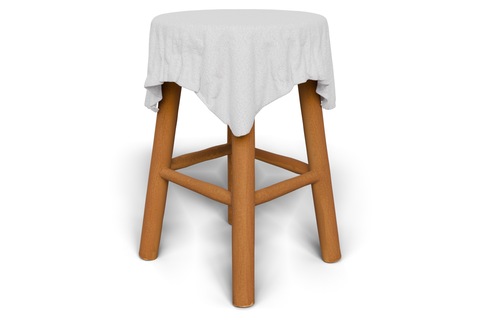} &
\includegraphics[width=\cmpw]{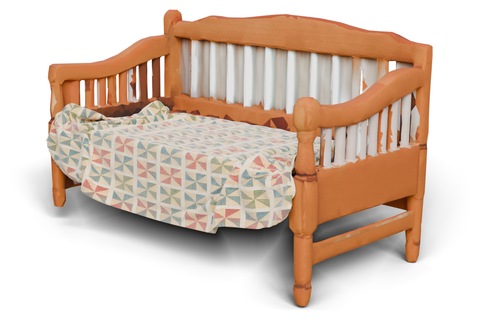} &
\includegraphics[width=\cmpw]{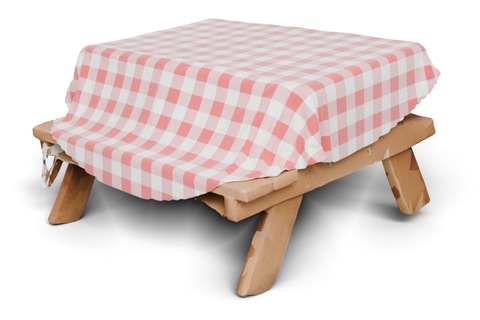}
\end{tabular}

  \caption{\textbf{Comparison with the rigid pipeline.} Top: the
  rigid pipeline places the generated deformable object as a rigid body that intersects or floats above the furniture. Bottom:
  the same scenes from the SoftPAT3D main run, as
  the solver returned them on their collision meshes.}
  \label{fig:compare}
\end{figure*}

\paragraph{Metrics.} All metrics are computed on the equilibrated
geometry of the solver against its collision meshes: crossings, the number of deformable surface edges that intersect another body, per thousand edges; penetration, the mean depth of the vertices inside a closed body; contact, the fraction of vertices within $5$\,mm of another body; the number of bodies that touch nothing or are unsupported; and the completion criterion of Eq.~\eqref{eq:complete}. Crossings are the edge analogue of the penetrating-triangle ratio of PAT3D.

\paragraph{Implementation details.} Thin shells are simulated at $\Delta t{=}0.02$\,s for up to $300$ implicit steps, stopping once $\max_i\lVert\bm{x}_i^{n}-\bm{x}_i^{n-40}\rVert<0.5$\,mm, with the cotton parameters of the material library for shells and its foam parameters for volumetric solids, effective rather than measured values (supplementary material). A sweep over time steps and mesh resolutions leaves every scene at static equilibrium with zero crossings (supplementary material).

\begin{table}[t]
\centering\scriptsize
\setlength{\tabcolsep}{2.5pt}
\begin{tabular*}{\linewidth}{@{\extracolsep{\fill}}l cccccc@{}}
\toprule
& Holodeck & Astra & rigid & Blender & Blender$^{*}$ & ours \\
\midrule
scenes & 32 & 29 & 34 & 34 & \bfScenes & 34 \\
deformable present & 8/32 & 29/32 & 34/34 & 34/34 & \bfPresent & 34/34 \\
soft bodies & 39 & 74 & 48 & 48 & \bfBodies & 43 \\
crossings (per 1k edges) & 71.2 & 40.2 & 5.1 & 196.2 & \bfCross & 0.0 \\
penetration (mm) & 0.0 & 1.7 & 14.8 & 1.9 & \bfPen & 0.0 \\
in contact (\%) & 10.3 & 38.5 & 1.8 & 22.8 & \bfContact & 39.9 \\
touching nothing & 22/39 & 1/74 & 14/48 & 4/48 & \bfFloat & 1/43 \\
unsupported & 23/39 & 5/74 & 15/48 & 0/48 & \bfUnsup & 0/43 \\
\midrule
geometric checks & \eEEHoloGeom & \eEEAstraGeom & \eEERigidGeom & \eEEBlGeom & \eEEBlFGeom & \eEEOursGeom \\
complete & \eEEHoloComplete & \eEEAstraComplete & \eEERigidComplete & \eEEBlComplete & \eEEBlFComplete & \eEEOursComplete \\
\bottomrule
\end{tabular*}

\caption{\textbf{Comparison with other systems.} SoftPAT3D has zero crossings and zero penetration, and its deformable objects contact their supports broadly and rest on them. Blender$^{*}$ has two scenes timed out. Deformable objects follow the labels of each system, and the
per-body rows pool the bodies each system placed. The last two rows apply our completion criterion to every system.}
\label{tab:compare}
\end{table}

\subsection{Comparison with other systems}
\label{sec:compare}

\begin{figure}[t]
  \centering
  \includegraphics{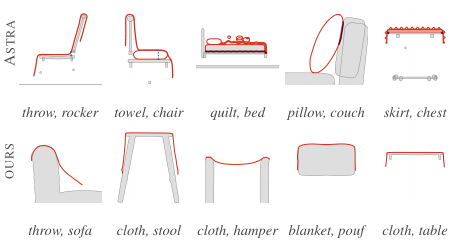}
  \caption{\textbf{Vertical sections,} cloth in red, furniture in grey, cloth inside the furniture in dark red, and a dashed line where the cloth floats above its support. Top: GPT-6 Astra; the blanket and the towel hang above their seats, the quilt runs through the mattress, the pillow sits inside the backrest, and the skirt ripples in mid-air. Bottom: ours; the cloth rests on the support, wraps its edges and hangs from them.}
  \label{fig:slices}
\end{figure}

Table~\ref{tab:compare} scores six conditions under one protocol over the thirty-two single-object scenes, the clothes sofa and the hamper. The rigid pipeline starts from the same generated assets and layout as ours, but simulating the cloth as rigid bodies leaves them intersecting the furniture at $5.1$ edges per thousand with $14.8$\,mm penetration, touching their supports over only $1.8\%$ of their vertices and floating in 14 of 48 bodies, as in Fig.~\ref{fig:compare}. Blender's cloth solver diverges in 24 of 34 scenes and crosses at 196 per thousand edges on the generated asset, and still crosses at \bfCross{} per thousand given our inputs. Holodeck places a deformable object in only 8 of 32 scenes, and Astra models fabric without a cloth solver. By contrast, SoftPAT3D has zero crossings and zero penetration, the largest contact area, $39.9\%$ of vertices, and no unsupported body; under Eq.~\eqref{eq:complete} applied to every system, with the prompt check on the render of each system, \eEEOursComplete{} of our 34 scenes are complete, against \eEEBlFComplete{} for Blender$^{*}$, \eEEAstraComplete{} for Astra, \eEEHoloComplete{} for Holodeck and \eEERigidComplete{} for the rigid pipeline. A render hides an intersection under the cloth, so these claims rest on the geometry; Fig.~\ref{fig:slices} shows a vertical section. Judged on renders under the protocol of PAT3D (Table~\ref{tab:render}), GPT-5.4 prefers our drape to that of the rigid pipeline in a side-swapped forced choice on \studyRigidOurs{} of \studyRigidN{} scenes, and its plausibility score rises from \plausrigid{} to \plausours{}. In a blind study, \pairRaters{} participants compare our render with a baseline's on \pairN{} prompts, systems unidentified and sides randomized, and prefer ours for physical validity and prompt satisfaction on every pair (supplementary material).

\begin{table}[t]
\centering\scriptsize
\setlength{\tabcolsep}{3pt}
\begin{tabular*}{\linewidth}{@{\extracolsep{\fill}}l r ccc@{}}
\toprule
 & scenes & plausibility & CLIP & VQAScore \\
\midrule
rigid pipeline & \plausrigidN{} & \plausrigid{} & \cliprigid{} & \vqarigid{} \\
ours, the same scenes & \plausrigidN{} & \textbf{\plausours{}} & \textbf{\clipours{}} & \textbf{\vqaours{}} \\
\midrule
GPT-6 Astra & \plausastraN{} & \plausastra{} & \clipastra{} & \vqaastra{} \\
ours, the same scenes & \plausastraN{} & \textbf{\plausoursastra{}} & \textbf{\clipoursastra{}} & \textbf{\vqaoursastra{}} \\
\bottomrule
\end{tabular*}

\caption{\textbf{Render metrics under the protocol of PAT3D.} GPT-5.4 physical plausibility (0 to 100), CLIP score and VQAScore on one render per scene, on the scenes each baseline rendered.}
\label{tab:render}
\end{table}

\subsection{End-to-end results}
\label{sec:settle}

\paragraph{Held-out prompts.} The thirty-two prompts written after the pipeline was frozen
are our primary measurement under Eq.~\eqref{eq:complete}. Every scene reaches static equilibrium with no crossing, five prompts lose an object upstream of the physics stage, and of the twenty-seven that reach it, \hoClosed{} are complete with the placement optimization of Sec.~\ref{sec:dvbd} in the loop (\hoClosed{} of \heldoutN{} overall). The physics stage alone completes \ablHoDComplete{}, and the grid of Eq.~\eqref{eq:grid} completes \hoCorrected{} of the \hoCorrectable{} scenes whose prompt failure names a reachable part in six re-simulations per scene (supplementary material). On the same held-out inputs Blender's cloth solver crosses at \hobBlCross{} per thousand edges and leaves \hobBlFloat{} of \hobBodies{} bodies unsupported, against zero and none for ours.

\paragraph{Development set.} Of the thirty-three development scenes, \ablDevDComplete{} are complete under the same criterion (Fig.~\ref{fig:gallery}), \devClosed{} with placement optimization, and none has a crossing; nine multi-body scenes with two to eight deformable objects (Fig.~\ref{fig:gallery}, bottom) reach equilibrium in one simulation and pass both checks.

\begin{figure*}[t]
  \centering
\setlength{\tabcolsep}{2pt}%
\providecommand{\stripw}{\dimexpr(\textwidth-0.27in-20pt)/6\relax}
\providecommand{\striph}{\dimexpr\stripw*9/16\relax}
\providecommand{\srow}[2]{\rotatebox{90}{\makebox[\striph]{\shortstack{\eyebrow{#1}\\\eyebrow{#2}}}}}
\providecommand{\tlab}[1]{{\scriptsize\color{black!70}#1}}
\noindent\begin{tabular}{@{}c@{\hspace{3pt}}cccccc@{}}
\srow{tidy: toys}{off bed} &
\includegraphics[width=\stripw]{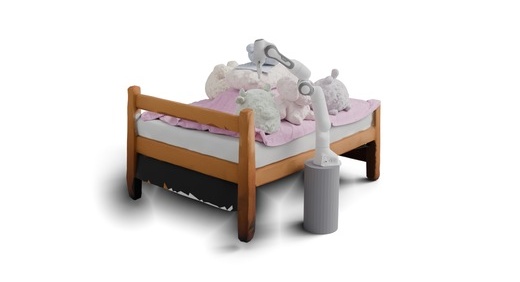} &
\includegraphics[width=\stripw]{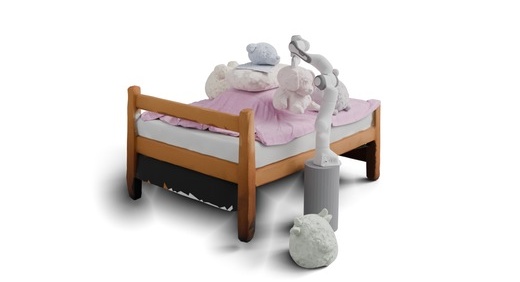} &
\includegraphics[width=\stripw]{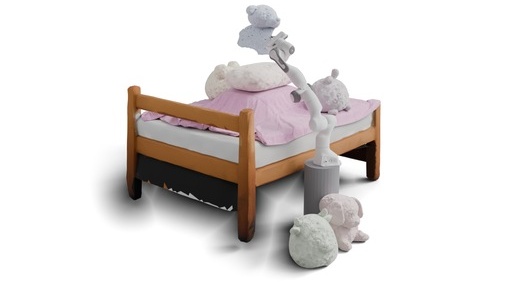} &
\includegraphics[width=\stripw]{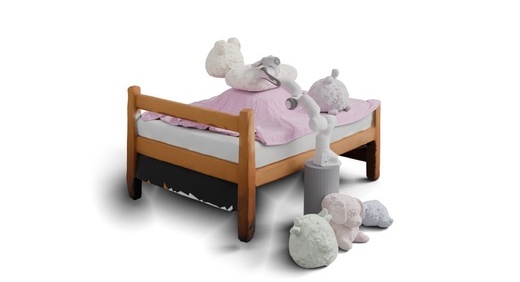} &
\includegraphics[width=\stripw]{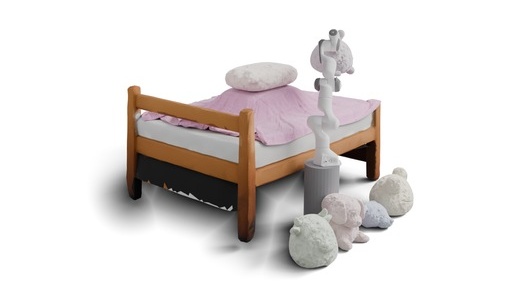} &
\includegraphics[width=\stripw]{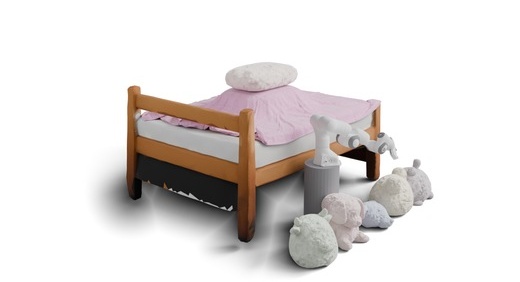} \\[10pt]
\srow{pull: cloth}{off stool} &
\includegraphics[width=\stripw]{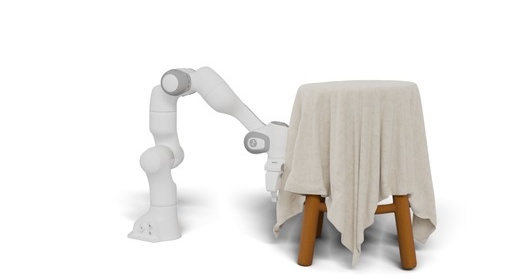} &
\includegraphics[width=\stripw]{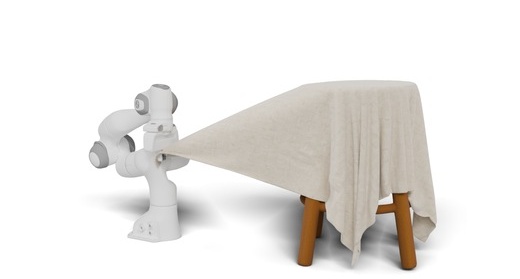} &
\includegraphics[width=\stripw]{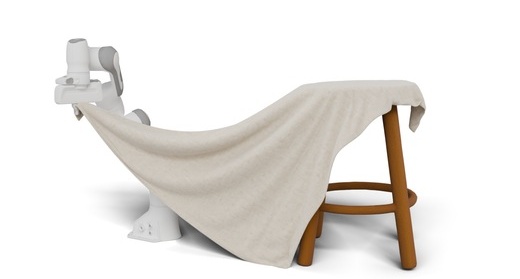} &
\includegraphics[width=\stripw]{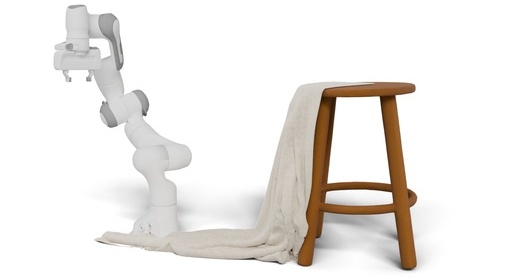} &
\includegraphics[width=\stripw]{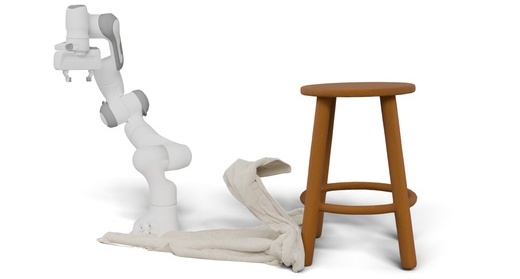} &
\includegraphics[width=\stripw]{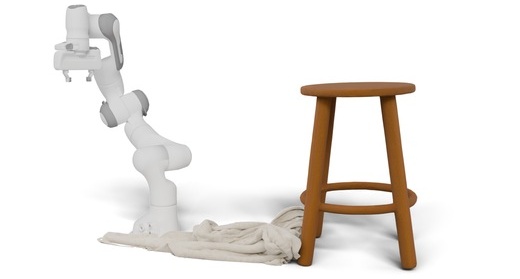}
\end{tabular}

\par\vspace{10pt}
\setlength{\tabcolsep}{2pt}%
\providecommand{\edw}{\dimexpr(\textwidth-0.27in-12pt)/4\relax}
\providecommand{\edc}[1]{\parbox[t]{\edw}{\centering\figcap{#1}}}
\noindent\begin{tabular}{@{}c@{\hspace{3pt}}cccc@{}}
\rotatebox{90}{\makebox[\dimexpr\edw*560/800\relax]{\shortstack{\eyebrow{edit: remove the throw}\\\eyebrow{and the toy, add a pillow}}}} &
\includegraphics[width=\edw]{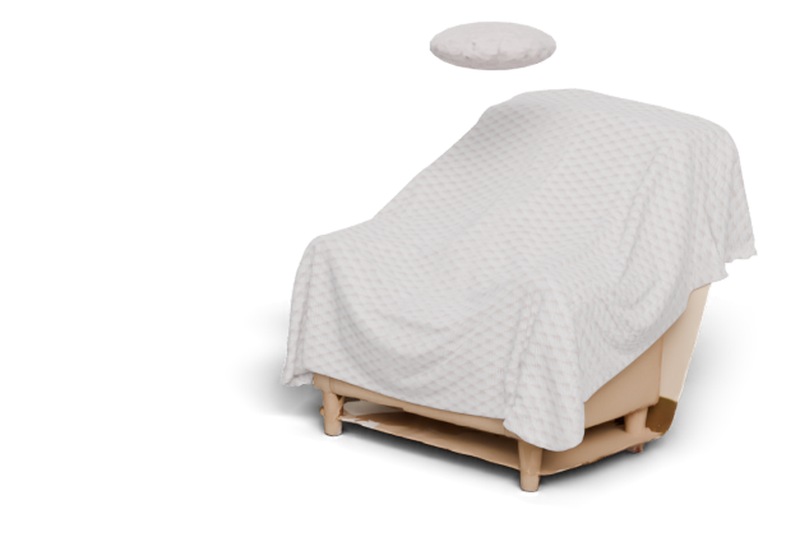} &
\includegraphics[width=\edw]{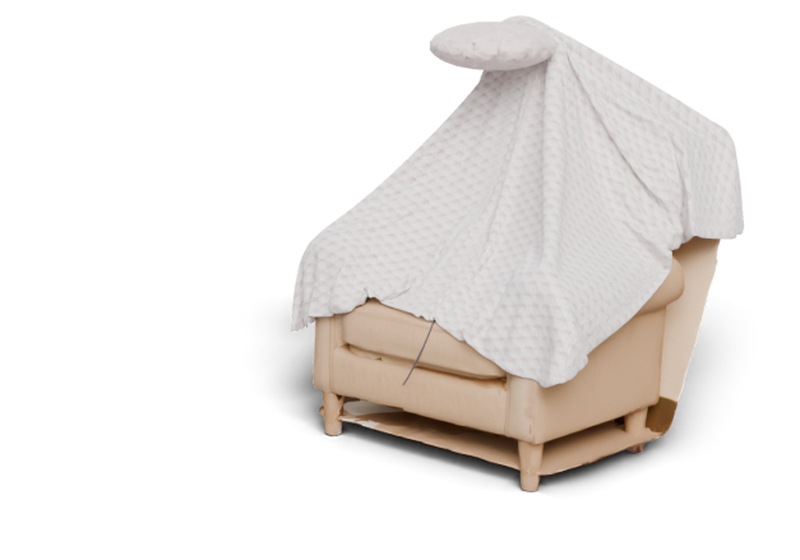} &
\includegraphics[width=\edw]{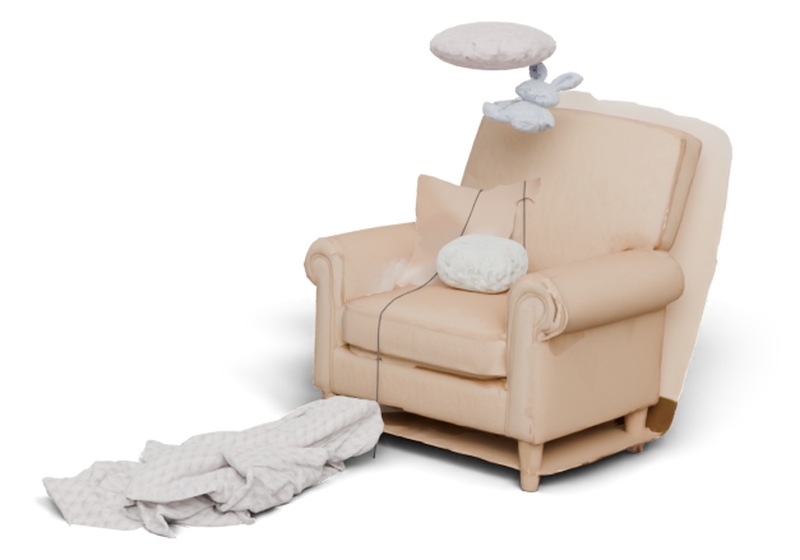} &
\includegraphics[width=\edw]{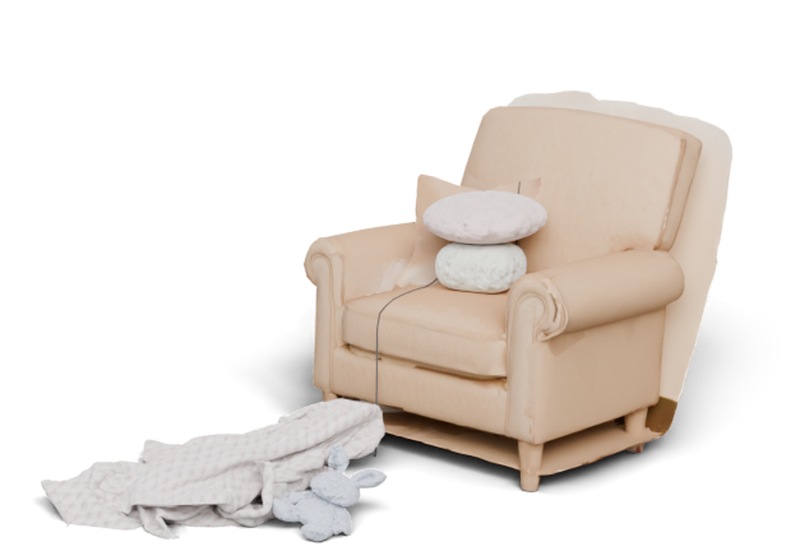} \\[0pt]
& \edc{the equilibrated nook, the pillow to add held above} & \edc{the throw gripped and lifted off the armchair} & \edc{the rabbit gripped and carried to the floor} & \edc{the pillow released onto the seat}
\end{tabular}

  \caption{\textbf{Scene editing and manipulation on equilibrated
  scenes.} Top: the five stuffed toys of the messy bed are set on the
  floor in turn while the quilt and pillows keep settling. Middle: a
  cloth pulled off a stool by one corner. Bottom: the throw is lifted
  off the armchair, the rabbit carried to the floor and a pillow placed
  on the seat, every body simulated; the arm is posed for visualization.}
  \label{fig:apps}
\end{figure*}

\subsection{Applications}
\begin{table}[t]
\centering\scriptsize
\setlength{\tabcolsep}{5pt}
\begin{tabular*}{\linewidth}{@{\extracolsep{\fill}}l cccc@{}}
\toprule
condition & S & G & J & C \\
\midrule
full stage, asset as rest shape & 22 & 19 & 14 & 14 (9) \\
\midrule
centred placement, hull-based support check & 32 & 28 & 17 & 17 \\
+ watertight collision mesh & 33 & 31 & 18 & 18 (13) \\
+ placement on the part the language model names & 33 & 30 & 20 & 20 (15) \\
+ its size estimate, full stage & 33 & 30 & 22 & \textbf{21} (16) \\
\bottomrule
\end{tabular*}

\caption{\textbf{Same-solver controls} on the \ablDevN{} development
scenes, same assets, solver and budget. First row: the generated asset's closed surface as the rest shape; below it the components of the physics stage are added one at a time; bold: the main run. In parentheses: the \ablDevSheetN{} cloth scenes alone. S: static equilibrium; G: geometric checks; J: prompt check; C: complete. Crossings are zero in every cell.}
\label{tab:control}
\end{table}
\label{sec:apps}

\paragraph{Robotic manipulation.} The same solver manipulates the equilibrated scene. A soft position
constraint grips a patch $\mathcal{G}$ of the deformable object and adds
\begin{equation}
\Psi_{\mathrm{grip}}(\bm{x})=\tfrac{k_g}{2}\sum_{i\in\mathcal{G}}
m_i\,\lVert\bm{x}_i-\bm{g}_i(t)\rVert^{2}
\label{eq:grip}
\end{equation}
to Eq.~\eqref{eq:ip}, where $\bm{g}_i(t)$ is the gripper path along which the constraint moves the patch, holds it and releases it; the arm in Fig.~\ref{fig:apps} is posed for visualization only. \emph{Pull}, \emph{lift} and \emph{drag} move a corner, the top centre or the patch, and \emph{press} compresses a stuffed body by half its height. Over
\ixEpisodes{} episodes on \ixScenes{} scenes, the patch reaches $90\%$
of its commanded displacement in \ixReached{} of the \ixMove{} pull,
lift and drag episodes, with a median grip error of \ixGripErr\,mm.

\paragraph{Scene editing.} The same mechanism removes a body by carrying it to the floor and adds one by holding and releasing it; on the messy bed (Fig.~\ref{fig:apps}) the added pillow's placement is optimized by the differentiable simulation (Sec.~\ref{sec:gradient}).

\subsection{Ablation studies}
\label{sec:control}

\begin{figure}[t]
  \centering
\setlength{\tabcolsep}{1.5pt}%
\providecommand{\ablw}{\dimexpr(\linewidth-9pt)/4\relax}
\providecommand{\ablc}[1]{\parbox[t]{\ablw}{\centering\eyebrow{#1}}}
\noindent\begin{tabular}{@{}cccc@{}}
\includegraphics[width=\ablw]{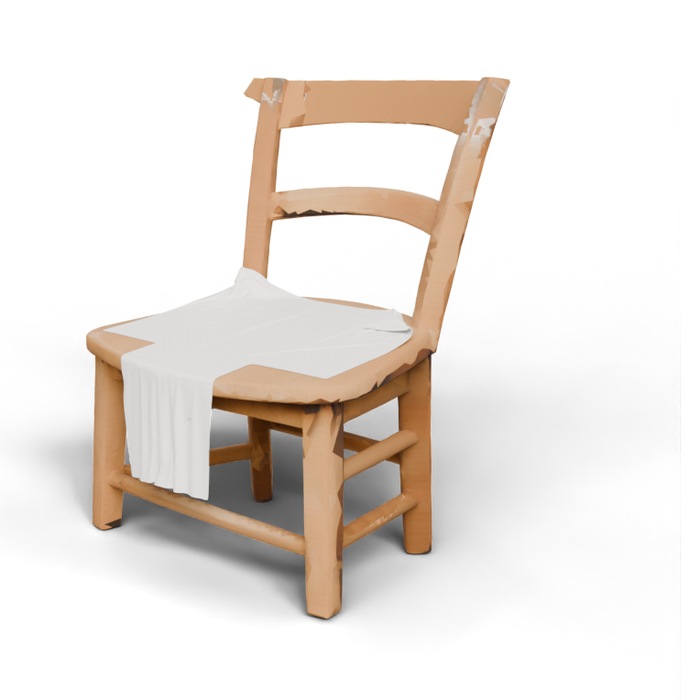} &
\includegraphics[width=\ablw]{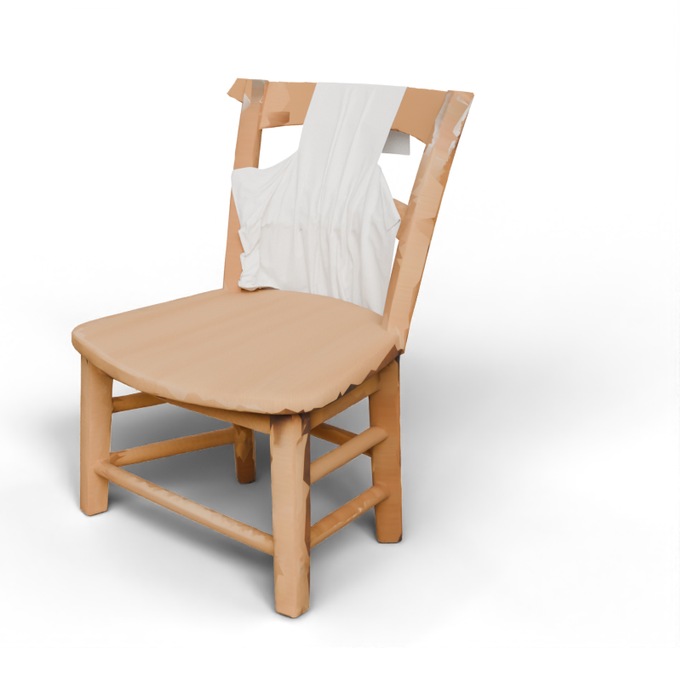} &
\includegraphics[width=\ablw]{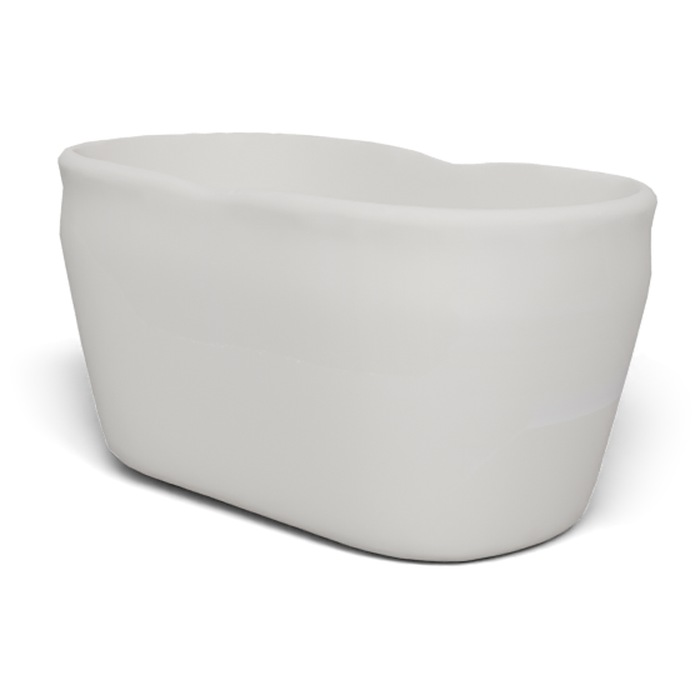} &
\includegraphics[width=\ablw]{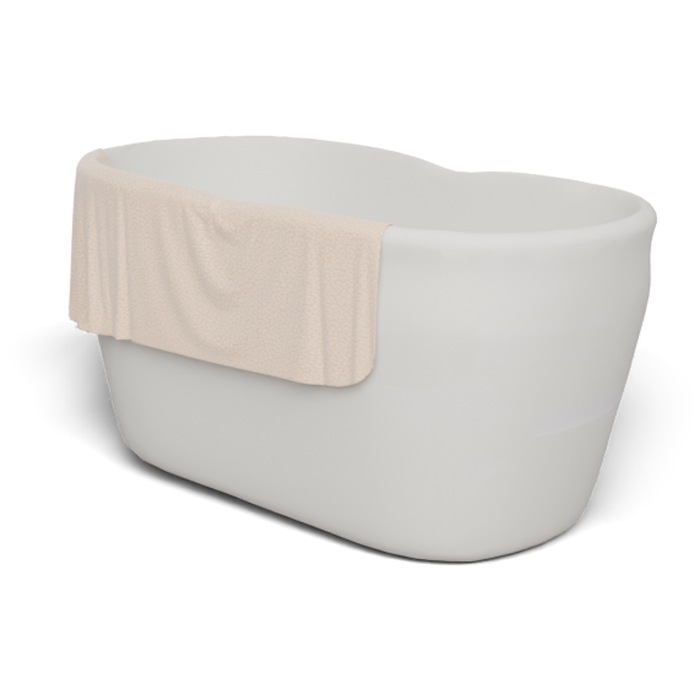} \\[-1pt]
\ablc{centred} & \ablc{named part} & \ablc{centred} & \ablc{named part} \\[-2pt]
\multicolumn{2}{c}{\figcap{apron over a chair back}} & \multicolumn{2}{c}{\figcap{towel over a tub rim}}
\end{tabular}

  \caption{\textbf{Placement ablation.} An apron and a towel placed centred on their support and on the part the language model names, both at the size it estimates.}
  \label{fig:abl}
\end{figure}

\paragraph{Same-solver controls.} In Table~\ref{tab:control}, the rest
shape accounts for the largest share: with the closed surface of the generated asset
in place of the sewn panel, the solver rejects the initial
state on 11 scenes and the completed cloth scenes fall from
\ablDevDSheetComplete{} to \assetDevSheetComplete{}. The
watertight collision mesh brings every scene to equilibrium, placing each object
on the part the language model names completes two more, and its
size estimate one more, from \ablDevAComplete{} for a sewn panel placed centred over the support to \ablDevDComplete{} for the full physics stage (Fig.~\ref{fig:abl}).

\paragraph{Differentiable simulation.}
\label{sec:gradient}
In the joint optimization of $N$ bodies over an obstructed support, it lands about twice as many bodies as a coordinate grid search with twice its forward budget, and remains the best of the four methods over eight supports and $32$ runs per $N$ (supplementary material), because the gradient crosses the obstruction that a line search does not. The pillow added to the messy bed (Sec.~\ref{sec:apps}) comes to rest $1.2$\,cm from its target under the adjoint's placement, against $67$\,cm under the language model's, and the same adjoint recovers a weight-to-stiffness ratio perturbed fourfold on \matScenes{} supports with a median error of \matOursMedPct\% over \matCases{} cases, against \matSearchMedPct\% for a random search at twice the budget (supplementary material). For a single body whose target lies on the grid's line, the grid of Eq.~\eqref{eq:grid} passes \budGrid{} of the \relcorrPos{} correctable scenes and the pipeline uses it; where an obstruction lies in between, the adjoint lands the body in $24$ of $32$ runs against $20$ for the grid (supplementary material).
 From the prompt alone, without the reference image, all \tsAllN{} scenes reach equilibrium and \tsAllGates{} pass both geometric checks, and \tsHoYes{} of the \tsHoN{} held-out prompts pass the prompt check against \tsHoImgYes{} with the image.\label{sec:t2s}

\section{Conclusion}
\label{sec:limits}

\begin{figure}[t]
  \centering
\setlength{\tabcolsep}{3pt}%
\providecommand{\failw}{\dimexpr(\linewidth-6pt)/2\relax}
\providecommand{\failc}[1]{\parbox[t]{\failw}{\centering\figcap{#1}}}
\noindent\begin{tabular}{@{}cc@{}}
\includegraphics[width=\failw]{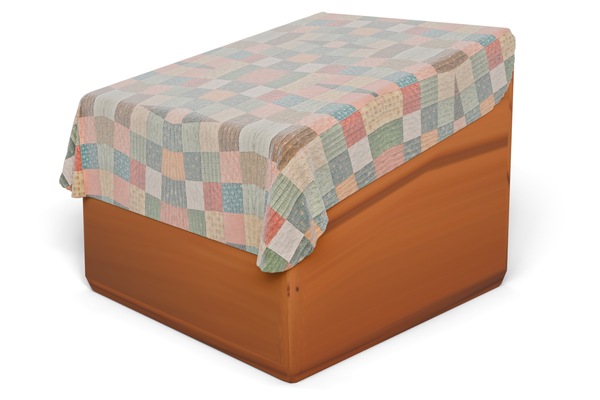} &
\includegraphics[width=\failw]{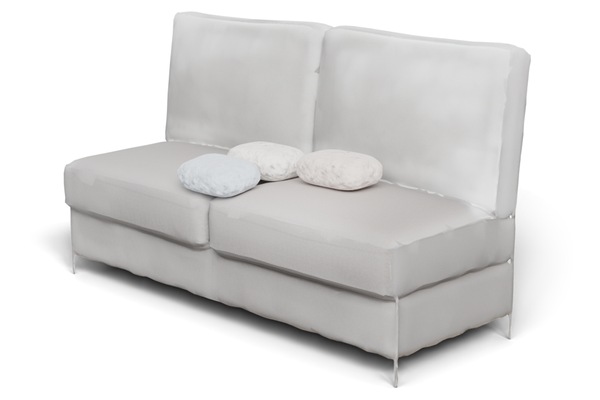} \\[-1pt]
\failc{quilt over the footboard of a bed} & \failc{three pillows lined up on a sofa}
\end{tabular}

  \caption{\textbf{Failure cases.} Left: the generated bed is a solid block with no footboard, so the quilt covers the whole bed. Right: pillows placed over the seat settle in a cluster, not in the row the prompt asks for.}
  \label{fig:fail}
\end{figure}

We presented SoftPAT3D, which constructs the inputs a deformable simulation requires and brings deformable objects to equilibrium in text-to-3D scene generation.
\textbf{Limitations.} A prompt that asks for a row of pillows or a folded towel is treated as a placement and fails the prompt check, and a support generated as a solid block has no part to place onto (Fig.~\ref{fig:fail}). The sewn panel keeps the generated object's kind in \idCat{} of \idN{} scenes and its colour in \idApp{}; its size and folds follow the class prior. Materials are effective values~\cite{forceaware}; plasticity is not modelled (cf.~\cite{codimrecon}).

\newpage  
\section*{LLM Usage Statement}
Large language models are components of the method and are evaluated as such: a language model returns the supporting part and the size of each object the prompt introduces (Sec.~\ref{sec:init}), and a vision--language model performs the prompt check (Sec.~\ref{sec:gates}). Beyond the method, LLM-based tools (Claude via Claude Code, and ChatGPT) were used at the assist level only, to edit and refine author-produced content: the text, the related work and citations, the code, the analysis of the experiments, the figures and the scene prompts. The authors take full responsibility for all submitted content.

\clearpage
\appendix
\twocolumn[{\begin{center}
{\Large\bfseries Supplementary Material}\\[4pt]
{\large SoftPAT3D: Physics-Augmented Text-to-3D Scene Generation with Deformables}\\[14pt]
\end{center}}]
\setcounter{figure}{0}
\renewcommand{\thefigure}{S\arabic{figure}}
\setcounter{table}{0}
\renewcommand{\thetable}{S\arabic{table}}

\begin{figure*}[!t]
  \centering
  \includegraphics[width=\textwidth]{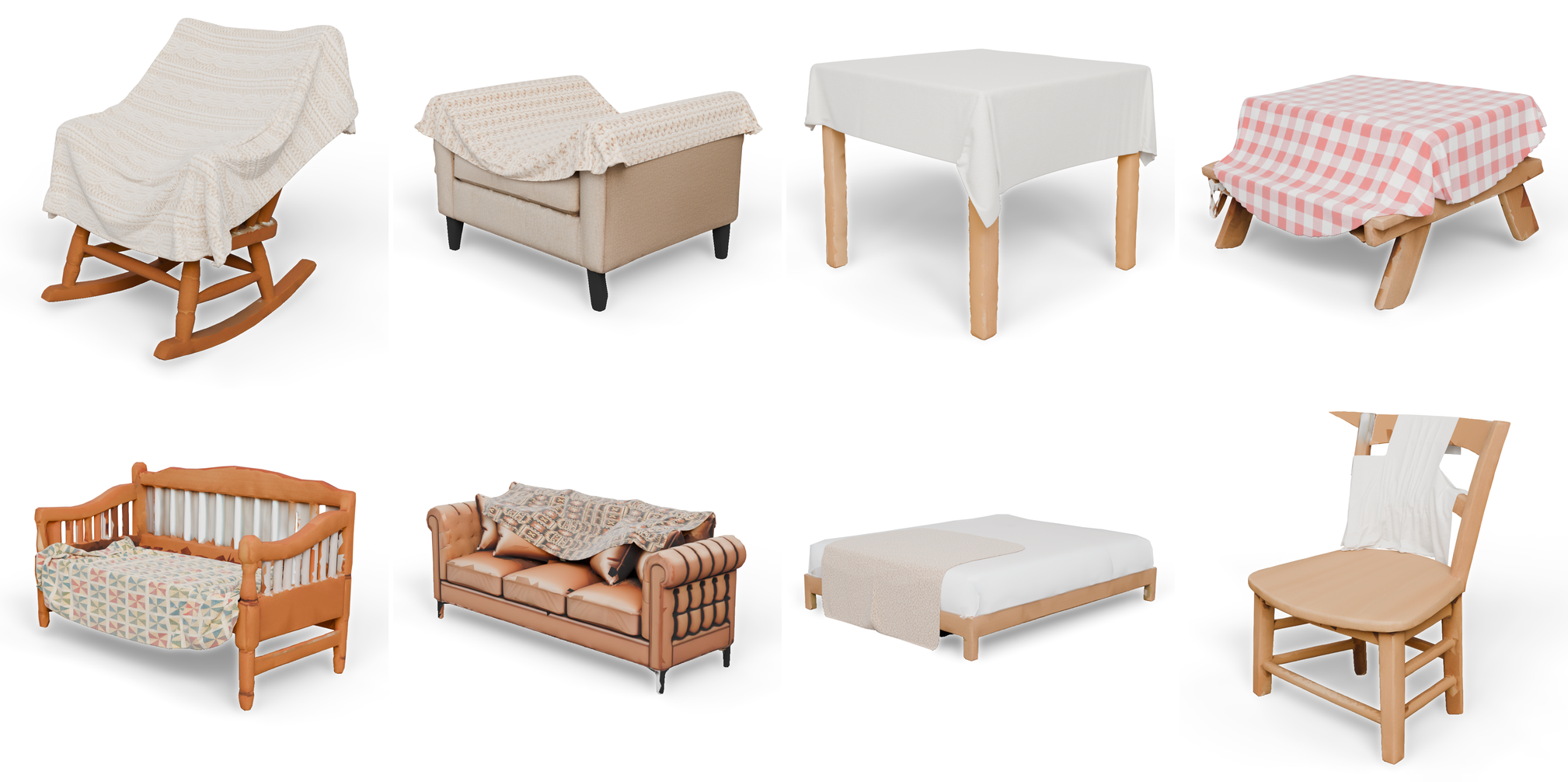}
  \caption{\textbf{Equilibrated scenes}, rendered from the solver's own
  output with the cloth texture extracted from each scene's reference
  image and the furniture's regenerated texture, under one studio
  environment map (HDRI).
  Top: a blanket over a rocking chair, a knit blanket over an armchair,
  a tablecloth on a square dining table, a gingham cloth on a picnic
  table. Bottom: a quilt on a daybed pushed flush to its back and
  hanging over the front rail, a patterned throw over the back of a
  leather sofa, a sheet folded over the foot of a bed, and an apron
  hung over the back of a kitchen chair. Eight of the thirty-two
  scenes are shown. Further scenes are shown in Fig.~\ref{fig:supp-all},
  and all thirty-two are reported with their numbers in
  Table~\ref{tab:sceneset}, including the ones that a geometric check
  rejected.
Because the renderer draws the solver's collider with the asset's
texture projected onto it, a render cannot show contact that the
simulation did not produce.}
  \label{fig:supp-gallery}
\end{figure*}

\section{Scene Set}
\label{supp:scenes}

In this section, we enumerate every scene that the pipeline is asked to equilibrate, including the rejected ones, so that each count in the main paper can be traced to its corresponding scenes. Table~\ref{tab:sceneset} is generated from the development runs rather than maintained by hand, and every rejected scene carries the check that rejects it. The final code's outcome for each scene is given in the last block of Table~\ref{tab:supp-2x2}, and Fig.~\ref{fig:supp-all} renders every scene that passes the geometric checks.

\begin{table*}[t]
\centering\footnotesize
\setlength{\tabcolsep}{4pt}
\begin{minipage}[t]{0.47\textwidth}
\begin{tabular*}{\linewidth}{@{\extracolsep{\fill}}l r r r r@{}}
\toprule
scene & bodies & crossings & contact (\%) & blocked (\%) \\
\midrule
apron & 1 & 0 & 23 & 17 \\
armchair & 1 & 0 & 38 & 0 \\
bathtub2 & 1 & 0 & 55 & 7 \\
bathtub & 1 & 0 & 4 & 11 \\
bedsheet & 1 & 0 & 56 & 0 \\
bench2b & 1 & 0 & 61 & 0 \\
bench & 1 & 0 & 65 & 0 \\
blanket & 1 & 0 & 32 & 0 \\
box & 1 & 0 & 98 & 0 \\
clothes sofa & 6 & 0 & 50 & 4 \\
coffee & 1 & 0 & 100 & 0 \\
couch & 1 & 0 & 3 & 0 \\
cushion & 1 & 0 & 4 & 0 \\
daybed & 1 & 0 & 34 & 0 \\
dining2 & 1 & 0 & 45 & 14 \\
dining & 1 & 0 & 27 & 18 \\
kite & 1 & 0 & 99 & 0 \\
picnic & 1 & 0 & 49 & 2 \\
\bottomrule
\end{tabular*}
\end{minipage}\hfill
\begin{minipage}[t]{0.47\textwidth}
\begin{tabular*}{\linewidth}{@{\extracolsep{\fill}}l r r r r@{}}
\toprule
scene & bodies & crossings & contact (\%) & blocked (\%) \\
\midrule
pillow2 & 1 & 0 & 9 & 2 \\
pillow3 & 1 & 0 & 6 & 0 \\
pillow & 1 & 0 & 9 & 0 \\
plush & 1 & 0 & 2 & 0 \\
rocking & 1 & 0 & 4 & 11 \\
skirt & 1 & 0 & 60 & 4 \\
stool2 & 1 & 0 & 39 & 15 \\
stool & 1 & 0 & 15 & 22 \\
throw & 1 & 0 & 51 & 0 \\
towel4 & 1 & 0 & 31 & 28 \\
trash & 1 & 0 & 10 & 0 \\
\midrule
bedroom & \multicolumn{4}{r}{small table scale (asset)} \\
ladder & \multicolumn{4}{r}{towel fell-off-support} \\
laundry & \multicolumn{4}{r}{laundry basket solid (asset)} \\
napkin & \multicolumn{4}{r}{wooden table support-off-scale} \\
steamer & \multicolumn{4}{r}{dumplings frozen-world} \\
tissue & \multicolumn{4}{r}{tissue centroid-off-support} \\
\bottomrule
\end{tabular*}
\end{minipage}

\caption{\textbf{The full scene set.} Every scene that any version of
the pipeline was asked to equilibrate is included, including the ones that
earlier versions rejected. Crossings are cloth edges passing through a
support, counted against the collider the solver used. ``Blocked'' is
the upward-ray measure of Sec.~\ref{supp:compare}, the fraction of
cloth vertices with support surface above them. It is large where a
body has come to rest on something other than the top. Scenes below the
rule were rejected and are labelled with what the containment or support
check found.}
\label{tab:sceneset}
\end{table*}

\begin{figure*}[!t]
  \centering
  \setlength{\tabcolsep}{0pt}%
\noindent\begin{tabular}{@{}c@{\hspace{0.03in}}c@{\hspace{0.03in}}c@{\hspace{0.03in}}c@{\hspace{0.03in}}c@{\hspace{0.03in}}c@{\hspace{0.03in}}c@{}}
\includegraphics[width=0.956in]{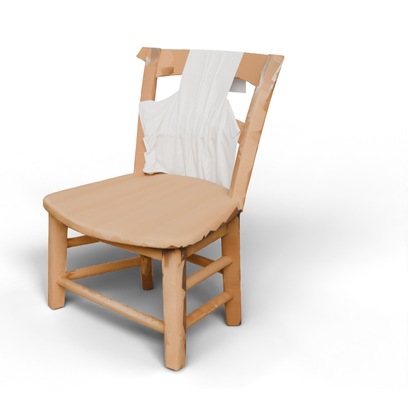} & \includegraphics[width=0.956in]{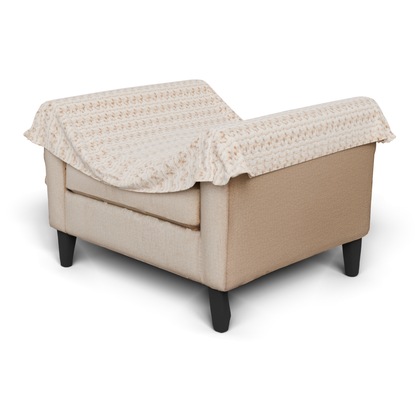} & \includegraphics[width=0.956in]{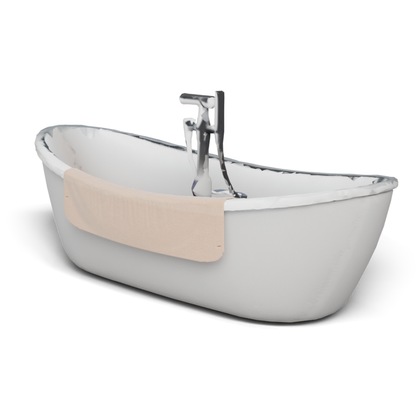} & \includegraphics[width=0.956in]{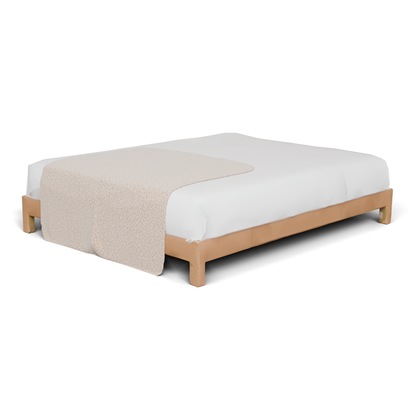} & \includegraphics[width=0.956in]{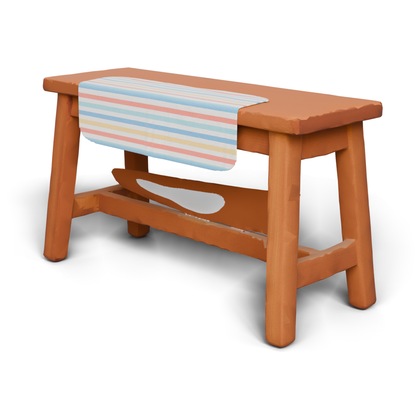} & \includegraphics[width=0.956in]{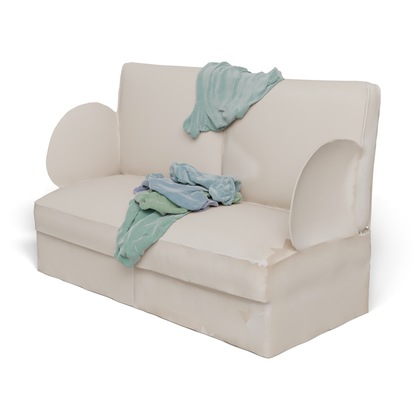} & \includegraphics[width=0.956in]{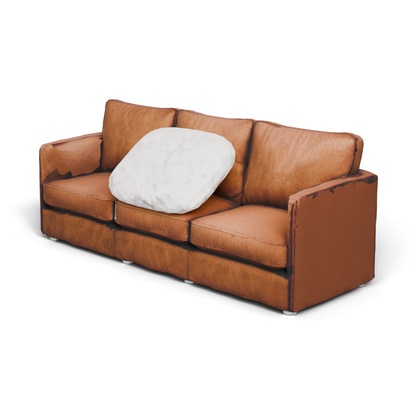} \\[1pt]
\footnotesize apron & \footnotesize armchair & \footnotesize bathtub & \footnotesize bedsheet & \footnotesize bench2b & \footnotesize clothes sofa & \footnotesize couch \\[4pt]
\includegraphics[width=0.956in]{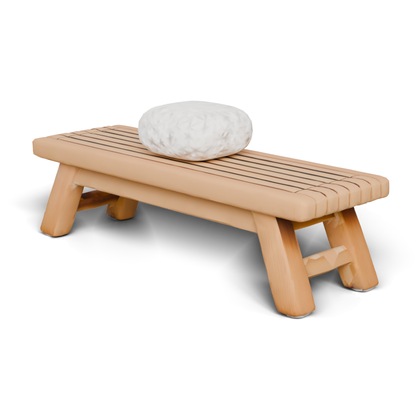} & \includegraphics[width=0.956in]{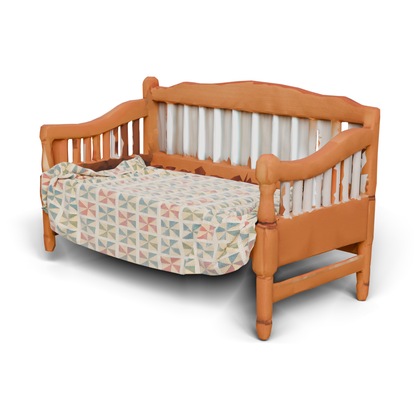} & \includegraphics[width=0.956in]{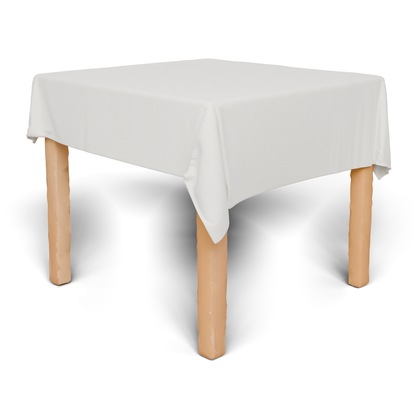} & \includegraphics[width=0.956in]{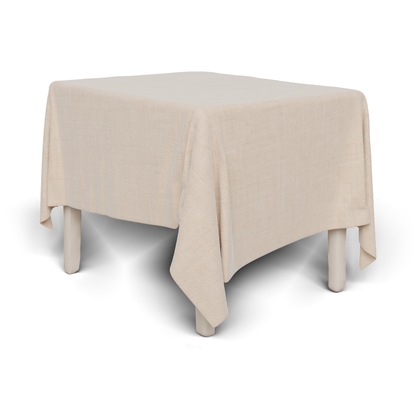} & \includegraphics[width=0.956in]{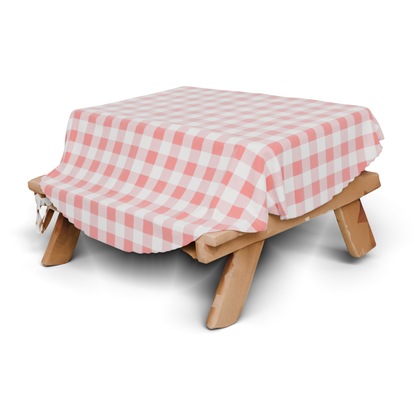} & \includegraphics[width=0.956in]{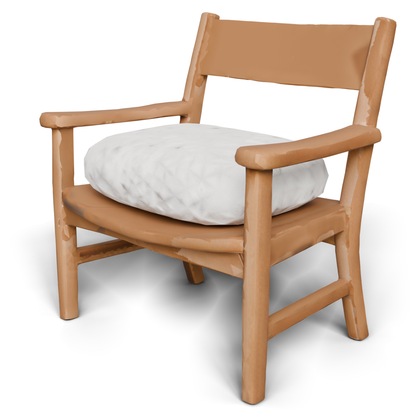} & \includegraphics[width=0.956in]{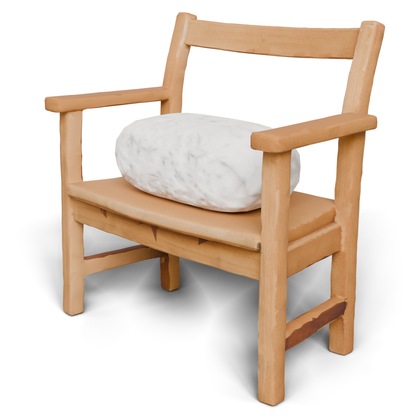} \\[1pt]
\footnotesize cushion & \footnotesize daybed & \footnotesize dining2 & \footnotesize dining & \footnotesize picnic & \footnotesize pillow2 & \footnotesize pillow3 \\[4pt]
\includegraphics[width=0.956in]{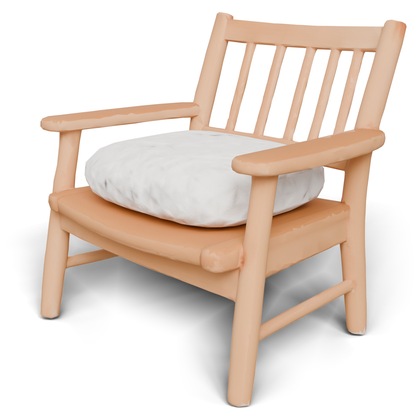} & \includegraphics[width=0.956in]{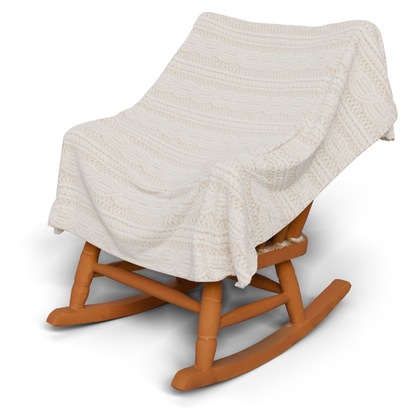} & \includegraphics[width=0.956in]{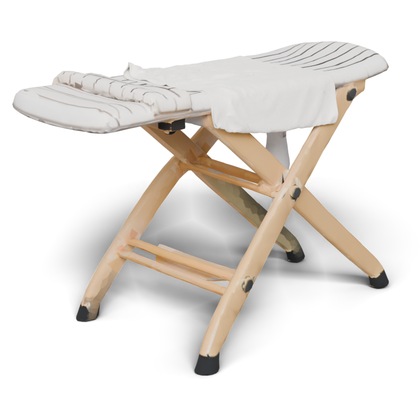} & \includegraphics[width=0.956in]{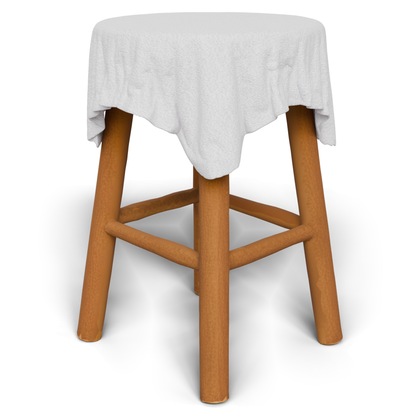} & \includegraphics[width=0.956in]{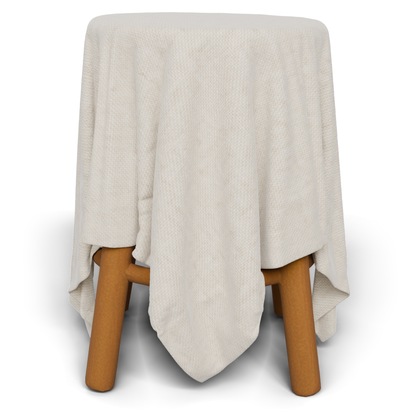} & \includegraphics[width=0.956in]{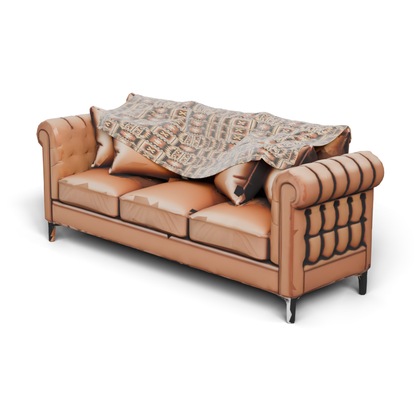} & \includegraphics[width=0.956in]{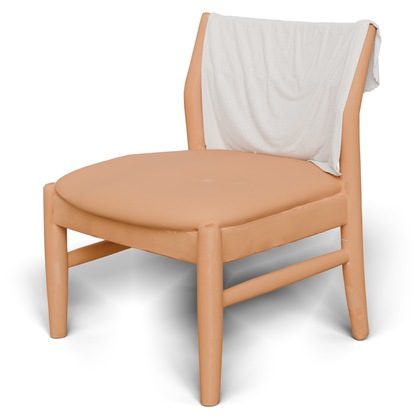} \\[1pt]
\footnotesize pillow & \footnotesize rocking & \footnotesize skirt & \footnotesize stool2 & \footnotesize stool & \footnotesize throw & \footnotesize towel4 \\[4pt]
\end{tabular}

  \caption{\textbf{Scenes of the main set that pass the geometric checks}, in the development simulations, under the same camera, lighting and texture pipeline as the gallery. Furniture is drawn as the solver's collider with the asset's texture projected onto it, and the cloth as the solver left it.}
  \label{fig:supp-all}
\end{figure*}

\section{Held-out Prompts}
\label{supp:heldout}

In this section, we provide the per-prompt record underlying the held-out count of the main paper. The thirty-two prompts of Table~\ref{tab:sceneset} are written while the pipeline is being built. After the pipeline is frozen, we write \heldoutN{} more prompts over supports it has not been run on and run them end to end, with no asset redraw, placement override or parameter change, under the same solver settings as the main set. Table~\ref{tab:heldout} lists every prompt with its outcome.

We manually verify the routing against the prompts. Among the 63 objects that reach the solver, every towel, blanket, jacket, scarf, skirt, quilt, shirt, sweater, throw and tablecloth is routed to the shell energy, every cushion, pillow and plush toy is routed to the volumetric-body energy, and every support is routed to the rigid energy, with no error.

The criterion is the one used throughout the paper. ``Complete'' means that the scene passes the containment and support checks, has zero cloth--support crossings on the colliders, and passes the prompt check of the main paper's Sec.~3.5, run on one render per scene with the same script and the same two-answer rule as the main set and the controls. A second pass of the check over the same twenty-four renders, with two new answers each, returned the same verdict on every scene. ``Upstream'' means that the scene never reached the solver, because the generation stage's own object list omitted one of the two requested objects. The outcome listed is that of the final code's run over the held-out assets, the full-stage cell of Sec.~\ref{supp:2x2}: \ablHoDComplete{} of \heldoutN{} complete.

\paragraph{Blender's cloth solver.} Given the same held-out inputs, the sewn sheet at our placement over our closed colliders exhibits crossings at \hobBlCross{} edges per thousand under Blender's cloth solver and leaves \hobBlFloat{} of its \hobBodies{} bodies touching nothing, against zero and none for ours, and its cloth touches the support over \hobBlContact\% of its vertices against \hobOursContact\% for ours (Table~\ref{tab:supp-heldout-blender}). This is the shared-input condition that Blender$^{*}$ runs on the development set, here over the \hobN{} held-out scenes that both solvers produced.

\begin{table*}[t]
\centering\scriptsize
\setlength{\tabcolsep}{4pt}
\begin{minipage}{0.7\textwidth}
\begin{tabular*}{\linewidth}{@{\extracolsep{\fill}}l r r r r r r r@{}}
\toprule
 & bodies & cross.\ (per 1k) & pen.\ (mm) & contact (\%) & floating & unsup. & zero-cross.\ scenes \\
\midrule
Blender$^{*}$ & 27 & 24.4 & 0.1 & 17.5 & 9/27 & 0/27 & 23/25 \\
\textbf{ours} & 27 & 0.0 & 0.0 & 52.5 & 0/27 & 0/27 & 25/25 \\
\bottomrule
\end{tabular*}

\end{minipage}
\caption{\textbf{Blender's cloth solver and ours on the held-out batch}, same sewn sheets, placements and colliders, over the \hobN{} held-out scenes both produced; the metrics of the main paper's Table~3.}
\label{tab:supp-heldout-blender}
\end{table*}

\paragraph{Results with placement optimization.} Seven of the scenes that fail the prompt check name a part that a placement can reach: the towel on the saddle, the counter towel, the skirt, the quilt on the footboard, the hamper towel, the marble-table cloth and the piano scarf. Sec.~\ref{supp:relcorr} re-solves them under the pipeline's rule, and five pass, while the closed loop completes \hoClosed{} of \heldoutN{}. The grid of Eq.~9 of the main paper passes these five at six re-simulations per scene (Table~\ref{tab:supp-heldout-compute}).

\begin{table}[tb]
\centering\footnotesize
\setlength{\tabcolsep}{3pt}
\begin{tabular*}{\linewidth}{@{\extracolsep{\fill}}l cccc@{}}
\toprule
configuration & S & G & J & C \\
\midrule
physics stage & \ablHoDSettled{} & \ablHoDGates{} & \ablHoDPrompt{} & \ablHoDComplete{} \\
+ placement grid of Eq.~9 (the pipeline) & \ablHoDSettled{} & \ablHoDGates{} & \hoClosedGrid{} & \hoClosedGrid{} \\
\bottomrule
\end{tabular*}

\caption{\textbf{Held-out completion at each stage of placement optimization.} Settled (S), passing the geometric checks (G), passing the prompt check (J) and complete (C) out of \heldoutN{}, Eq.~9 is the placement grid of the main paper, which is the single-body correction of the pipeline, run at six re-simulations per scene.}
\label{tab:supp-heldout-compute}
\end{table}

\begin{table*}[t]
\centering\scriptsize
\setlength{\tabcolsep}{3pt}
\begin{tabular}{l p{0.62\linewidth} l}
\toprule
scene & prompt & outcome\\
\midrule
bedfoot & A folded wool blanket lying across the foot of a neatly made bed. & complete \\
bookshelf & A knitted blanket draped over the top of a low wooden bookshelf. & complete \\
counter & A kitchen towel draped over the edge of a small kitchen island counter. & incomplete \\
drawers & A pleated skirt draped over the top of a chest of drawers. & incomplete \\
footboard & A patchwork quilt draped over the wooden footboard of a bed. & incomplete \\
ironing2 & A white dress shirt lying flat on an ironing board. & complete \\
ottoman & A fleece blanket draped over a round leather ottoman. & complete \\
stepstool & A towel draped over the top step of a small wooden step stool. & complete \\
stool3 & A tablecloth draped over a tall wooden stool. & complete \\
stump & A picnic blanket draped over a large tree stump. & complete \\
suitcase & A sweater draped over a closed hard-shell suitcase lying flat on the floor. & complete \\
trunk & A quilt draped over a wooden storage trunk. & complete \\
barstool2 & A dish towel draped over the seat of a metal bar stool. & complete \\
benchcushion & A long seat cushion lying on a wooden garden bench. & incomplete \\
bicycle & A towel draped over the saddle of a bicycle leaning on its kickstand. & incomplete \\
crib & A baby blanket draped over the side rail of a wooden crib. & complete \\
deskchair & A denim jacket draped over the back of a wooden desk chair. & complete \\
hamper & A folded bath towel resting on top of a closed wicker laundry hamper. & incomplete \\
marble & A small square cloth draped over a round marble side table. & incomplete \\
metalchair & A tea towel draped over the back of a metal kitchen chair. & complete \\
piano & A silk scarf draped over the closed lid of an upright piano. & incomplete \\
radiator & A towel draped over a white wall radiator. & complete \\
runner & A long narrow cloth draped across a rectangular wooden dining table. & incomplete \\
sofapillows & Three throw pillows lined up on a grey fabric sofa. & incomplete \\
sink & A hand towel draped over the edge of a white bathroom sink on a cabinet. & incomplete \\
gardenchair & A beach towel draped over a wooden garden chair. & incomplete \\
glasstable & A knitted blanket thrown over a low glass coffee table. & incomplete \\
cot & A wool blanket spread over a camping cot, outdoors. & incomplete \\
nightstand & A cloth napkin draped over the edge of a wooden nightstand. & incomplete \\
sweaters & Two sweaters piled on the seat of a wooden chair. & incomplete \\
toybox & A plush toy sitting on top of a closed wooden toy box. & incomplete \\
velvet & A fleece throw draped over the seat and one armrest of a velvet armchair. & incomplete \\
\bottomrule
\end{tabular}

\caption{\textbf{The held-out batch.} The \heldoutN{} prompts written
after the pipeline was frozen, each with the outcome of the final
code's run on the same assets, the full-stage cell of
Sec.~\ref{supp:2x2}, which is the number reported in the main paper.
``Complete'' is the full criterion of the main set. Every prompt ends with a setting clause, ``in a simple room'' or similar, omitted here. \ablHoDComplete{} of \heldoutN{} are
complete.}
\label{tab:heldout}
\end{table*}

\section{Same-solver Controls}
\begin{table}[tb]
\centering\scriptsize
\begin{tabular}{@{}p{0.15\linewidth}p{\dimexpr0.85\linewidth-2\tabcolsep\relax}@{}}
\toprule
\multicolumn{2}{@{}l}{\textit{Placement} (Sec.~3.4 of the main paper)} \\
part & the planner names a surface, a part (seat, backrest, arm, rim) and a drop height; objects seen in the image keep their layout place \\
straddle & a sheet that fits inside the seat is centred on the seat's front edge rather than on the seat \\
flush & a sheet deeper than a backed seat is pushed against the back, so the excess hangs over the front \\
throw & a throw on seating goes over the backrest unless the planner names another part \\
order & coverings first, smaller items on top; a sheet in a container is shrunk below the opening; line-mounted cloth starts vertical, top row pinned \\
lift & every release is tested for intersection and lifted until it is free \\
\midrule
\multicolumn{2}{@{}l}{\textit{Completion} (Sec.~3.5 of the main paper)} \\
containment & the deformable object's centre of mass must project inside its support's ground projection \\
support & a support nearly as solid as its bounding box, or far outside its class's size range, is a generation failure \\
prompt & one render, one question, put twice to a vision--language model; pass when both answers agree \\
complete & every requested deformable object present, static equilibrium, no deformable edge crossing a collider, both geometric checks and the prompt check passed \\
\bottomrule
\end{tabular}

\caption{\textbf{Placement rules and completion tests.} The rules of Sec.~3.4 and the tests of Sec.~3.5 of the main paper
convert the language model's answer into an intersection-free release, and the
tests decide whether an equilibrated scene is complete.}
\label{tab:supp-rules}
\end{table}

\label{supp:controls}

In this section, we provide the per-scene record of the same-solver controls of Table~5 of the main paper, and we identify the scene for which each rule was introduced. Table~\ref{tab:supp-control-scenes} lists, for each of the four conditions, whether the simulation reaches static equilibrium, whether the two geometric checks pass, whether the vision--language model, asked once on one render, finds the deformable object where the prompt puts it, and the number of crossings against the colliders.

The conditions are set up as follows. The two without the watertight collision mesh, base and placement, are run on the nine scenes whose support the earlier support check replaced by a convex hull: the bathtub, the bed, the two benches, the coffee table, the steamer, the trash bin, the tissue box and the napkin's table. On every other scene, the earlier support check returns the same collider, so those rows repeat the matching closed-collider condition. The centred placement puts the sheet at the geometric centre of its support with the class size and no part, with the language model, the three placement rules and the container and line-mount rules disabled. Note that, because the language model is not deterministic in the size it returns, the runs without the watertight collision mesh fix the size at the value the final run is given where that value is known.

\begin{table}[tb]
\centering\scriptsize
\setlength{\tabcolsep}{3pt}
\begin{tabular*}{\linewidth}[t]{@{\extracolsep{\fill}}l cccc c cccc c cccc c cccc@{}}
\toprule
& \multicolumn{4}{c}{base} & & \multicolumn{4}{c}{+ collision mesh} & & \multicolumn{4}{c}{+ placement} & & \multicolumn{4}{c}{full} \\
\cmidrule(lr){2-5}\cmidrule(lr){7-10}\cmidrule(lr){12-15}\cmidrule(lr){17-20}
scene & S & G & J & x & & S & G & J & x & & S & G & J & x & & S & G & J & x \\
\midrule
apron & y & y & n & 0 & & y & y & n & 0 & & y & y & y & 0 & & y & y & y & 0 \\
armchair & y & y & n & 0 & & y & y & n & 0 & & y & y & y & 0 & & y & y & y & 0 \\
bathtub2 & y & y & n & 0 & & y & y & n & 0 & & y & y & y & 0 & & y & y & y & 0 \\
bathtub & y & y & y & 0 & & y & y & y & 0 & & y & y & y & 0 & & y & y & y & 0 \\
bedsheet & y & y & n & 0 & & y & y & n & 0 & & y & y & n & 0 & & y & y & n & 0 \\
bench2b & y & y & y & 0 & & y & y & y & 0 & & y & y & y & 0 & & y & y & y & 0 \\
bench & y & y & y & 0 & & y & y & y & 0 & & y & y & y & 0 & & y & y & y & 0 \\
blanket & y & y & n & 0 & & y & y & n & 0 & & y & y & n & 0 & & y & y & n & 0 \\
box & y & y & n & 0 & & y & y & n & 0 & & y & y & y & 0 & & y & y & y & 0 \\
clothes sofa & y & y & n & 0 & & y & y & n & 0 & & y & y & y & 0 & & y & y & y & 0 \\
coffee & y & y & y & 0 & & y & y & y & 0 & & y & y & y & 0 & & y & y & y & 0 \\
couch & y & y & y & 0 & & y & y & y & 0 & & y & y & y & 0 & & y & y & y & 0 \\
cushion & y & y & y & 0 & & y & y & y & 0 & & y & y & y & 0 & & y & y & y & 0 \\
daybed & y & y & y & 0 & & y & y & y & 0 & & y & y & y & 0 & & y & y & y & 0 \\
dining2 & y & y & y & 0 & & y & y & y & 0 & & y & y & y & 0 & & y & y & y & 0 \\
dining & y & y & n & 0 & & y & y & n & 0 & & y & y & n & 0 & & y & y & n & 0 \\
kite & y & y & n & 0 & & y & y & n & 0 & & y & y & n & 0 & & y & y & n & 0 \\
ladder & y & n & n & 0 & & y & n & n & 0 & & y & n & y & 0 & & y & n & y & 0 \\
napkin & y & n & n & 0 & & y & n & n & 0 & & y & n & n & 0 & & y & n & n & 0 \\
picnic & y & y & y & 0 & & y & y & y & 0 & & y & y & y & 0 & & y & y & y & 0 \\
pillow2 & y & y & y & 0 & & y & y & y & 0 & & y & y & y & 0 & & y & y & y & 0 \\
pillow3 & y & y & y & 0 & & y & y & y & 0 & & y & y & y & 0 & & y & y & y & 0 \\
pillow & y & y & n & 0 & & y & y & n & 0 & & y & y & n & 0 & & y & y & n & 0 \\
plush & y & y & y & 0 & & y & y & y & 0 & & y & y & y & 0 & & y & y & y & 0 \\
rocking & y & y & y & 0 & & y & y & y & 0 & & y & y & y & 0 & & y & y & y & 0 \\
skirt & y & y & y & 0 & & y & y & y & 0 & & y & y & y & 0 & & y & y & y & 0 \\
steamer & y & n & n & 0 & & y & y & n & 0 & & y & n & n & 0 & & y & n & n & 0 \\
stool2 & y & y & y & 0 & & y & y & y & 0 & & y & y & y & 0 & & y & y & y & 0 \\
stool & y & y & y & 0 & & y & y & y & 0 & & y & y & y & 0 & & y & y & y & 0 \\
throw & y & y & y & 0 & & y & y & y & 0 & & y & y & y & 0 & & y & y & y & 0 \\
tissue & y & n & n & 0 & & y & y & n & 0 & & y & n & n & 0 & & y & n & n & 0 \\
towel4 & y & y & n & 0 & & y & y & n & 0 & & y & y & n & 0 & & y & y & n & 0 \\
trash & -- & -- & -- & -- & & y & y & y & 0 & & -- & -- & -- & -- & & y & y & y & 0 \\
\bottomrule
\end{tabular*}

\caption{\textbf{Per-scene results of the same-solver controls} on the
development set. S: reached static equilibrium; G: containment and
support checks; J: prompt check; x: crossings per thousand edges. A
dash marks a run without an equilibrium state to score.}
\label{tab:supp-control-scenes}\label{tab:supp-control-scenes-ho}
\end{table}

Read scene by scene, the table shows where each component changes an outcome. The closed colliders alone change one outcome: the tissue box's convex hull has no recess to hold the sheet, and the trash bin's convex hull has no inside at all, so the bin's two runs without the watertight collision mesh end in a solver assertion and are the two entries not at equilibrium. The placement rules change nine outcomes. The apron, the armchair's blanket, the box, the ladder's towel and the second bathtub are found by the prompt check on the wrong part of the right support when released at the centre, and pass once we use the part the language model names.

Each rule of the stage is introduced for a particular scene. Table~\ref{tab:supp-ablation} records these scenes, with each component replaced by what precedes it and the scene re-solved. The reported numbers are obtained from the re-solved states. The table shows the scenes from which the rules originate; however, their value over the set is what the controls above measure.

\begin{table}[tb]
\centering\scriptsize
\setlength{\tabcolsep}{3pt}
\setlength{\tabcolsep}{2.4pt}%
\begin{tabular*}{\linewidth}{@{\extracolsep{\fill}}p{0.30\linewidth}p{0.53\linewidth}@{\hspace{3pt}}r@{}}
\toprule
component $\to$ what preceded it & effect without it & scenes \\
\midrule
sewing-pattern rest shape $\to$ generated asset & layers closer than $\hat d$, world invalid at step 0, $0.0$\,mm displacement; the render still shows a drape & 6 of 6 \\
intersection-free closure $\to$ watertightness test & nine supports fall back to convex hulls; cloth equilibrates on a block, a wedge, a tub with no interior & 9 of 32 \\
convex-cell collider $\to$ single hull & Hausdorff error to the asset $82.8$ vs $8.4$\,mm; the towel hangs through the chair's open back & 1 of 1 \\
placement rules $\to$ planner's point & the sheet rests as a flat rectangle centred on the seat and passes every check & 6 of 32 \\
support de-occlusion $\to$ masked crop & the generator fills the mask with cloth fused into the support; the sheet drapes over it & 32 of 33 \\
gradient release correction $\to$ straight drop & obstructed targets within $3$\,cm: 1 straight, 4 random search, 6 gradient & 14 targets \\
\bottomrule
\end{tabular*}

\caption{\textbf{Origin of each placement rule.} Each component replaced by what preceded it, on the scenes
where it was introduced.}
\label{tab:supp-ablation}
\end{table}

\subsection{Size and placement}
\label{supp:2x2}

The placement determines two aspects, namely where the sheet is released and how large it is, and Table~3 of the main paper separates these two factors on both sets. Every cell is solved on 2026-10-21 from one code version, under the final collision-mesh closing, and on the same assets. The language model is queried once per scene, in the full-stage cell, and its two answers, a part and a size in metres, are recorded and fixed in the other cells, so that its non-determinism is excluded from the comparison.

The four cells correspond to the centred placement at the class size, the same release at the size the language model estimates, the part the language model names at the class size with the second version of the part rules, and the part the language model names and size, which is the full stage. In the centred placement, the sheet is sized to the support's top-surface extent, clamped to the class's range, and released at the support's centre. A scene whose deformable object is a volumetric body has no language-model query, and its part-only and size-only cells copy the full-stage and the centred cell, which the tables mark ``as D'' and ``as A''. For these \ablDevVolN{} development and \ablHoVolN{} held-out scenes, the placement factor is the second placement rules alone and the size factor is nothing, which is why Table~5 of the main paper also counts the sheet scenes by themselves. Table~\ref{tab:supp-2x2} lists every development scene; the full-stage cell on the held-out batch is the per-scene record of Table~\ref{tab:heldout}.

\paragraph{Run-to-run variation.} We run the full-stage cell twice more with the same code over the same assets, so that only the language model's answer and the solver's own nondeterminism differ between runs. The three runs equilibrate \ablDevDSettled{}, \repOneDevSettled{} and \repTwoDevSettled{} development scenes and complete \repDevList{}; on the held-out batch they complete \repHoList{}. Table~3 of the main paper should be read against this spread. A difference of one scene is within the spread of repeated runs of the same row, and the three development scenes that the placement rules add are above it.

\paragraph{Generated asset as the rest shape.} The first row of Table~3 runs the full stage with every cloth-class object simulated as the closed surface of its own generated asset, certified closed the same way as the colliders, with the language model, the placement rules and the geometric checks unchanged, under the option \texttt{SOFT\_ASSET\_SHELL}. The solver rejects the initial configuration on 11 development and 11 held-out scenes, because the two layers of the lifted shell lie closer together than the activation distance $\hat d$: on the development set the apron, the armchair's blanket, the bedsheet, the clothes pile, both tablecloths, the kite, the skirt, the stool's cloth, the throw and the chair's towel, and on the held-out set the bar stool, the bed foot, the bookshelf, the counter, the glass table, the ironing board, the marble table, the piano, the step stool and the stump, where the garden chair also runs out of memory. The remaining scenes equilibrate, \assetDevSettled{} and \assetHoSettled{}, pass the geometric checks on \assetDevGates{} and \assetHoGates{}, and complete \assetDevComplete{} and \assetHoComplete{}, of which \assetDevSheetComplete{} and \assetHoSheetComplete{} are sheet scenes, against \ablDevDSheetComplete{} and \ablHoDSheetComplete{} for the sewn panel. The volumetric bodies are unchanged by the switch. A shell that equilibrates keeps the generator's snapshot of a drape. The vision--language model accepts it where that snapshot fits the support it lands on, a towel over a bathtub's rim or a cloth on a stool, and rejects it where the prompt asks for a part that the snapshot does not reach.

\paragraph{Preserved properties of the generated object.} The sewn template replaces the generated surface, so the question is which properties of the generated object survive. Table~\ref{tab:supp-identity} answers this question with the vision--language model shown the reference image beside the record render and asked, of the deformable object alone, whether its kind and its colour and pattern survive. The kind of object survives in \idCat{} of \idN{} scenes; the exceptions are objects whose construction the material library lacks, the kite, the dumplings and the folded napkin, two runners the model reads as tablecloths, the shirt and the two skirts whose garment the sewn panel does not reproduce, and the folded towels. The colour and pattern survive in \idApp{}; where they do not, the texture extracted from the reference image came out plain white, as on the apron, the denim jacket and the plush bear, or lost its print, as on the daybed quilt. The size relative to the furniture and the drape configuration are not expected to survive, since the template's size comes from the class prior and the language model rather than from the generated surface, and the simulation produces its own folds and overhang. In geometry, the template's largest extent is a median \idExtMed{} of the generated asset's as the layout placed it (interquartile range \idExtQa{} to \idExtQb{} over \idBodies{} bodies). Therefore, the construction keeps what the prompt names and what the reference image colours, and replaces the generator's size and folds by design; the loss of pattern on a third of the scenes is a limitation of the texture extraction rather than of the template.

\begin{table}[tb]
\centering\footnotesize
\begin{tabular*}{\linewidth}{@{\extracolsep{\fill}}l c@{}}
\toprule
criterion & scenes (of 60) \\
\midrule
same kind of object & 48 \\
same colour and pattern & 34 \\
\bottomrule
\end{tabular*}

\caption{\textbf{Properties of the generated object preserved by the sewn template.} GPT-5.4 is shown the reference image and the record render of each of the \idN{} record-run scenes and asked, of the deformable object alone, whether its kind and its colour and pattern survive.}
\label{tab:supp-identity}
\end{table}

\begin{table}[tb]
\centering\scriptsize
\setlength{\tabcolsep}{1.8pt}
\setlength{\tabcolsep}{1.5pt}%
\begin{tabular*}{\linewidth}[t]{@{\extracolsep{\fill}}l cccc c cccc c cccc c cccc@{}}
\toprule
& \multicolumn{4}{c}{A} & & \multicolumn{4}{c}{C} & & \multicolumn{4}{c}{B} & & \multicolumn{4}{c}{D} \\
\cmidrule(lr){2-5}\cmidrule(lr){7-10}\cmidrule(lr){12-15}\cmidrule(lr){17-20}
scene & S & G & J & x & & S & G & J & x & & S & G & J & x & & S & G & J & x \\
\midrule
apron & y & y & n & 0 & & y & y & n & 0 & & y & y & y & 0 & & y & y & y & 0 \\
armchair & y & y & n & 0 & & y & y & y & 0 & & y & y & n & 0 & & y & y & y & 0 \\
bathtub2 & y & y & n & 0 & & y & y & n & 0 & & y & y & y & 0 & & y & y & y & 0 \\
bathtub & y & y & y & 0 & & y & y & n & 0 & & y & y & y & 0 & & y & y & n & 0 \\
bedsheet & y & y & n & 0 & & y & y & n & 0 & & y & y & n & 0 & & y & y & n & 0 \\
bench2b & y & y & y & 0 & & y & y & n & 0 & & y & y & y & 0 & & y & y & y & 0 \\
bench & y & y & y & 0 & & y & y & n & 0 & & y & y & y & 0 & & y & y & y & 0 \\
blanket & y & y & n & 0 & & y & y & n & 0 & & y & y & n & 0 & & y & y & n & 0 \\
box & y & y & n & 0 & & y & y & y & 0 & & y & y & n & 0 & & y & y & y & 0 \\
clothes sofa & y & y & n & 0 & & y & y & n & 0 & & y & y & n & 0 & & y & y & n & 0 \\
coffee & y & y & y & 0 & & y & y & y & 0 & & y & y & y & 0 & & y & y & y & 0 \\
couch & y & y & y & 0 & & \multicolumn{4}{c}{as A} & & \multicolumn{4}{c}{as D} & & y & y & y & 0 \\
cushion & y & y & y & 0 & & \multicolumn{4}{c}{as A} & & \multicolumn{4}{c}{as D} & & y & y & y & 0 \\
daybed & y & y & y & 0 & & y & y & y & 0 & & y & y & y & 0 & & y & y & y & 0 \\
dining2 & y & y & y & 0 & & y & y & y & 0 & & y & y & y & 0 & & y & y & y & 0 \\
dining & y & y & n & 0 & & y & y & n & 0 & & y & y & n & 0 & & y & y & n & 0 \\
kite & y & y & n & 0 & & y & y & n & 0 & & y & y & n & 0 & & y & y & n & 0 \\
ladder & y & n & n & 0 & & y & n & y & 0 & & y & n & n & 0 & & y & n & y & 0 \\
napkin & y & n & n & 0 & & y & n & n & 0 & & y & n & n & 0 & & y & n & n & 0 \\
picnic & y & y & y & 0 & & y & y & y & 0 & & y & y & y & 0 & & y & y & y & 0 \\
pillow2 & y & y & y & 0 & & \multicolumn{4}{c}{as A} & & \multicolumn{4}{c}{as D} & & y & y & y & 0 \\
pillow3 & y & y & y & 0 & & \multicolumn{4}{c}{as A} & & \multicolumn{4}{c}{as D} & & y & y & y & 0 \\
pillow & y & y & n & 0 & & \multicolumn{4}{c}{as A} & & \multicolumn{4}{c}{as D} & & y & y & n & 0 \\
plush & y & y & y & 0 & & \multicolumn{4}{c}{as A} & & \multicolumn{4}{c}{as D} & & y & y & y & 0 \\
rocking & y & y & y & 0 & & y & y & y & 0 & & y & y & y & 0 & & y & y & y & 0 \\
skirt & y & y & y & 0 & & y & y & y & 0 & & y & y & y & 0 & & y & y & y & 0 \\
steamer & y & y & n & 0 & & \multicolumn{4}{c}{as A} & & \multicolumn{4}{c}{as D} & & y & y & n & 0 \\
stool2 & y & y & y & 0 & & y & y & y & 0 & & y & y & y & 0 & & y & y & y & 0 \\
stool & y & y & y & 0 & & y & y & y & 0 & & y & y & y & 0 & & y & y & y & 0 \\
throw & y & y & y & 0 & & y & y & y & 0 & & y & y & y & 0 & & y & y & y & 0 \\
tissue & y & y & n & 0 & & y & n & n & 0 & & y & n & n & 0 & & y & n & n & 0 \\
towel4 & y & y & n & 0 & & y & y & n & 0 & & y & y & n & 0 & & y & y & n & 0 \\
trash & y & y & y & 0 & & y & y & y & 0 & & y & y & y & 0 & & y & y & y & 0 \\
\bottomrule
\end{tabular*}

\caption{\textbf{Per-scene results of the size and placement conditions} under one code version, on the development set. A: centred placement at the class size; C: centred placement at the size the language model estimates; B: placement on the part the language model names, class size; D: that part and size, which is the full stage. Columns as in
Table~\ref{tab:supp-control-scenes}; x is the crossing count.}
\label{tab:supp-2x2}\label{tab:supp-2x2-ho}
\end{table}

\section{Implementation Details}
\label{supp:impl}

In this section, we provide the algorithm, the solver settings, the timings, and the preprocessing rules under which every scene of the paper runs. It begins with the collider's certificate, since every contact in the paper is computed against it. On the towel-scene chair (Table~\ref{tab:cells}) the single convex hull fabricates up to $330$\,mm of geometry across the open back where the towel hangs through the frame, so a contact model built on it produces drapes that rest on fabricated geometry, which a render does not reveal. However, the convex-cell decomposition bounds the fabricated geometry at $29$\,mm and reports it. The analytic field matches finite differences to the order that the release-pose gradient requires.

\begin{table}[tb]
\centering\footnotesize
\setlength{\tabcolsep}{2.5pt}
\begin{tabular}{lcc}
\toprule
collider & mean/max (mm) & closed \\
\midrule
single convex hull      & $82.8$ / $330.5$ & \checkmark \\
\textbf{convex cells ($K{=}35$)} & $\mathbf{8.4}$ / $\mathbf{29.4}$ & \checkmark \\
raw generated mesh      & -- & \boldmath$\times$ \\
\midrule
$\nabla\tilde{\phi}$ vs.\ finite diff. &
\multicolumn{2}{c}{$1.6\times10^{-7}$} \\
Hessian-vector vs.\ finite diff. &
\multicolumn{2}{c}{$1.9\times10^{-6}$} \\
\bottomrule
\end{tabular}
\caption{\textbf{Collider validation} on the towel-scene chair.
Convex cells reduce the fabricated geometry by an order of magnitude
compared with the single convex hull, whose $330$\,mm maximum spans the
chair's open back, where a draped towel hangs through the frame. Both
are closed by construction, which gives the inside/outside test that
the raw mesh cannot, and the analytic contact field matches finite
differences to the order that the release-pose gradient requires.}
\label{tab:cells}
\end{table}

\paragraph{Algorithm.} Fig.~\ref{fig:supp-algo} presents the stage as it runs, together with the section of the main paper from which each step originates. Every scene in the paper follows exactly these steps with the same settings. However, the demonstration runs, such as the teaser of Sec.~\ref{supp:rooms}, differ only in the placement options named there.

\begin{figure}[tb]
\small
\begin{tabular}{@{}p{0.97\linewidth}@{}}
\toprule
\textbf{Input:} the prompt and the generated scene, that is, the
reference image, masks, lifted assets and language model layout.\quad
\textbf{Output:} an equilibrated
simulation state, or a rejection with its cause. \\
\midrule
1. \emph{Route} every object: rigid support, shell, volumetric body or
rod, by the keyword material library; take its material from the material library
(Sec.~3.2). \\
2. \emph{Supports:} de-occlude the support's crop and regenerate the
asset, align it to the layout, close it (open-edge test, convex
cells), and write its certificate (Sec.~3.4). \\
3. \emph{Rest shapes:} cut a sheet to the outline its class names, or
sew the library garment pattern, at the size the language model estimates or the class
prior; stuff the bodies that are stuffed (Sec.~3.2). \\
4. \emph{Release:} resolve the part the prompt names on the collider,
place the rest shape above it, lifted clear of the bodies already
placed; check that step zero is intersection-free (Sec.~3.4). \\
5. \emph{Equilibrate:} minimize the incremental potential with the
per-object energies for up to $300$ steps of $0.02$\,s; stop once the
largest displacement over $40$ steps is under $0.5$\,mm (Sec.~3.3). \\
6. \emph{Checks:} crossings, support, floor and asset tests on the
equilibrated geometry; then the prompt check on the render (Sec.~3.5). \\
7. \emph{Correction:} if the prompt check fails and the prompt names a
reachable part, move the release by the adjoint of a differentiable
simulation or by the pipeline grid, re-solve, and re-run the geometric checks and
the prompt check (Sec.~3.6). \\
\bottomrule
\end{tabular}
\caption{\textbf{The equilibration stage.} One pass over a generated scene.}
\label{fig:supp-algo}
\end{figure}

\paragraph{Parameters.} The solver settings remain the same for every scene. We use a time step of $0.02$\,s; at most $300$ implicit steps, with the stop test of Fig.~\ref{fig:supp-algo} applied after step $200$; a contact distance $\hat d = 1$\,mm, since the library's default of $10$\,mm keeps every cloth a centimetre off its support; a contact resistance of $1$\,GPa; Coulomb friction of $0.6$ between every pair, from the material library's cloth entries; Newton termination at a velocity residual of $0.02$\,m/s, i.e.\ $\lVert\Delta\bm{x}\rVert_\infty/\Delta t<0.02$; gravity $9.8$\,m/s$^{2}$; sheets of $128\times128$ vertices; a $3$\,mm standoff between rigid bodies at release; and soft position constraints at a strength ratio of $100$ for the manipulation episodes, gripping the vertices within $6$\,cm of the chosen point, or $8$\,cm for press. Friction follows the lagged potential of IPC, $D(\bm{x},\bm{x}_n)=\sum_k \mu\,\lambda_k\, f_0(\lVert\bm{u}_k\rVert)$ with $\lambda_k$ the previous step's normal force at contact pair $k$, $\bm{u}_k$ its tangential slip and $f_0$ the smoothed absolute value with velocity threshold $\epsilon_v=10^{-3}$\,m/s. Three quantities are chosen per scene by the language model and never manually: the sheet's size, from the class prior or the language model's estimate (Sec.~\ref{supp:2x2}), the support part that the placement aims at, and the material class. However, the only parameters set by hand in the paper are the placement options of the demonstration runs (Sec.~\ref{supp:rooms}), which is why those runs enter no count.

\paragraph{Materials.} Materials are selected from the material library according to the object's name. The shells are listed by Young's modulus, Poisson ratio, thickness, density and bending stiffness as follows: cotton $10^{6}$\,Pa, $0.30$, $0.4$\,mm, $375$\,kg/m$^{3}$ ($150\,\mathrm{g/m^2}$), $1.0$; fleece $5\times10^{5}$, $0.30$, $1.2$\,mm, $250$, $0.5$; silk $5\times10^{5}$, $0.35$, $0.2$\,mm, $350$, $0.2$; and wool $8\times10^{5}$, $0.30$, $1.5$\,mm, $280$, $2.0$. The material library also includes leather, with $5\times E$ and $2.5\times h$ of cotton, denim, with $2\times E$, and paper, with $E{=}10^5$\,Pa, $k_b{=}20$ and $h{=}0.1$\,mm. The membrane values correspond to real fabric, and $\mu_f{=}0.6$ for every entry. Volumetric bodies use stable neo-Hookean materials, with Young's modulus, Poisson ratio and density of $8\times10^{3}$\,Pa, $0.30$ and $100$\,kg/m$^{3}$ for foam, $5\times10^{3}$, $0.35$ and $80$ for plush, and $3\times10^{4}$, $0.35$ and $120$ for mattress; dough has $E{=}1.5{\times}10^{4}$\,Pa and $\nu{=}0.45$, which is near-incompressible. Finally, rods are rope, cable, cord, string or hose (Table~\ref{tab:supp-rods}).

\paragraph{Optimization settings.} For placement optimization, we run Adam for $8$ iterations on $\theta=(c_x, c_z)$ with step size $\eta=\min(\max(0.04,\,1.2\,d/8),\,0.15)$ for a target $d$ metres away. We clip the gradient as $\bm{g}\leftarrow\bm{g}\min(1,1/\lVert\bm{g}\rVert)$ and keep the best iterate. The pipeline grid comprises three placements toward the target, at the language model's and at the class size.

\paragraph{Timing and memory.} We report wall-clock medians over the scheduler's records of every completed job of each kind, with one GPU per job. Generating a scene, which covers the reference image, the segmentation, one 3D generation per object and the layout, takes $13$\,min on an A6000. The stage end to end on an A5000, from de-occlusion through texture, equilibration, the geometric checks and the collider render, takes $18$\,min over $n{=}13$ held-out chains. A single-object equilibration run alone takes \ctrlFullTime{}, the median over the control runs; a studio render takes $5.7$\,min, a manipulation episode \ixSecPerStep\,s per step, and the prompt check under a minute. Every single-object run and every multi-body scene of up to eight bodies fits on a $24$\,GB A5000.

\paragraph{Classification.} Rigid compound names, such as rack, box, bin, basket, hamper, crate, holder, stand, shelf, frame, pot and steamer, are matched first. This ordering ensures that a \emph{tissue box} is a container rather than cloth. Shell keywords are matched next, followed by volumetric-body keywords. Catalog names can embed the scene identifier, as in \texttt{steamer\_001\_dumpling1}, and we strip it before matching; otherwise the dumplings classify as steamers.

\paragraph{Sizes and placements.} A sheet takes the size specified by the placement language model. Where the language model specifies no size, we set the sheet size from the support top-surface extent with a $10$\,cm overhang, $w=w_{\mathrm{s}}+0.2$\,m, $\ell=\ell_{\mathrm{s}}+0.2$\,m. Sheets are released just above the support top, and the $k$-th shell of a stack is released at $y_k=y_{\mathrm{top}}+h_{\mathrm{drop}}+0.06\,k$\,m. A sheet planned \emph{inside} a container-class support shrinks to $55\%$ of the container's own opening, $w=0.55\,w_{\mathrm{open}}$, measured from the container's bounding box rather than from the scene union, because with a table in the scene the union bounding box put tissue balls above the table; a sheet planned on top of the container keeps its size. Backrest-dominated supports release the sheet over the seat band, the largest upward-facing area band, with its size set to the top-surface extent plus a $0.3$\,m hang. Clotheslines and flags use a hanging placement, a vertical sheet whose top row, or pole-side column for flags, is pinned, so that hanging is imposed as a boundary condition. Crumpled paper is a perturbed icosphere at the paper entry; the crumple is placed in the rest shape because the solver has no plasticity.

\paragraph{Panel outlines.} Single-panel objects are cut on the simulation grid to an outline before simulation: a rounded rectangle for linens, towels and blankets, an ellipse for round cloths, a diamond for the kite, and a bib-and-skirt outline for the apron. Faces with any removed vertex are removed as well, and the grid is reindexed. Every named garment in the paper, the skirt, the shirt, the jacket, the sweater, the hoodie, the trousers and the dress, is instead sewn from the library pattern of its class (Sec.~3.2 of the main paper).

\paragraph{Mesh conditioning.} Decimation to $\le$12k faces \emph{precedes} repair, because repair upsamples and decimation after repair re-introduces the intersections that repair removed. Repair is unconditional.

\paragraph{Numerical sensitivity.} Table~\ref{tab:supp-dtsweep} re-solves six development scenes at steps of $0.005$, $0.01$ and $0.04$\,s with the same release and with the simulated time held at $6$\,s, and at sheet resolutions $64$ and $192$. Every run reaches static equilibrium with zero crossings and zero penetration and passes the containment and support checks. The contact fraction is unchanged to within four points on five of the six scenes. However, the two small napkins at resolution $192$ produced a defective sheet mesh and are not scored.

\begin{table}[tb]
\centering\scriptsize
\setlength{\tabcolsep}{3pt}
\begin{tabular*}{\linewidth}{@{\extracolsep{\fill}}cccccc@{}}
\toprule
step (s) & $n$ & settled & cross. & pen.\ (mm) & contact (\%) \\
\midrule
$0.005$ & 128 & 6/6 & 0 & 0.0 & 5--100 \\
$0.01$ & 128 & 6/6 & 0 & 0.0 & 9--100 \\
$0.02$ & 128 & 6/6 & 0 & 0.0 & 8--100 \\
$0.04$ & 128 & 6/6 & 0 & 0.0 & 4--100 \\
$0.02$ & 64 & 6/6 & 0 & 0.0 & 2--100 \\
$0.02$ & 192 & 4/4 & 0 & 0.0 & 2--90 \\
\bottomrule
\end{tabular*}

\caption{\textbf{Numerical sensitivity.} Six development scenes re-solved at other time steps (simulated time fixed at $6$\,s) and sheet resolutions. Settled: scenes reaching static equilibrium; crossings and penetration are summed and maximised over the scenes; contact: range of the contact fraction.}
\label{tab:supp-dtsweep}
\end{table}

\paragraph{Cloth textures.} The cloth texture is initialized from the cloth region of the reference image. The largest axis-aligned rectangle inside the eroded mask is cropped, padded to a square, and passed to the image editor, gpt-image-1.5, with the instruction to return a seamless, flat, evenly lit texture of that material only, while ignoring anything in front of or behind it. The result is then downsampled to $512^2$. Its residual low-frequency luminance is divided out, and its seams are cross-faded after a half-period shift so that the tile wraps. Bump presets for terry, knit, crinkle and weave are chosen from prompt and name keywords, and the height signal multiplies the albedo through a colour ramp, because normals alone are not visible on a flat-lit sheet.

\section{Multi-body Scenes and Rods}
\label{supp:rooms}

In this section, we record the multi-body scenes of Fig.~3 of the main paper. We record the rods, which are an exploratory extension, and the teaser scene, which is a demonstration run.

\paragraph{Scenes.} Fig.~3 of the main paper presents scenes that contain several deformable objects on one support in a single simulation under one contact barrier. Table~\ref{tab:supp-rooms} lists all such scenes, together with the body count produced by the routing stage and the wall time of the whole job, including meshing, placement and simulation, on one GPU.

Every body is placed according to the rules of the single-object scenes, with two additions. A body released over another deformable object is lifted until its nearest point is $15$\,mm from every surface already placed, in both directions, and a body that a support crossing has lifted afterwards is checked against the deformable objects again, so that two throws released on the same backrest ridge keep more than their activation distance $\hat d$ between them. Two further rules apply to volumetric bodies: a toy is lowered onto its layer like any other body rather than left where it was placed over a ridge, and a volumetric body is kept inside the support's ground projection by its own half extent at release.

\begin{table*}[t]
  \centering\footnotesize
  \setlength{\tabcolsep}{3.5pt}
  \begin{tabular}{l>{\raggedright\arraybackslash}p{5.0cm}ccc>{\raggedright\arraybackslash}p{5.6cm}}
    \toprule
    scene & deformable objects & shells & volumetric bodies & wall & result \\
    \midrule
    messy bed & quilt, pillows, plush bear & 2 & 2 & 8.9 & equilibrated, passes \\
    made bed & sheet and three pillows & 1 & 3 & 8.4 & equilibrated, passes \\
    wardrobe & shirt and trousers on a bed & 2 & 0 & 8.0 & equilibrated, passes$^\dagger$ \\
    dinner & tablecloth and napkin on a chair & 2 & 0 & 8.6 & equilibrated, passes$^\dagger$ \\
    messy bed 2 & quilt, two pillows, five toys & 1 & 7 & 11.5 & equilibrated, passes \\
    reading nook & blanket, two pillows, plush rabbit, charging cable & 1 & 3 & 6.9 & equilibrated, passes (one rod) \\
    picnic & gingham cloth, coil of rope, plush dog & 1 & 1 & 6.7 & equilibrated, passes (one rod); the beach towel was omitted upstream \\
    set table & tablecloth, four napkins, telephone cord & 5 & 0 & 27.6 & equilibrated, passes (one rod) \\
    child's bed & patchwork quilt, hoodie, two pillows, teddy bear & 2 & 3 & 13.2 & equilibrated, passes$^\dagger$ \\
    closet & six garments & 6 & 0 & 125 & equilibrated, passes the geometric checks$^\ddagger$ \\
    \bottomrule
  \end{tabular}
  \caption{\textbf{Multi-body scenes.} Shells and volumetric bodies are
deformable objects, and every scene also has one rigid support. Wall
time is in minutes for the whole job on one A5000 or A6000.
$^\dagger$The smaller shell of the pair is a membrane-only shell.
$^\ddagger$Solved on a 96\,GB card with the library garments
decimated to about 5{,}200 faces each, four of the six as membrane-only
shells. It is not shown, because its wardrobe reached the solver as
three loose shelves, which is an asset defect.}
  \label{tab:supp-rooms}
\end{table*}

\begin{figure*}[t]
  \centering
  \setlength{\tabcolsep}{0pt}%
\noindent\begin{tabular}{@{}c@{\hspace{0.03in}}c@{\hspace{0.03in}}c@{\hspace{0.03in}}c@{\hspace{0.03in}}c@{}}
\includegraphics[width=1.351in]{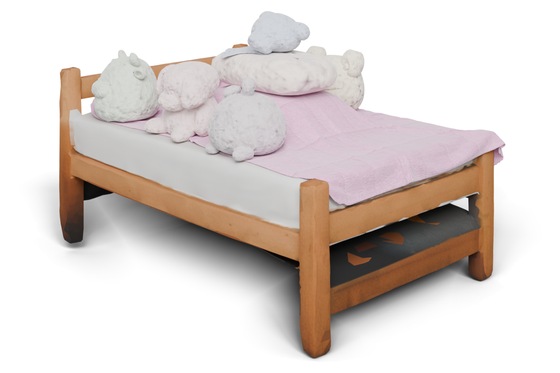} & \includegraphics[width=1.351in]{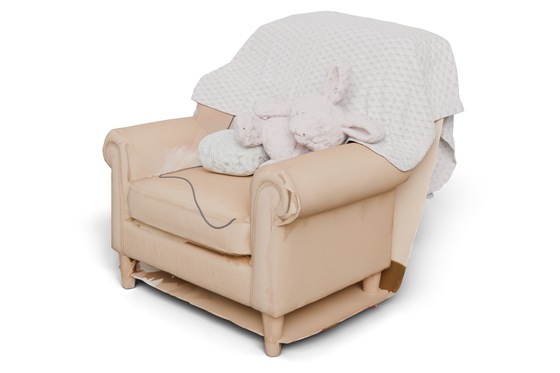} & \includegraphics[width=1.351in]{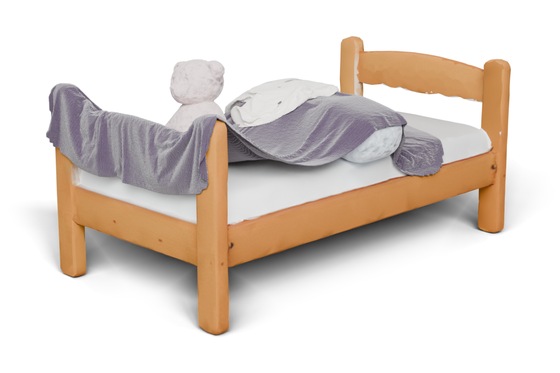} & \includegraphics[width=1.351in]{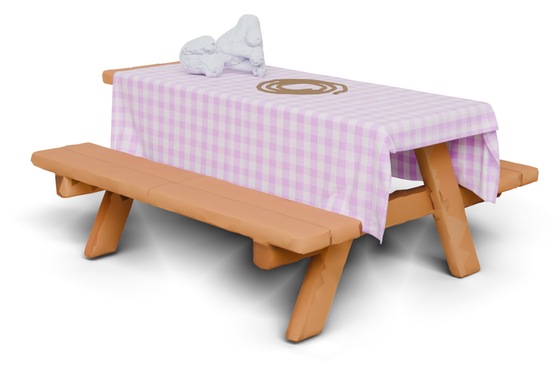} & \includegraphics[width=1.351in]{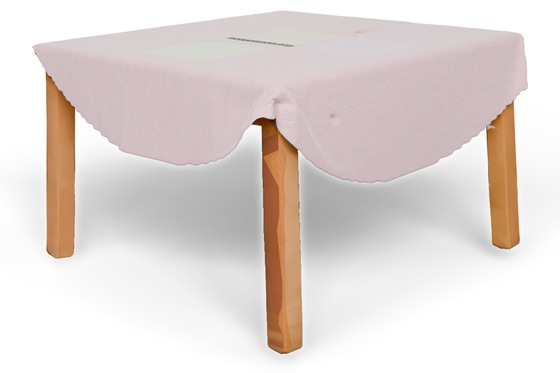} \\[1pt]
\footnotesize messy bed & \footnotesize reading nook & \footnotesize child's bed & \footnotesize picnic & \footnotesize set table
\end{tabular}

  \caption{\textbf{Multi-body scenes}, each the last simulation recorded in Table~\ref{tab:supp-rooms}: two to eight deformable objects, quilts, pillows, garments, stuffed toys and a cord, equilibrated in one simulation.}
\label{fig:supp-rooms-all}
\end{figure*}

\begin{figure*}[t]
  \centering
\setlength{\tabcolsep}{2pt}%
\providecommand{\srodw}{1.33in}
\providecommand{\snm}[1]{\figcap{#1}}
\noindent\begin{tabular}{@{}ccccc@{}}
\includegraphics[width=\srodw]{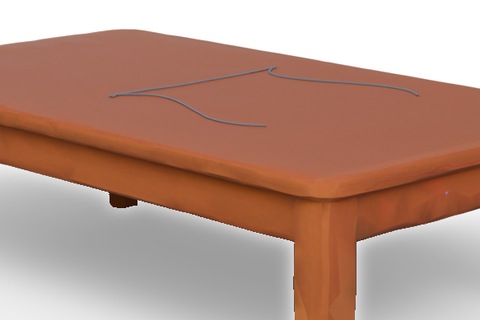} &
\includegraphics[width=\srodw]{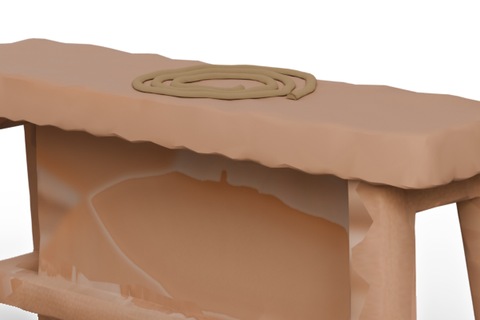} &
\includegraphics[width=\srodw]{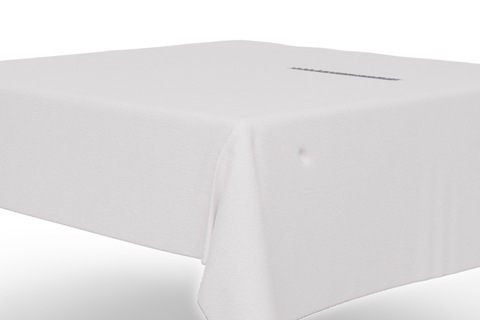} &
\includegraphics[width=\srodw]{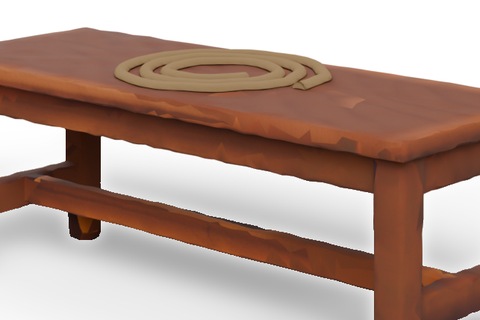} &
\includegraphics[width=\srodw]{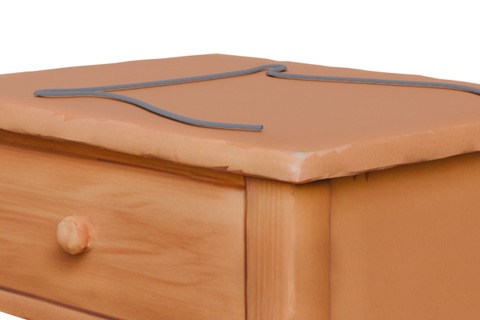} \\[1pt]
\snm{charging cable} & \snm{rope} & \snm{telephone cord on a tablecloth} & \snm{``a rope on a bench''} & \snm{``a cable on a side table''}
\end{tabular}

  \caption{\textbf{Rods.} Three rods laid on supports of the scene
  set, a cable, a rope and a telephone cord on an equilibrated tablecloth,
  and the two scenes whose prompt named the rod, equilibrated in the same
  simulation as the shells.}
  \label{fig:supp-rods}
\end{figure*}

\begin{figure*}[t]
  \centering
  \input{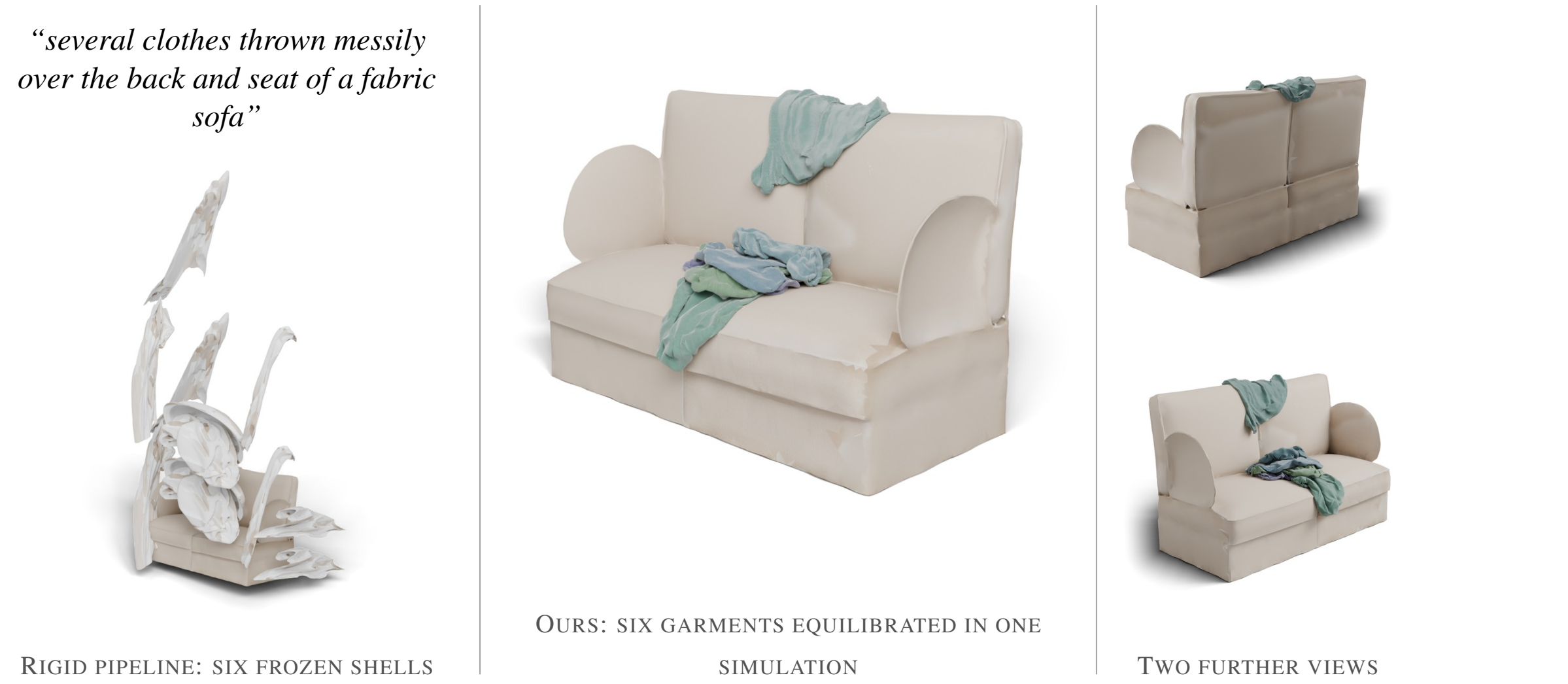}
  \caption{\textbf{Six garments on a sofa.} The prompt asks for several
  clothes thrown messily over a sofa. Left: the rigid pipeline lifts
  each garment to a frozen shell, and its pile intersects the backrest
  and hangs in the air. Middle and right: our equilibration stage brings
  the six garments to rest on the seat and back in one simulation, with
  none of their $12{,}089$ cloth edges crossing the sofa, seen from the
  generation viewpoint and from two novel views.}
  \label{fig:supp-pile}
\end{figure*}

\paragraph{Rods.} The three rods of Fig.~\ref{fig:supp-rods} are introduced using the prompts ``a charging cable on the coffee table'', ``a coil of rope on the bench'' and ``a telephone cord on the table'' on supports from the main set. They are placed on the largest upward-facing band of the support, above any deformable object already present there. Rods enter no count in the main paper.

A rod is a polyline with a section radius, discretized at $2.5$ radii or $8$\,mm, whichever is larger, with a Hookean stretch term and the Kirchhoff bending term of the main paper's Sec.~3.3 on a straight rest shape. The loop, coil and helix it starts from are initial conditions that the simulation relaxes rather than rest shapes. Table~\ref{tab:supp-rods} gives the material table and the scenes. For contact, we use a thin offset of $0.3$\,mm rather than the section radius, because the solver's initial distance check applies the largest thickness in the scene to every pair, and with a $3$\,mm rod the check reported the table as too close to itself. The tube drawn in the figure is the centreline swept by the section radius, so the picture can overlap the support where the centreline does not.

All three rods equilibrate within the step budget under the early-stop rule of every scene and pass the containment and support checks. Measured against their closed supports, every centreline stays outside its support; the cable's and the rope's tubes enter by less than their section radius, and the cord's not at all. Two more scenes, ``a coil of rope on a wooden bench'' and ``a charging cable on a nightstand'', name the rod as the prompt's main object. The generator produces a rope or cable asset, the routing stage classifies it as a rod from its name and replaces it by a centreline of the matching curve, and the simulation proceeds as above. Because the layout stage sizes the rope to the bench, $1.2$\,m, and the cable to a metre, the extent of a generated rod is capped by its curve, $0.45$\,m for a coil and $0.6$\,m for a loop or helix, and by $0.85$ of the support's smaller extent. Both scenes equilibrate and pass (Fig.~\ref{fig:supp-rods}, last two panels).

\begin{table*}[t]
  \centering\footnotesize
  \setlength{\tabcolsep}{3.5pt}
  \begin{minipage}[t]{0.38\textwidth}\centering
  \begin{tabular}[t]{lccccc}
    \toprule
    material & $E$ (Pa) & $k_{\text{bend}}$ & $r$ (mm) & $\rho$ (kg/m$^3$) & $\mu$ \\
    \midrule
    rope   & $2\times10^{7}$ & $5\times10^{5}$ & 12  & 900  & 0.7 \\
    cable  & $1\times10^{8}$ & $5\times10^{6}$ & 2.5 & 1200 & 0.5 \\
    cord   & $1\times10^{8}$ & $2\times10^{6}$ & 3   & 1100 & 0.5 \\
    string & $5\times10^{7}$ & $2\times10^{5}$ & 1.5 & 900  & 0.5 \\
    hose   & $5\times10^{7}$ & $3\times10^{6}$ & 8   & 1000 & 0.6 \\
    \bottomrule
  \end{tabular}
  \end{minipage}\hfill
  \begin{minipage}[t]{0.59\textwidth}\centering
  \begin{tabular}[t]{llcrcc}
    \toprule
    scene & rod & start & vertices & extent & wall (min) \\
    \midrule
    coffee table & charging cable & loop  & 166 & 0.50\,m & 4.2 \\
    bench        & rope           & coil  & 86  & 0.42\,m & 2.9 \\
    dining table & telephone cord & helix & 857 & 0.60\,m & 7.6 \\
    bench (prompt object)      & rope           & coil & 78 & 0.38\,m & 2.8 \\
    nightstand (prompt object) & charging cable & loop & 81 & 0.24\,m & 3.1 \\
    \bottomrule
  \end{tabular}
  \end{minipage}
  \caption{\textbf{Rods.} Left: the material table, chosen from the
  rod's name. $k_{\text{bend}}$ is the stiffness passed to the
  Kirchhoff term, $\alpha=k\pi r^{4}/4$. Right: the five rod scenes of
  Fig.~\ref{fig:supp-rods}, with the curve each rod was released as and
  the wall time in minutes. The last two are scenes whose prompt named
  the rod as the main object. The telephone cord lies on the dining
  table's tablecloth, a shell equilibrated in the same simulation.}
  \label{tab:supp-rods}
\end{table*}

\paragraph{Teaser scene.} Fig.~1 of the main paper shows a fabric sofa with twelve shirts, which is a demonstration run that enters no count in the paper. The rigid pipeline freezes the pieces at the locations where the layout lifts them, as a stack above the sofa with every piece floating. Each garment is sewn from the library T-shirt pattern at $1.3\times$ its catalogue size, $0.94$\,m across the shoulders. Because the prompt names both the back and the seat, we release the twelve garments alternately over the backrest ridge and the seat with a small random yaw and tilt. Each garment is lifted clear of the bodies already placed, with the last one by up to $4.5$\,m, and equilibrated in one simulation. The seat band alone, which the placement would choose for a single garment, cannot hold twelve $0.94$\,m shirts. The simulation passes both geometric checks, with no garment edge crossing the sofa and $26\%$ of the cloth vertices in contact. However, the run shown in Fig.~1, re-solved with the state written out every ten steps for the equilibration strip, reaches static equilibrium at step $200$ and passes both geometric checks.

\section{Placement Optimization}
\label{supp:relcorr}

In this section, we provide the per-scene record of the placement optimization of Sec.~3.6 of the main paper, and the derivation of the adjoint it uses.

\paragraph{Single body at equal compute.} On the \budN{} correction scenes whose prompt names a reachable part, the target lies on the line from the placement toward the named part, the loss is a function of two placement coordinates whose minimum the six grid placements bracket, and the grid of Eq.~9 of the main paper passes \budGrid{} scenes in six re-simulations with the pipeline's solver; the three it does not pass are the hamper towel and the marble-table cloth, whose complaint is the fold rather than the placement, and the chair-back towel, whose equilibrated centroid is at the target already. The adjoint enters where the target does not lie on the grid's line. Table~\ref{tab:supp-budget} collects the single-body rows of the obstructed benchmark of Sec.~\ref{supp:multi}, in which an obstruction lies between the placement and the target, so that no point on the grid's line reaches it, whereas the gradient carries the body across: the adjoint lands the body in $4$ of $8$ runs on the four original supports and in $24$ of $32$ on eight, against $2$ and $20$ for the grid at twice the forward budget. The pipeline therefore uses the grid where the target lies on its line and the adjoint where it does not, for several bodies (Sec.~\ref{supp:multi}), and for the material recovery of Sec.~\ref{supp:material}, which has no grid.

\begin{table}[tb]
\centering\scriptsize
\setlength{\tabcolsep}{3pt}
\begin{tabular*}{\linewidth}{@{\extracolsep{\fill}}l r cc@{}}
\toprule
single body, target past an obstruction & runs & grid & adjoint \\
\midrule
four supports, two seeds (Table~\ref{tab:supp-multi-grid}) & 8 & 2 & \textbf{4} \\
eight supports, four seeds (Table~\ref{tab:supp-multi-grid}) & 32 & 20 & \textbf{24} \\
\bottomrule
\end{tabular*}

\caption{\textbf{Single-body placement past an obstruction.} Runs in which the body lands on its target, for the grid at twice the adjoint's forward budget and for the adjoint. No point on the grid's line reaches a target behind an obstruction, whereas the gradient carries the body across it.}
\label{tab:supp-budget}
\end{table}

\paragraph{Adjoint derivation.} The surrogate simulation comprises a sequence of implicit steps. With $\bm{x}_n$ denoting the positions after step $n$, each step solves the stationarity condition of Eq.~(1) of the main paper by Newton's method,
\begin{align}
\bm{R}_n(\bm{x}_{n+1}, \bm{x}_n, \bm{x}_{n-1}, \theta) ={}&
\tfrac{1}{\Delta t^{2}}\mathbf{M}(\bm{x}_{n+1} - 2\bm{x}_n + \bm{x}_{n-1}) \nonumber\\
&- \bm{f}_{\mathrm{ext}} + \nabla\Psi(\bm{x}_{n+1};\theta) \nonumber\\
&+ \nabla B(\bm{x}_{n+1}) + \nabla D(\bm{x}_{n+1}, \bm{x}_n) \nonumber\\
&= \bm{0},
\label{eq:supp-residual}
\end{align}
with $D$ the lagged friction potential, which depends on the previous positions. The placement $\theta$ enters through the initial positions
$\bm{x}_0 = \bm{x}_{-1}$ (and, for a joint correction, through the rest
shape and the material in $\Psi$). For a loss $L(\bm{x}_N)$ on the equilibrium state, the gradient
follows from the Lagrangian $L + \sum_n \bm{w}_n^{\!\top}\bm{R}_n$.
Stationarity in each $\bm{x}_k$ gives a reverse sweep of linear
solves with the step's own Newton matrix $\mathbf{H}_n =
\partial\bm{R}_n/\partial\bm{x}_{n+1}$,
\begin{align}
\mathbf{H}_{k-1}^{\!\top}\bm{w}_{k-1} ={}& -\Big[\frac{\partial L}{\partial\bm{x}_k}
+ \Big(\frac{\partial\bm{R}_k}{\partial\bm{x}_k}\Big)^{\!\top}\bm{w}_k \nonumber\\
&\quad + \Big(\frac{\partial\bm{R}_{k+1}}{\partial\bm{x}_k}\Big)^{\!\top}\bm{w}_{k+1}\Big],
\quad k = N, \ldots, 1,
\label{eq:supp-adjoint}
\end{align}
with $\bm{w}_N = \bm{w}_{N+1} = \bm{0}$, and then $\mathrm{d}L/\mathrm{d}\theta = \sum_n \bm{w}_n^{\!\top}\, \partial\bm{R}_n/\partial\theta$. The backward pass requires one linear solve per executed step with a matrix that the forward pass already factorized, so its cost matches the forward pass and its state is linear in vertices and steps. Positions are stored at segment checkpoints, and the steps between them are recomputed. The barrier's continuous-collision line search is not differentiated.

At convergence, the step satisfies Eq.~\eqref{eq:supp-residual} exactly. Therefore, the adjoint is exact wherever the set of active contact pairs does not change with $\theta$. This is the compact-contact regime, in which it matches finite differences on the single-body scenes. The sweep is the same as in the block-descent and GPU-IPC adjoints that the main paper builds on, applied to the Newton-solved steps of the surrogate.

\section{Manipulation and Editing}
\label{supp:manip}

In this section, we record every manipulation episode of Sec.~4.4 of the main paper, as well as the tidying and editing sequences on the second messy bed and the fold test.

\paragraph{Episodes.} Table~\ref{tab:supp-ixbench} lists every episode. Each episode starts from the equilibrium state of its scene, grips the vertices within a radius of $6$\,cm, or $8$\,cm for press, of a chosen point with a soft position constraint, moves the target along the prescribed path over $60$ steps of $0.02$\,s, holds for $15$ steps, releases, and runs for $90$ more steps. Pull moves the corner farthest from the body's centre $0.8$\,m outward and $0.45$\,m up. Lift raises the top centre by $0.5$\,m. Drag moves the same patch $0.4$\,m along the support at constant height. Press drives the crown of a volumetric body down by half its height. The constraint is soft, so a heavy, stiff body such as the daybed quilt follows the grip only where a free corner is available; on the daybed, where the quilt lies flush to the back and hangs over the front rail, a $6$\,cm patch at the free corner reaches $99\%$ of its target. Finally, all \ixEpisodes{} episodes finish with finite state at \ixSecPerStep\,s per step on one A5000.

\paragraph{Press.} The indentation denotes the vertical travel of the gripped patch, and recovery denotes the fraction of that travel that the patch recovers after release. The indentations are $8.1$, $7.4$, $7.3$ and $7.5$\,cm of about $10$\,cm commanded for the couch cushion, the loose cushion and the two pillows, and each recovers to within a centimetre of its rest height. The plush toy, which is $48$\,cm tall and pressed by $23$\,cm, topples and comes to rest on its side, which is the expected outcome of that command.

\paragraph{Tidying.} The tidying episode of the manipulation figure in the main paper comprises five grips in one simulation of the second messy bed. The toys are ordered along the long axis of the bed. Each toy is gripped on the top fifth of its vertices within $7$\,cm of the crown, raised $35$\,cm over $90$ steps along three eased segments, up, across to a point $45$\,cm beyond the footboard, and down to the floor, held for $8$ steps and released, and the scene runs $30$ steps before the next grip and $60$ after the last. The grip error is $3.6$\,mm on average over the $700$ steps, and it rises to $58$\,mm for a few steps while the second and fourth toys are carried over the footboard. Every toy rests on the floor afterwards, within $2$\,cm of its target in the plane, and the same episode on the child's bed, with one bear and one grip, behaves in the same way with a $3$\,mm mean.

\begin{figure*}[t]
  \centering
\setlength{\tabcolsep}{2pt}%
\providecommand{\edc}[1]{\parbox[t]{0.235\textwidth}{\centering\figcap{#1}}}
\begin{tabular}{@{}cccc@{}}
\includegraphics[width=0.235\textwidth]{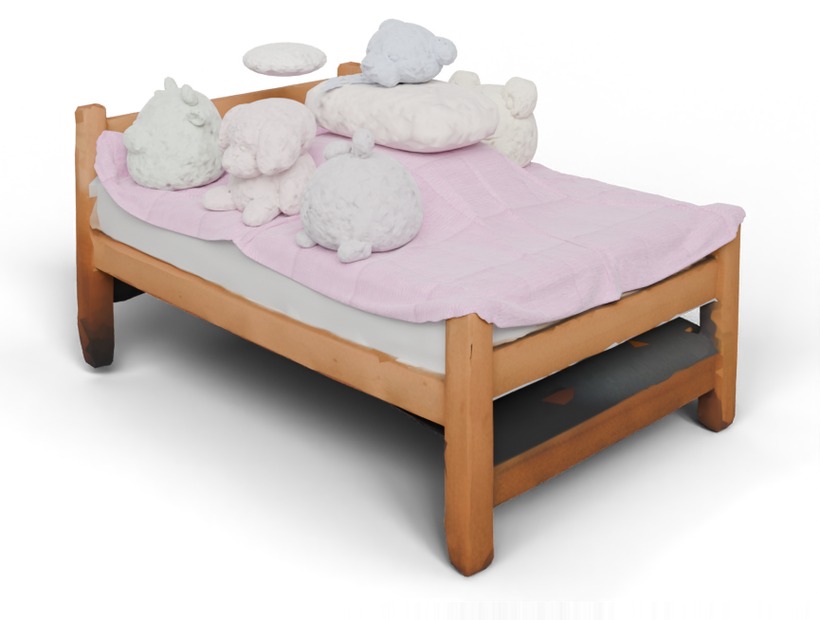} &
\includegraphics[width=0.235\textwidth]{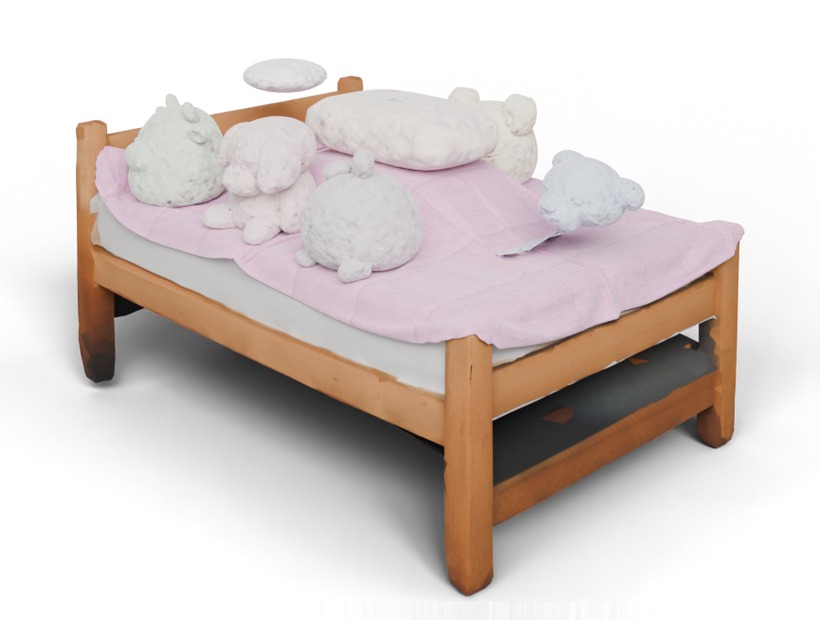} &
\includegraphics[width=0.235\textwidth]{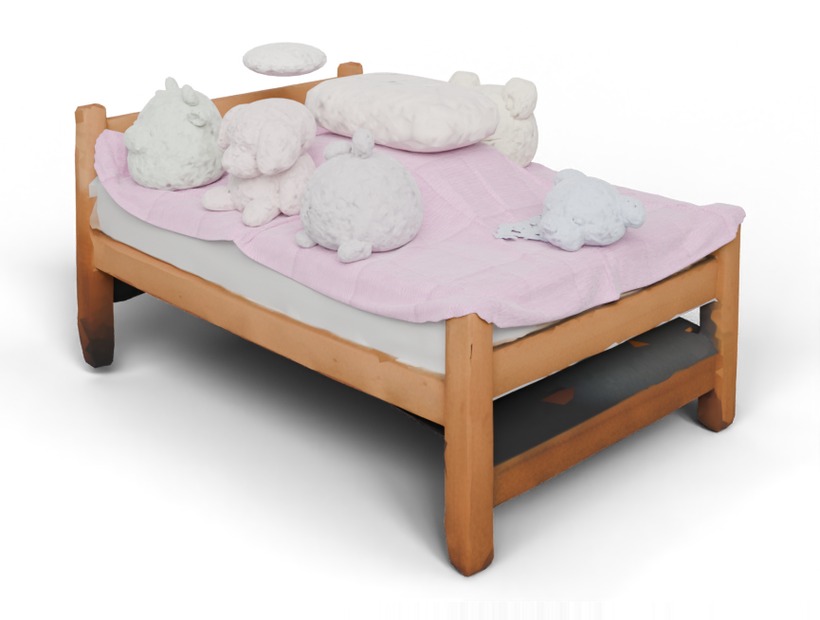} &
\includegraphics[width=0.235\textwidth]{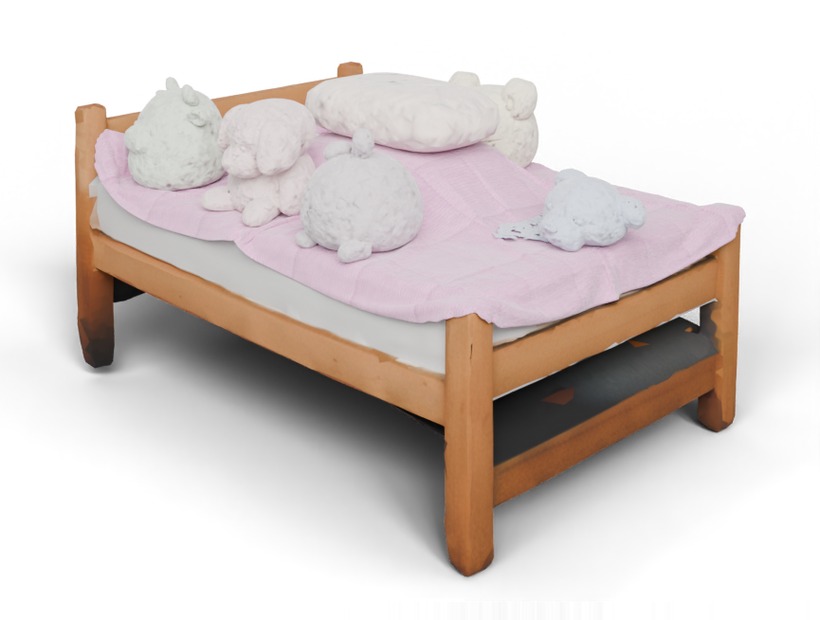} \\
\edc{(a) the equilibrated bed; the pillow to add is held at its corrected release} &
\edc{(b) the toy carried from the pillows to the foot of the bed} &
\edc{(c) the toy set down at the foot of the bed; the scene re-equilibrates} &
\edc{(d) the held pillow released at the head of the bed}
\end{tabular}

  \caption{\textbf{Scene editing.} A move and an addition on the equilibrated
  second messy bed in one simulation, with every body simulated throughout
  (Sec.~4.4 of the main paper): the plush toy on the pillows is gripped,
  carried to the foot of the bed and set down on the quilt, and a foam
  pillow introduced by the prompt is held while the scene equilibrates at
  a release that the differentiable surrogate corrected toward the head
  of the bed, and then released. It comes to rest $1.2$\,cm from the
  target, compared with $67$\,cm from the language model's placement.}
  \label{fig:supp-edit}
\end{figure*}

\paragraph{Scene editing.} Fig.~\ref{fig:supp-edit} and Sec.~4.4 of the main paper edit an equilibrated scene with the same constraint, and every body remains simulated throughout. In one simulation of the second messy bed, a foam pillow introduced by the prompt is held at its placement by its own soft position constraint while the quilt, the two pillows and the five toys equilibrate for the usual $300$ steps. Then, the plush toy on top of the pillows is gripped on its crown, carried to the foot of the bed and set down on the quilt, with $90$ steps of travel, $8$ held and $30$ free, which is a move; a deletion carries the body to the floor beside the support instead. The held pillow is then released and runs until it is at rest, $61$ steps, which is an addition, followed by $60$ free steps. The carried toy comes to rest $14$\,mm from its target in the plane with a grip error of $0.9$\,mm mean and $7.4$\,mm peak, the pillows re-equilibrate where the toy was, and the $249$ steps after the equilibration phase take $119$\,s on one A5000 with finite state throughout.

The addition is where the differentiable simulation enters, as PAT3D's layout optimization does for a rigid layout. The language model's placement for the new pillow is the centre of the seat band, so the pillow lands in the middle of the quilt, $0.67$\,m from where a pillow belongs, the head of the bed, the end whose support rises highest. The placement is optimized by the surrogate of Sec.~\ref{supp:relcorr}, with the equilibrated quilt, pillows, toys and the bed as static colliders and the stuffed pillow approximated by its closed shell with thickness $5$\,mm, $E{=}10^{5}$\,Pa, bending $10$ and density matched to the foam. Adam on the placement offset for $8$ iterations, with the squared distance of the equilibrium centroid to the head target as the loss, takes the loss from $0.446$ to $3\times10^{-4}$ in $395$\,s and moves the placement by $0.66$\,m. Passed back to the pipeline simulation through its placement override, the pillow comes to rest $1.4$\,cm from the target, against $67$\,cm before, and the surrogate's predicted landing point is within $1$\,cm of the pipeline's, the same transfer as for the sheets. This is the deletion and addition of PAT3D's scene-editing demonstration carried out on deformable objects, with the placement of the added body optimized through the equilibration simulation rather than set manually.

\begin{figure*}[!t]
  \centering
  \setlength{\tabcolsep}{0pt}%
\noindent\begin{tabular}{@{}c@{\hspace{0.03in}}c@{\hspace{0.03in}}c@{\hspace{0.03in}}c@{\hspace{0.03in}}c@{\hspace{0.03in}}c@{}}
\includegraphics[width=1.121in]{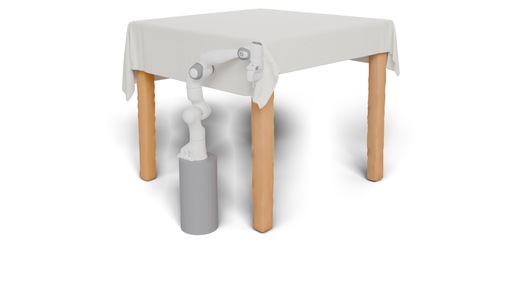} & \includegraphics[width=1.121in]{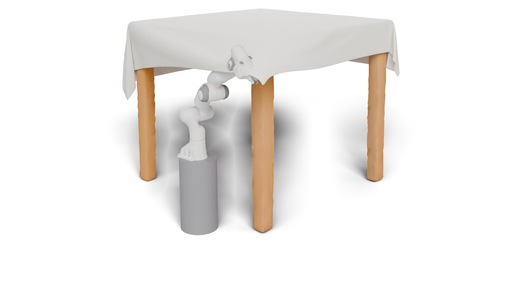} & \includegraphics[width=1.121in]{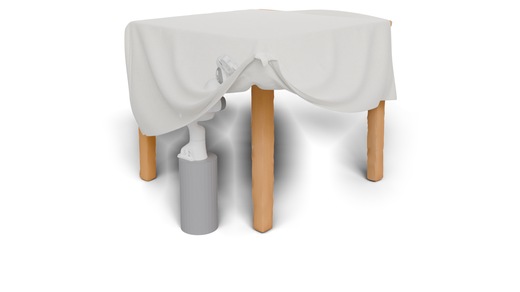} & \includegraphics[width=1.121in]{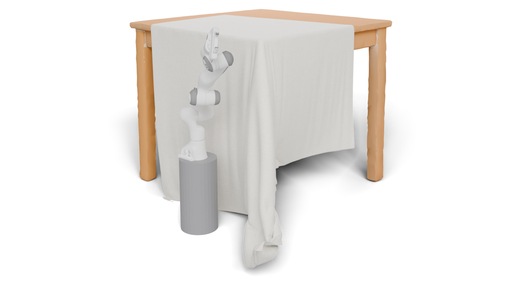} & \includegraphics[width=1.121in]{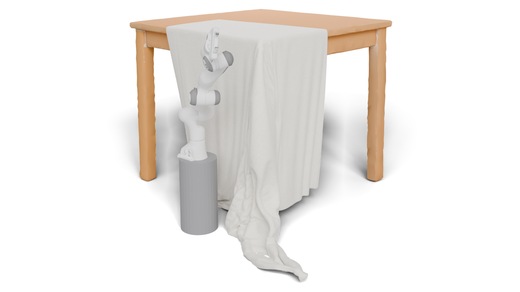} & \includegraphics[width=1.121in]{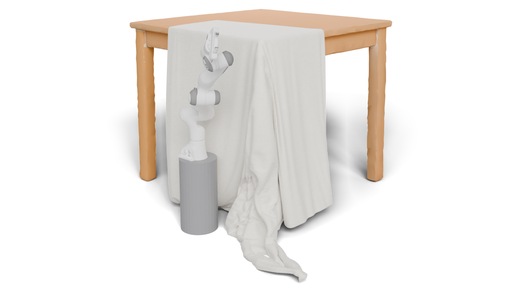}
\end{tabular}

  \caption{\textbf{Pull.} The tablecloth of the dining scene is
  gripped at its farthest corner after equilibration and drawn off the
  table, through the same soft position constraint as every other
  episode.}
  \label{fig:supp-pull-dining}
\end{figure*}

The same edit with one of the bed's pillows as the deleted body, gripped from under the quilt and carried out sideways, drags the quilt and the toys on it to the floor. This is the physically expected outcome of that command. However, it is not a useful edit, and the toy on top is the one to remove. Fig.~\ref{fig:supp-pull-dining} shows the tablecloth pull, which the main paper's figure omits for lack of space.

\begin{figure*}[!t]
  \centering
  \setlength{\tabcolsep}{0pt}%
\noindent\begin{tabular}{@{}c@{\hspace{0.03in}}c@{\hspace{0.03in}}c@{\hspace{0.03in}}c@{\hspace{0.03in}}c@{\hspace{0.03in}}c@{}}
\includegraphics[width=1.121in]{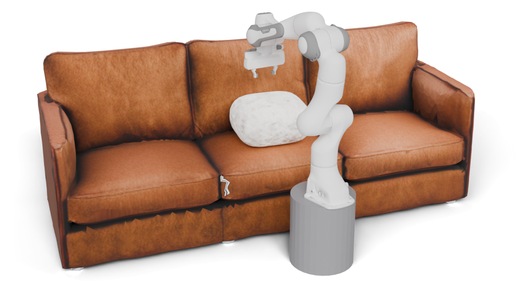} & \includegraphics[width=1.121in]{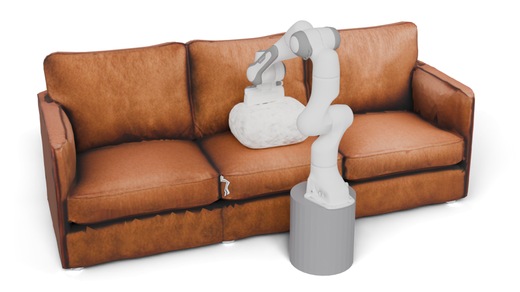} & \includegraphics[width=1.121in]{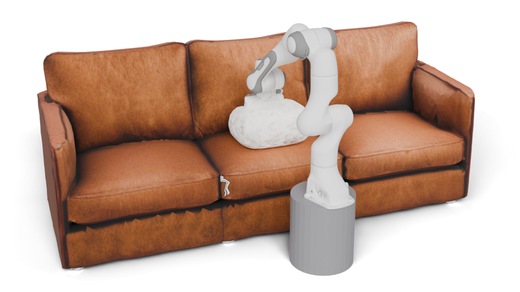} & \includegraphics[width=1.121in]{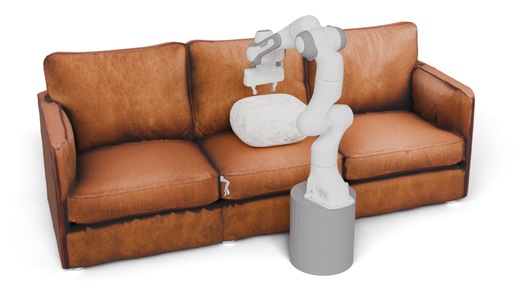} & \includegraphics[width=1.121in]{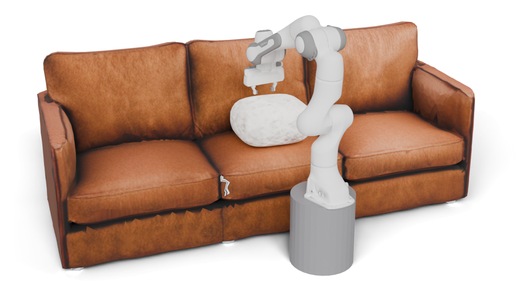} & \includegraphics[width=1.121in]{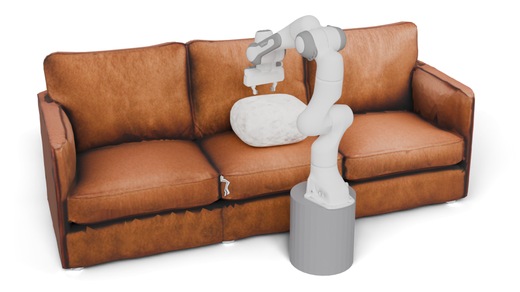}
\end{tabular}

  \caption{\textbf{Press.} A stuffed pillow on the leather couch is
  gripped at its crown and driven down by half its height. The
  couch's seat, a rigid collider in this scene, does not deform, so the
  indentation is all in the pillow.}
  \label{fig:supp-press-couch}
\end{figure*}

\begin{figure*}[!t]
  \centering
  \setlength{\tabcolsep}{0pt}%
\noindent\begin{tabular}{@{}c@{\hspace{0.03in}}c@{\hspace{0.03in}}c@{\hspace{0.03in}}c@{\hspace{0.03in}}c@{\hspace{0.03in}}c@{}}
\includegraphics[width=1.121in]{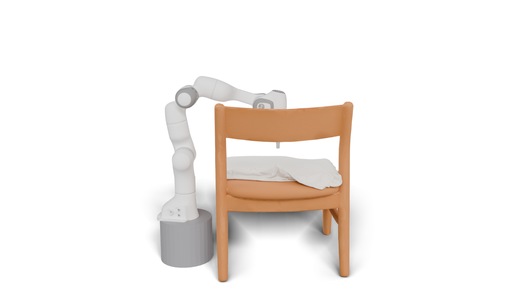} & \includegraphics[width=1.121in]{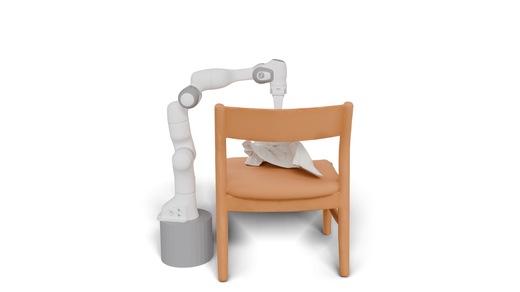} & \includegraphics[width=1.121in]{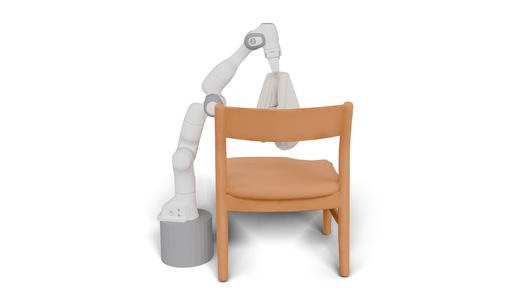} & \includegraphics[width=1.121in]{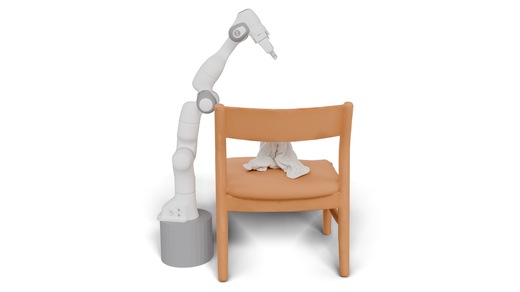} & \includegraphics[width=1.121in]{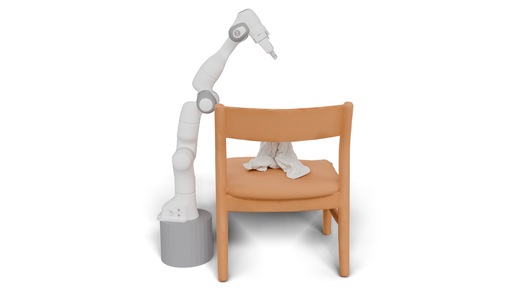} & \includegraphics[width=1.121in]{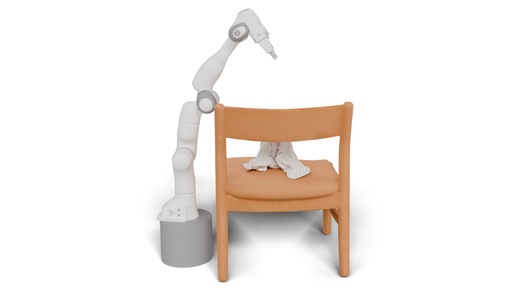}
\end{tabular}

  \caption{\textbf{Lift.} The towel on the chair is gripped at its top
  centre after equilibration and raised $0.5$\,m. It hangs from the grip,
  and on release it falls back onto the seat.}
  \label{fig:supp-lift-towel}
\end{figure*}

\paragraph{Videos.} Eleven videos accompany this supplement: the tidying episode of the second messy bed, the pull episodes of the stool cloth and the dining tablecloth (Fig.~\ref{fig:supp-pull-dining}), the press on the couch pillow (Fig.~\ref{fig:supp-press-couch}), the lift of the chair towel (Fig.~\ref{fig:supp-lift-towel}), and six equilibration runs with the state written out every five steps under a fixed camera: a throw onto a rocking chair, a cloth onto a picnic table, a sheet onto a bed, a quilt onto a daybed, a towel over a chair back, and the second messy bed.

\begin{table*}[t]
\centering\footnotesize
\setlength{\tabcolsep}{4pt}
\begin{tabular*}{\linewidth}{@{\extracolsep{\fill}}l l r ccc ccc c@{}}
\toprule
scene & episode & grip & err (mm) & max (mm) & reached & equilibrated & held & released & finite \\
\midrule
armchair & pull & 54 & 8.3 & 44.4 & 0.99 & 18.5 & 0.0 & 0.0 & \checkmark \\
bedsheet & pull & 21 & 19.3 & 75.6 & 0.99 & 10.6 & 4.4 & 0.0 & \checkmark \\
bench2b & pull & 144 & 6.1 & 42.6 & 1.00 & 48.1 & 0.0 & 0.0 & \checkmark \\
daybed & pull & 34 & 12.9 & 73.8 & 0.99 & 8.0 & 0.0 & 0.0 & \checkmark \\
dining2 & pull & 28 & 14.1 & 92.5 & 0.99 & 16.0 & 11.1 & 4.3 & \checkmark \\
dining & pull & 42 & 9.9 & 76.2 & 1.00 & 21.9 & 8.4 & 0.1 & \checkmark \\
picnic & pull & 25 & 15.4 & 67.3 & 0.99 & 11.2 & 5.6 & 5.8 & \checkmark \\
rocking & pull & 48 & 7.6 & 43.2 & 0.99 & 1.8 & 1.7 & 1.6 & \checkmark \\
stool2 & pull & 178 & 6.6 & 43.5 & 1.00 & 37.2 & 0.0 & 0.0 & \checkmark \\
stool & pull & 97 & 4.0 & 45.3 & 1.00 & 15.0 & 11.0 & 0.0 & \checkmark \\
apron & lift & 357 & 1.6 & 15.0 & 1.00 & 16.7 & 0.1 & 5.6 & \checkmark \\
bench & lift & 787 & 1.5 & 19.7 & 1.00 & 43.0 & 0.0 & 2.4 & \checkmark \\
blanket & lift & 90 & 5.3 & 21.0 & 0.99 & 5.9 & 1.5 & 3.5 & \checkmark \\
box & lift & 69 & 7.5 & 19.3 & 0.98 & 38.3 & 6.6 & 18.5 & \checkmark \\
coffee & lift & 1714 & 0.4 & 11.3 & 1.00 & 66.5 & 0.0 & 26.3 & \checkmark \\
kite & lift & 392 & 1.1 & 13.9 & 1.00 & 67.9 & 0.0 & 0.0 & \checkmark \\
throw & lift & 186 & 2.6 & 17.4 & 0.99 & 17.1 & 2.1 & 5.1 & \checkmark \\
towel4 & lift & 375 & 1.4 & 12.6 & 1.00 & 71.3 & 0.0 & 2.5 & \checkmark \\
bench & drag & 786 & 2.1 & 10.9 & 1.00 & 43.8 & 14.2 & 0.0 & \checkmark \\
daybed & drag & 74 & 5.7 & 27.4 & 0.99 & 8.8 & 6.1 & 5.6 & \checkmark \\
dining & drag & 65 & 5.4 & 26.2 & 1.00 & 21.6 & 15.7 & 16.0 & \checkmark \\
picnic & drag & 27 & 10.5 & 19.4 & 0.98 & 10.5 & 7.0 & 6.5 & \checkmark \\
stool & drag & 137 & 3.6 & 20.2 & 1.00 & 14.9 & 11.9 & 12.6 & \checkmark \\
couch & press & 75 & 14.9 & 42.4 & 0.78 & 1.2 & 4.7 & 1.4 & \checkmark \\
cushion & press & 54 & 16.5 & 41.7 & 0.74 & 2.7 & 5.2 & 2.9 & \checkmark \\
pillow2 & press & 22 & 17.8 & 41.3 & 0.72 & 5.7 & 8.4 & 5.8 & \checkmark \\
pillow3 & press & 50 & 16.4 & 39.7 & 0.75 & 6.1 & 9.3 & 6.3 & \checkmark \\
plush & press & 50 & 15.8 & 59.8 & 0.95 & 2.0 & 4.4 & 0.0 & \checkmark \\
\bottomrule
\end{tabular*}

\caption{\textbf{All manipulation episodes.} ``Grip'' is the number
of constrained vertices. Contact columns are the fraction of body
vertices within $5$\,mm of the support at static equilibrium, at the
end of the hold, and $1.8$\,s after release.}
\label{tab:supp-ixbench}
\end{table*}

\paragraph{Fold test.} CoDimRecon evaluates a folded sheet of paper by testing whether the crease is retained, which is enabled by its plastic hinges. Our shells are elastic, so a fold is maintained by friction and gravity alone, and we test whether this suffices for cloth. On \foldN{} equilibrated sheet scenes the gripper takes the corner farthest from the sheet's centre, carries it in an arc over the sheet to past the centre and sets it down, holds, and releases, after which the sheet runs free for $120$ steps (Table~\ref{tab:supp-fold}). A tenth episode, on the second dining table, is not scored because the constraint lost the patch during the carry. Retained is the horizontal travel of the corner that survives the release, as a fraction of its travel at the hold.

The fold remains held on \foldHeld{} of the \foldN{} sheets, with a median retention of \foldRetainedMed\%, wherever the folded corner comes to rest on cloth that itself lies on the support, and the contact fraction, halved by the fold, remains at the location where the hold occurs. However, it opens on the \foldOpened{} sheets whose hanging overhang outweighs the friction of the folded part, specifically the tablecloth on the dining table and the cloth on the stool, where the corner slides back over the edge. A crease in the paper sense requires plasticity, which is not modelled.

\begin{table}[tb]
\centering\scriptsize
\setlength{\tabcolsep}{3pt}
\begin{tabular*}{\linewidth}{@{\extracolsep{\fill}}l r r r r@{}}
\toprule
scene & carry (cm) & retained (\%) & contact, settled (\%) & contact, end (\%) \\
\midrule
bedsheet & 200 & 104 & 30 & 15 \\
bench & 59 & 100 & 69 & 37 \\
blanket & 141 & 99 & 10 & 4 \\
daybed & 147 & 100 & 6 & 4 \\
dining & 96 & 5 & 22 & 20 \\
picnic & 184 & 101 & 11 & 4 \\
stool & 47 & 19 & 14 & 14 \\
throw & 105 & 102 & 14 & 10 \\
towel4 & 68 & 98 & 60 & 25 \\
\bottomrule
\end{tabular*}

\caption{\textbf{Fold episodes.} Carry: horizontal travel of the gripped corner at the hold. Retained: the fraction of that travel surviving the release. Contact: fraction of sheet vertices within $5$\,mm of the support, at the settled state and at the end of the episode.}
\label{tab:supp-fold}
\end{table}

\section{Comparison with Other Systems}
\label{supp:compare}

In this section, we specify the conditions and the metrics of the comparison of Sec.~4.2 of the main paper and reports the studies that accompany it: the material recovery, the placement benchmarks, and the two systems that build their own scenes. The comparison holds everything fixed except the physics. Previous work and our stage read the same prompt, run the same reference-image, segmentation, asset-generation and layout stages, and therefore place the same generated meshes in the same arrangement. The difference lies in what happens to the deformable objects after layout: previous work leaves each generated asset frozen in its layout transform, whereas our stage gives the deformable objects an elastic energy and solves for the resting state.

\subsection{Conditions}
\label{sec:conditions-list}

Four conditions run on the same prompts.

\begin{description}
\item[Previous work.] The scene bundle as previous work emits it, taken after its own
physics optimization stage rather than before, so that the comparison
is against the complete prior pipeline and not against a raw layout.
That stage optimizes rigid placement, which is the most that a rigid
pipeline can do for a piece of cloth. Every generated asset, the cloth
included, stays frozen in the transform it comes to rest in.
\item[Blender cloth on the same assets.] This condition tests whether a generic cloth solver suffices. The
generated asset of the deformable object, as the rigid pipeline placed
it, is passed to Blender's cloth solver with the cotton preset, 12
substeps, the painted support and the floor as collision objects, and
150 frames, and is scored like everything else. Self-collision is off, because the asset
self-intersects by construction and the solver then explodes on the
first frame. A second run releases each cloth two centimetres above
the highest support surface under its ground projection, which is the one
adjustment a user would make manually.
\item[Blender cloth on our inputs.] This is the shared-input baseline. The sewn sheet at our placement,
our closed colliders and the floor are passed to the same Blender
cloth settings with self-collision on, since the sheet is
intersection-free, for $300$ frames. Volumetric bodies keep their volume
through Blender's pressure term, whereas ours are a tetrahedral foam,
so for sheets only the solver differs from ours and for volumetric bodies
the volume model differs as well.
\item[Ours.] The full stage, released from the rule-based pose. The
  gradient loop is not run.
\end{description}

\begin{table*}[t]
  \centering\footnotesize
  \setlength{\tabcolsep}{6pt}
  \begin{tabular*}{\linewidth}{@{\extracolsep{\fill}}l cccc@{}}
\toprule
intermediate & rigid & Blender, asset & Blender, ours & ours \\
\midrule
image, segmentation, assets & $\bullet$ & $\bullet$ & $\bullet$ & $\bullet$ \\
layout, support graph & $\bullet$ & $\bullet$ & $\bullet$ & $\bullet$ \\
cloth rest shape & asset & asset & sewn sheet & sewn sheet \\
support geometry & painted & painted & de-occluded, closed & de-occluded, closed \\
release pose & layout & layout & rules (Sec.~3.7) & rules (Sec.~3.7) \\
solver & rigid IPC & Blender cloth & Blender cloth & IPC, routed \\
gradient correction & no & no & no & no \\
\bottomrule
\end{tabular*}

  \caption{\textbf{Pipeline components shared by each condition of Table~2.}
  A dot marks an intermediate taken unchanged from the pipeline.}
  \label{tab:supp-share}
\end{table*}

\subsection{Metrics}

We evaluate geometry rather than renders. All five quantities are computed from the meshes alone and require no ground truth, so they apply to generated scenes. Because our stage rescales the scene to real-world sizes before solving, lengths measured in the layout frame are converted with the ratio of the two support bounding boxes. Every millimetre below is at the same physical scale on both sides.

Each deformable object is measured against every other body in its scene, rather than against the bodies that its own producer labelled rigid. The label is not consistent across systems. Our sofas arrive as one rigid mesh, while GPT-6 Astra decomposes the same sofa into an upholstered shell, two padded arms, three seat cushions and three welts and calls them soft. Scoring by label would credit a shirt lying on a cushion as resting on nothing, and it would fail to detect every blanket that passes through one.

For the same reason, the aggregates pool vertices instead of averaging over bodies. A body is any component that a system decides to export separately. Astra writes one scarf as a scarf plus thirty-four fringe strands, and a per-body mean therefore allows the trim to outweigh the garment it hangs from.

\begin{description}
\item[Crossings.] Cloth edges whose segment passes through another
  body's surface, reported per one thousand edges. This is the most
  visible failure.
\item[Penetration.] Mean distance from a cloth vertex to the nearest
  surface, over the vertices that fall inside a closed body. This
  measures how deep the penetration is rather than how often it occurs.
\item[Contact.] Fraction of cloth vertices within five millimetres of
  another surface. Draped fabric touches what it rests on over a broad
area, whereas a frozen asset touches only at the points that its
layout transform brought close.
\item[Floating.] Deformable objects more than two centimetres from
  everything else in the scene, that is, touching nothing at all.
\item[Unsupported.] Deformable objects with no chain of contacts reaching
  the ground. This is a stricter version of the previous measure, since
a body can touch something and still be unsupported if the body it
touches is itself unsupported. This case occurs in practice. Astra's
bench hovers $31$\,mm above its own floor and its armchair $12$\,mm,
so a towel correctly resting on that bench is unsupported without
floating, and the defect belongs to the furniture rather than to the
cloth. The ground counts as support for this chain but is not a body
for the previous measure, so a towel lying on the floor touches
nothing and is still supported, which is how the bathtub towel scores.
\item[Blocked.] The fraction of cloth vertices with support surface
  directly above them, measured by casting a ray upward from each
  vertex. A cloth resting \emph{on} something has nothing of that
something above it, so this measure is near zero for a normal drape
and large when a body has come to rest on an up-facing surface that is
not the top, or hangs beneath one, for example a skirt under an
overhanging table top, a lower rail, a stretcher, or the slats under a
seat. It is reported and not optimized against. It is the one measure
here that separates a valid resting state from the intended one.
\item[Off support.] Deformable objects whose centroid falls outside the
  ground projection of the supports.
\end{description}

With $\mathcal{E}$ and $\mathcal{V}$ denoting the edges and vertices of a deformable object, $\partial\mathcal{O}$ denoting the union of the other bodies' surfaces, $\mathcal{V}_{\mathrm{in}}\subset\mathcal{V}$ denoting the vertices inside a closed body, and $\mathcal{S}$ denoting the supports, we define the per-body quantities as
\begin{align}
X &= \frac{10^{3}}{|\mathcal{E}|}\,\big|\{e\in\mathcal{E}: e\cap\partial\mathcal{O}\neq\emptyset\}\big|, \nonumber\\
P &= \frac{1}{|\mathcal{V}_{\mathrm{in}}|}\sum_{i\in\mathcal{V}_{\mathrm{in}}}\operatorname{dist}(\bm{x}_i,\partial\mathcal{O}), \nonumber\\
C &= \frac{1}{|\mathcal{V}|}\,\big|\{i: \operatorname{dist}(\bm{x}_i,\partial\mathcal{O})<5\,\mathrm{mm}\}\big|, \nonumber\\
B &= \frac{1}{|\mathcal{V}|}\,\big|\{i: \mathrm{ray}(\bm{x}_i,+\hat{\bm{y}})\cap\mathcal{S}\neq\emptyset\}\big|,
\label{eq:supp-metrics}
\end{align}
A body is floating when $\min_i\operatorname{dist}(\bm{x}_i,\partial\mathcal{O}\cup\text{floor})>2$\,cm. It is unsupported when no chain of contacts within $2$\,cm links it to the ground. It is off its support when $\pi(\bar{\bm{c}})\notin\mathcal{F}$ for its centroid $\bar{\bm{c}}$ and the supports' ground projection $\mathcal{F}$. Scene aggregates pool edges and vertices over the bodies before the ratios are taken.

\begin{figure*}[!t]
  \centering
  \setlength{\tabcolsep}{0pt}%
\noindent\begin{tabular}{@{}c@{\hspace{3pt}}c@{\hspace{0.03in}}c@{\hspace{0.03in}}c@{\hspace{0.03in}}c@{\hspace{0.03in}}c@{\hspace{0.03in}}c@{\hspace{0.03in}}c@{\hspace{0.03in}}c@{}}
 & \eyebrow{folded sheet}& \eyebrow{dining table}& \eyebrow{stool}& \eyebrow{rocking chair}& \eyebrow{towel, chair}& \eyebrow{armchair}& \eyebrow{daybed}& \eyebrow{picnic table}\\[1pt]
\rotatebox{90}{\makebox[0.539in]{\iconbad\,\eyebrow{rigid}}} & \includegraphics[width=0.808in]{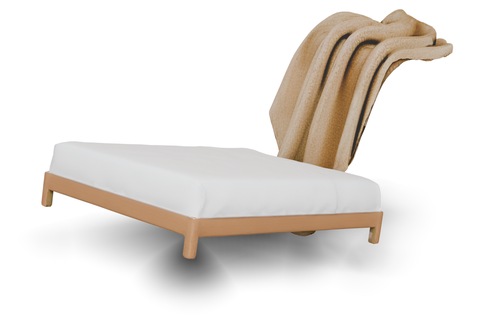} & \includegraphics[width=0.808in]{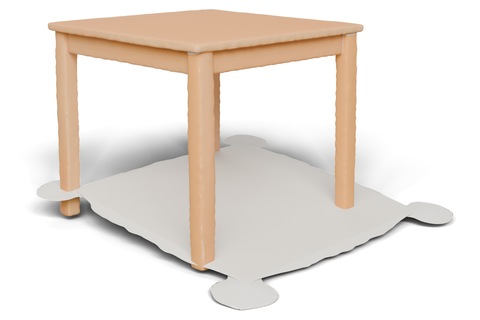} & \includegraphics[width=0.808in]{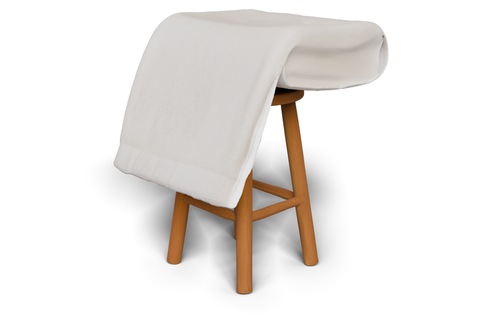} & \includegraphics[width=0.808in]{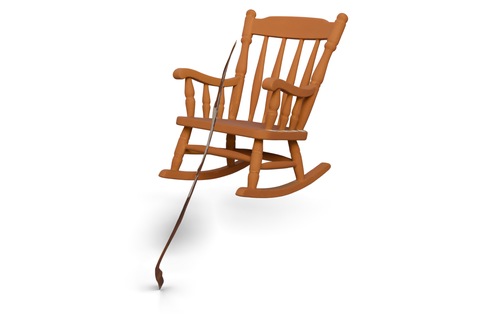} & \includegraphics[width=0.808in]{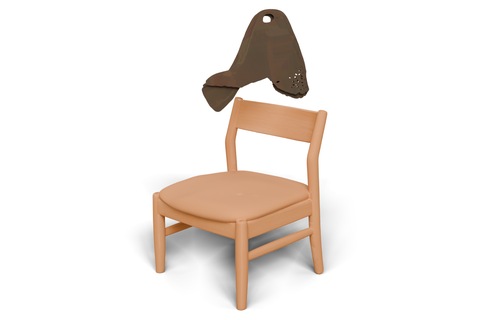} & \includegraphics[width=0.808in]{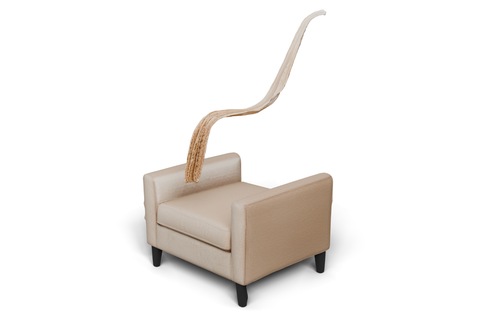} & \includegraphics[width=0.808in]{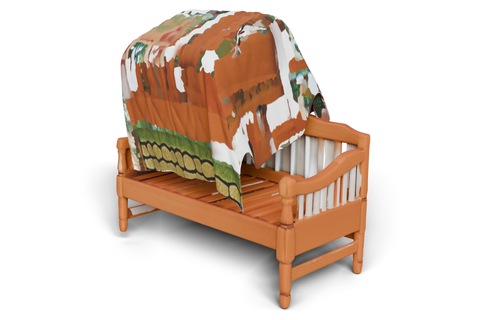} & \includegraphics[width=0.808in]{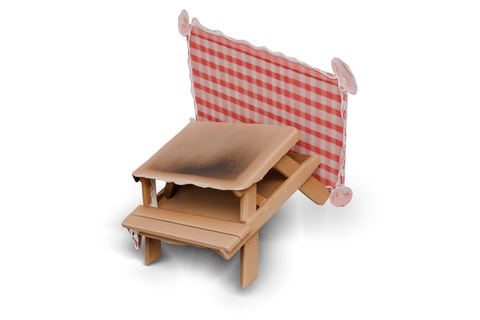} \\[3pt]
\rotatebox{90}{\makebox[0.539in]{\iconok\,\eyebrow{ours}}} & \includegraphics[width=0.808in]{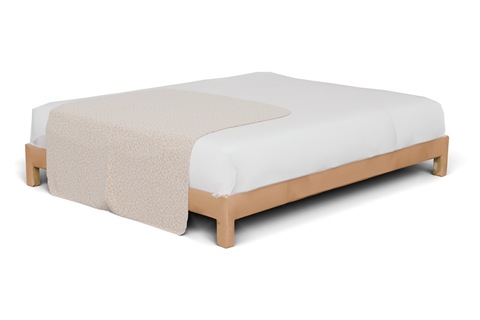} & \includegraphics[width=0.808in]{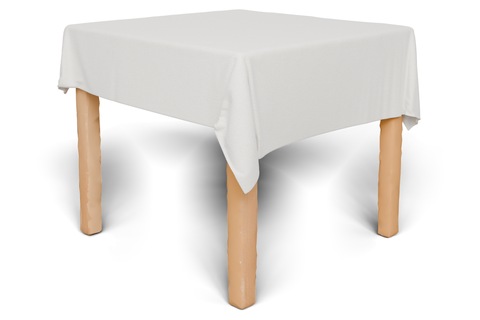} & \includegraphics[width=0.808in]{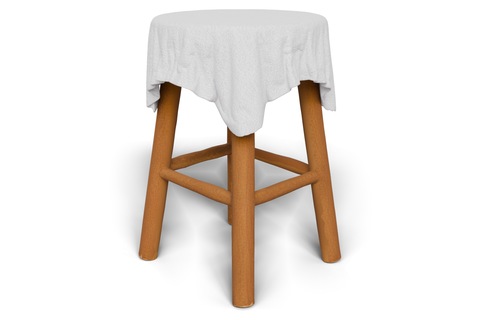} & \includegraphics[width=0.808in]{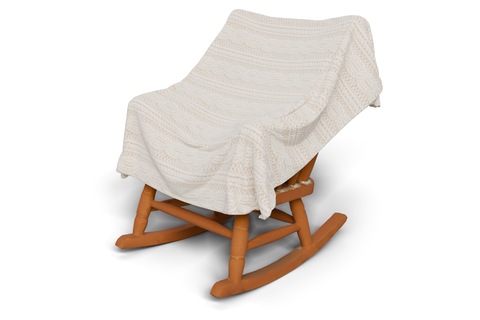} & \includegraphics[width=0.808in]{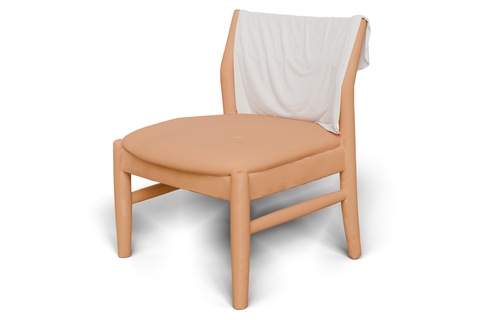} & \includegraphics[width=0.808in]{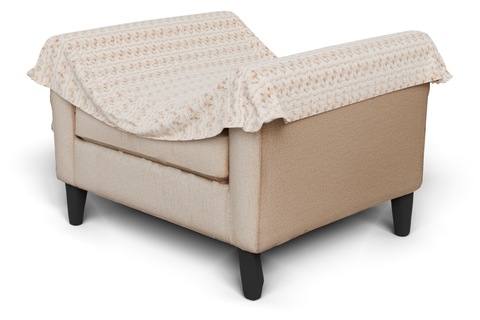} & \includegraphics[width=0.808in]{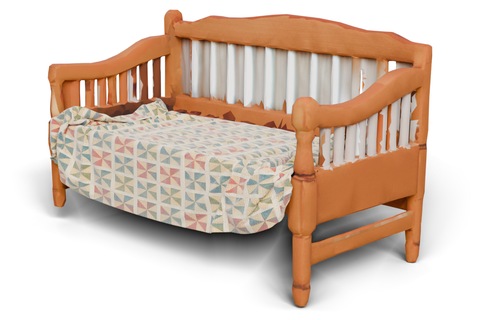} & \includegraphics[width=0.808in]{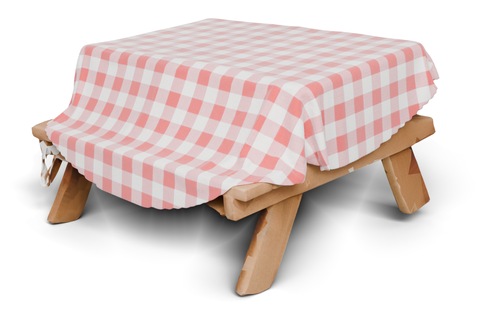} \\[3pt]
\end{tabular}

  \caption{\textbf{The rigid pipeline and ours on the same prompts and assets.} Top
  row of each pair: previous work, every generated object kept rigid.
  Bottom row: the same scene after our stage. The cloth in the top row
  keeps the pose the generator gave it, which is why it hovers, tilts,
  and passes through the furniture it is meant to rest on.}
  \label{fig:supp-compare-grid}
\end{figure*}

\subsection{Results}
The aggregate over the scene set is presented in Table~\ref{tab:supp-compare-summary}, and the per-scene record is presented in Table~\ref{tab:supp-compare-full}. Both come from the evaluator above, and the main paper discusses them.

\paragraph{Completion criterion for every system.} Table~\ref{tab:supp-e2e} applies the main paper's completion criterion to every system of the comparison. For each scene, \emph{present} means that the system places the requested deformable object, \emph{finished} means that its run does not diverge, $X{=}0$ means that no deformable edge crosses another body, $G$ means that the object's centre of mass projects inside the ground projection of the body it rests on, that it rests on that body rather than floating or lying on the floor, and that the body is not a solid block, and $J$ denotes the prompt check of Sec.~3.5, which we apply to the same vision--language model on each system's render: our studio camera for the rigid pipeline, Astra and Blender, and Holodeck's own top-down render, since its scenes are not in our frame. We run the prompt check on every scene for Holodeck, Astra, the rigid pipeline and ours, and on the scenes that pass $X{=}0$ and $G$ for the Blender conditions, which is sufficient for the completion count.

Blender's solver on our inputs completes \eEEBlFComplete{} scenes, and every completed scene is a pillow, cushion, plush toy or a sheet that lies flat, and \eEEBlQComplete{} at 40 substeps. The rigid pipeline completes none, since its rigid shell either crosses the furniture or floats above it. Holodeck completes \eEEHoloComplete{} and Astra \eEEAstraComplete{}. SoftPAT3D completes \eEEOursComplete{} of the 34 scenes, and every scene that reaches the solver passes $X{=}0$.

\begin{table}[tb]
\centering\scriptsize
\setlength{\tabcolsep}{3pt}
\begin{tabular*}{\linewidth}{@{\extracolsep{\fill}}l r rrrrrr@{}}
\toprule
system & scenes & present & finished & $X{=}0$ & $G$ & $J$ & complete \\
\midrule
Holodeck & 32 & 8 & 8 & 5 & 2 & 2 & 2 \\
Astra & 29 & 28 & 28 & 14 & 15 & 19 & 7 \\
rigid & 34 & 34 & 34 & 14 & 15 & 11 & 0 \\
Blender & 34 & 34 & 22 & 11 & 11 & 0 & 0 \\
Blender$^{*}$ & 34 & 32 & 32 & 23 & 19 & 8 & 8 \\
Blender$^{*}$, 40 substeps & 34 & 30 & 30 & 25 & 14 & 7 & 7 \\
ours & 34 & 34 & 34 & 34 & 30 & 22 & 21 \\
\bottomrule
\end{tabular*}

\caption{\textbf{Completion criterion applied to every system.} Scenes of the comparison set in which the deformable object is present, the run finished, no edge crosses another body, both geometric checks pass with the object resting on its support, the prompt check passes, and all of these hold, Eq.~6 of the main paper.}
\label{tab:supp-e2e}
\end{table}

\paragraph{Blender's cloth solver on the generated asset.} Blender's cloth solver on the generated asset, placed where the rigid pipeline put it, diverges in $24$ of $34$ scenes. The cloth's bounding box grows past four times the support's, crosses the furniture at $196$ edges per thousand, and leaves $23\%$ of cloth vertices in contact. The asset starts inside its support, from which a cloth solver cannot recover, and in the scenes that do not diverge it falls through the seat to the floor. Released two centimetres above the support, it still diverges in $24$ of $34$, crosses at $190$ per thousand, and reaches $33\%$ contact, because the asset's own two layers and folds are now the rest shape that the solver preserves. This condition isolates what the whole stage adds over the solver alone, and the shared-input run below isolates the solver.

\paragraph{Blender's cloth solver on our inputs.} Given the sewn sheet, the placement and the closed colliders from which our solve starts, Blender diverges in \bfDiv{} of \bfScenes{} scenes. It crosses at \bfCross{} edges per thousand with \bfPen\,mm penetration, and leaves \bfContact\% of cloth vertices in contact, \bfFloat{} bodies touching nothing and \bfUnsup{} unsupported. Two scenes, the napkin on a plate and the bamboo steamer, do not finish within the job limit and are not counted. At $40$ substeps instead of the preset's $12$, with everything else the same, it diverges in \bfqDiv{} of the \bfqScenes{} scenes that finish within twelve hours, which excludes the apron, the ladder, the napkin and the steamer. It crosses at \bfqCross{} per thousand with \bfqPen\,mm penetration, and leaves \bfqContact\% of cloth vertices in contact and \bfqFloat{} bodies touching nothing. More substeps remove about two thirds of the crossings, partly because the ladder, one of the difficult scenes below, is excluded. However, they remove none of the floating bodies, so step size is not what separates the two solvers.

The crossings are concentrated in a few scenes. Five scenes account for all but a few of them: the two bathtubs, the ladder, the laundry basket and the clothes pile, where the support is concave or the bodies stack and Blender's collision response allows the sheet to pass through. Elsewhere, the sheet stays intersection-free but does not stay in place. In eleven scenes it slides off the support and ends in contact with nothing, among them the apron, the bedsheet, the throw, the skirt and the napkin on the coffee table, and only three scenes reach the contact that our simulation reaches. This is the column that separates the preprocessing from the solver in Table~3 of the main paper: the preprocessing removes the divergence, and the solver decides where the cloth comes to rest.

\begin{table}[tb]
  \centering
  \scriptsize
  \setlength{\tabcolsep}{2.5pt}
  \begin{tabular*}{\linewidth}{@{\extracolsep{\fill}}l cccccc@{}}
\toprule
& Holodeck & Astra & rigid & Blender & Blender$^{*}$ & ours \\
\midrule
scenes & 32 & 29 & 34 & 34 & \bfScenes & 34 \\
deformable present & 8/32 & 29/32 & 34/34 & 34/34 & \bfPresent & 34/34 \\
soft bodies & 39 & 74 & 48 & 48 & \bfBodies & 43 \\
crossings (per 1k edges) & 71.2 & 40.2 & 5.1 & 196.2 & \bfCross & 0.0 \\
penetration (mm) & 0.0 & 1.7 & 14.8 & 1.9 & \bfPen & 0.0 \\
in contact (\%) & 10.3 & 38.5 & 1.8 & 22.8 & \bfContact & 39.9 \\
touching nothing & 22/39 & 1/74 & 14/48 & 4/48 & \bfFloat & 1/43 \\
unsupported & 23/39 & 5/74 & 15/48 & 0/48 & \bfUnsup & 0/43 \\
\midrule
geometric checks & \eEEHoloGeom & \eEEAstraGeom & \eEERigidGeom & \eEEBlGeom & \eEEBlFGeom & \eEEOursGeom \\
complete & \eEEHoloComplete & \eEEAstraComplete & \eEERigidComplete & \eEEBlComplete & \eEEBlFComplete & \eEEOursComplete \\
\bottomrule
\end{tabular*}

  \caption{\textbf{Aggregate comparison} over the comparison set. The rigid pipeline,
  both Blender runs and ours share prompts and generated assets.
  Holodeck and Astra build their own scenes from the prompts.}
  \label{tab:supp-compare-summary}
\end{table}

\begin{table}[tb]
  \centering
  \scriptsize
  \setlength{\tabcolsep}{3pt}
  \begin{tabular*}{\linewidth}{@{\extracolsep{\fill}}l cc cc cc cc@{}}
\toprule
& \multicolumn{2}{c}{cross/1k} & \multicolumn{2}{c}{pen. (mm)} & \multicolumn{2}{c}{contact (\%)} & \multicolumn{2}{c}{floating} \\
\cmidrule(lr){2-3}\cmidrule(lr){4-5}\cmidrule(lr){6-7}\cmidrule(lr){8-9}
scene & prev. & ours & prev. & ours & prev. & ours & prev. & ours \\
\midrule
apron & 0.0 & 0.0 & 0.0 & 0.0 & 0.0 & 23.1 & 1 & 0 \\
armchair & 0.0 & 0.0 & 0.0 & 0.0 & 0.0 & 37.5 & 1 & 0 \\
bathtub2 & 25.8 & 0.0 & 2.0 & 0.0 & 3.2 & 54.8 & 0 & 0 \\
bathtub & 0.0 & 0.0 & 0.0 & 0.0 & 0.0 & 4.0 & 1 & 0 \\
bedsheet & 0.0 & 0.0 & 0.0 & 0.0 & 0.0 & 55.7 & 1 & 0 \\
bench2b & 0.0 & 0.0 & 0.0 & 0.0 & 0.0 & 60.6 & 1 & 0 \\
bench & 18.0 & 0.0 & 14.8 & 0.0 & 2.5 & 64.5 & 0 & 0 \\
blanket & 0.0 & 0.0 & 0.0 & 0.0 & 0.0 & 32.0 & 1 & 0 \\
box & 0.0 & 0.0 & 0.0 & 0.0 & 0.0 & 97.6 & 1 & 0 \\
clothes sofa & 3.2 & 0.0 & 40.1 & 0.0 & 0.5 & 19.3 & 1 & 0 \\
coffee & 0.0 & 0.0 & 15.7 & 0.0 & 0.0 & 100.0 & 0 & 0 \\
couch & 0.0 & 0.0 & 0.0 & 0.0 & 0.0 & 3.4 & 1 & 0 \\
cushion & 14.2 & 0.0 & 1.0 & 0.0 & 13.7 & 4.4 & 0 & 0 \\
daybed & 5.8 & 0.0 & 21.1 & 0.0 & 0.5 & 34.2 & 0 & 0 \\
dining2 & 9.8 & 0.0 & 22.7 & 0.0 & 0.5 & 44.7 & 0 & 0 \\
dining & 16.8 & 0.0 & 7.6 & 0.0 & 4.6 & 26.6 & 0 & 0 \\
kite & 0.0 & 0.0 & 0.0 & 0.0 & 0.0 & 99.1 & 1 & 0 \\
ladder & 11.5 & 0.0 & 6.0 & 0.0 & 5.2 & 28.0 & 0 & 0 \\
laundry & 1.3 & 0.0 & 0.0 & 0.0 & 0.8 & 76.0 & 0 & 0 \\
napkin & 1.0 & 0.0 & 0.3 & 0.0 & 1.4 & 11.3 & 0 & 0 \\
picnic & 7.8 & 0.0 & 3.6 & 0.0 & 1.7 & 48.9 & 0 & 0 \\
pillow2 & 16.8 & 0.0 & 9.5 & 0.0 & 5.1 & 9.1 & 0 & 0 \\
pillow3 & 0.0 & 0.0 & 0.0 & 0.0 & 0.0 & 6.5 & 1 & 0 \\
pillow & 0.2 & 0.0 & 1.2 & 0.0 & 0.3 & 9.2 & 0 & 0 \\
plush & 0.8 & 0.0 & 2.6 & 0.0 & 0.5 & 1.8 & 0 & 0 \\
rocking & 0.0 & 0.0 & 0.0 & 0.0 & 0.0 & 4.2 & 1 & 0 \\
skirt & 37.5 & 0.0 & 12.1 & 0.0 & 2.1 & 60.4 & 0 & 0 \\
steamer & 1.7 & 0.0 & 1.7 & 0.0 & 4.7 & 21.3 & 0 & 0 \\
stool2 & 1.8 & 0.0 & 1.2 & 0.0 & 1.4 & 38.8 & 0 & 0 \\
stool & 0.0 & 0.0 & 0.0 & 0.0 & 0.0 & 14.9 & 1 & 0 \\
throw & 11.8 & 0.0 & 18.4 & 0.0 & 2.9 & 50.9 & 0 & 0 \\
tissue & 12.8 & 0.0 & 0.9 & 0.0 & 1.9 & 0.0 & 0 & 1 \\
towel4 & 0.0 & 0.0 & 0.0 & 0.0 & 0.0 & 30.8 & 1 & 0 \\
trash & 9.5 & 0.0 & 0.0 & 0.0 & 2.2 & 9.9 & 0 & 0 \\
\bottomrule
\end{tabular*}

  \caption{\textbf{Per-scene comparison.} ``prev.'' is the rigid pipeline, the
  generated asset frozen in its layout transform. ``Ours'' replaces
  it by the sewn sheet and solves. Deformable objects are taken from
the pipeline's material library rather than filtered by name.
Table~\ref{tab:supp-objects} counts them per scene.}
  \label{tab:supp-compare-full}
\end{table}

\begin{table}[tb]
  \centering
  \scriptsize
  \setlength{\tabcolsep}{4pt}
  \begin{tabular*}{\linewidth}{@{\extracolsep{\fill}}l cccc@{}}
\toprule
scene & requested & generated & soft bodies solved & soft bodies scored \\
\midrule
apron & 2 & 2 & 1 & 1 \\
armchair & 2 & 2 & 1 & 1 \\
bathtub2 & 2 & 2 & 1 & 1 \\
bathtub & 2 & 2 & 1 & 1 \\
bedsheet & 2 & 2 & 1 & 1 \\
bench2b & 2 & 2 & 1 & 1 \\
bench & 2 & 2 & 1 & 1 \\
blanket & 2 & 2 & 1 & 1 \\
box & 2 & 2 & 1 & 1 \\
clothes sofa & 2 & 7 & 6 & 6 \\
coffee & 2 & 2 & 1 & 1 \\
couch & 2 & 2 & 1 & 1 \\
cushion & 2 & 2 & 1 & 1 \\
daybed & 2 & 2 & 1 & 1 \\
dining2 & 2 & 2 & 1 & 1 \\
dining & 2 & 2 & 1 & 1 \\
kite & 2 & 2 & 1 & 1 \\
ladder & 2 & 2 & 1 & 1 \\
napkin & 3 & 3 & 1 & 1 \\
picnic & 2 & 2 & 1 & 1 \\
pillow2 & 2 & 2 & 1 & 1 \\
pillow3 & 2 & 2 & 1 & 1 \\
pillow & 2 & 2 & 1 & 1 \\
plush & 2 & 2 & 1 & 1 \\
rocking & 2 & 2 & 1 & 1 \\
skirt & 2 & 2 & 1 & 1 \\
steamer & 2 & 3 & 2 & 2 \\
stool2 & 2 & 2 & 1 & 1 \\
stool & 2 & 2 & 1 & 1 \\
throw & 2 & 2 & 1 & 1 \\
tissue & 3 & 3 & 1 & 1 \\
towel4 & 2 & 2 & 1 & 1 \\
trash & 2 & 2 & 1 & 1 \\
\midrule
total & 68 & 74 & 39 & 39 \\
\bottomrule
\end{tabular*}

  \caption{\textbf{Objects per scene.} What each scene of the comparison set asked for, received
  and scored: objects in the parsed request, objects the generator
  produced, deformable objects the solver simulated, and deformable
  objects the evaluator scored. The two columns on the right agree in
  every scene. The clothes sofa's six garments come from one request
  entry, and the steamer's two are its dumplings.}
  \label{tab:supp-objects}
\end{table}

\paragraph{Human raters.} To validate the prompt check against human judgments, we ask \hcRaters{} raters from outside the author list to answer the same question on the same \hcN{} main-run renders, one view each, in a randomized order, without the model's verdict (the rating page and the responses are in the released code). The raters agree with one another on \hcInter\% of the renders and are unanimous on \hcUnanYes{} accepted scenes. Each rater returns the check's verdict on \hcMeanAgree\% of the renders on average, \hcMeanAgreeN{} of \hcN{}, and departs from it on at most six, nearly all of them placements in which the object sits on the right furniture at a position the raters read differently, a blanket at the foot of a bed or a towel on a glass table. Therefore, the completion counts of the main paper agree with human judgment on all but a few main-run scenes. Neither rating study collected personal data beyond a nickname, and neither involved risk to the raters, who took part voluntarily with informed consent; they required no ethics-board review under our institution's rules.

\paragraph{User study.} We further evaluate the systems under human judgment. \pairRaters{} participants from outside the author list are shown, for each of \pairN{} prompts drawn at random from the comparison set, our rendering beside that of one baseline, the rigid pipeline (\pairRigid{} pairs), GPT-6 Astra (\pairAstra{}) or Holodeck (\pairHolo{}), with the systems unidentified, the sides randomized and the pairs in random order. For each pair, participants answer two questions: which scene is more physically valid, and which better matches the description. Our method is preferred by majority on \pairPhysOurs{} of \pairN{} pairs for physical validity and on \pairPromptOurs{} of \pairN{} for prompt satisfaction, and receives \pairPhysVotes{} and \pairPromptVotes{} of the \pairVotes{} individual votes on the two questions; no vote goes to a baseline, and the participants agree with one another on \pairInterPhys\% and \pairInterPrompt\% of their answers. The two questions separate physical validity from semantic consistency, and our advantage holds on both, including against GPT-6 Astra, whose renderings depict convincing fabric that does not rest on the furniture.

\subsection{Material recovery}
\label{supp:material}

The release benchmark requires the loop of Sec.~3.6 to move a cloth, and this benchmark requires it to explain a drape. On each support a sheet is equilibrated with a known weight-to-stiffness ratio, which fixes how far it sags between its contacts, and the equilibrium state is taken as the target. The ratio is then perturbed by a factor of four, heavier or lighter, and three conditions attempt to bring the equilibrated drape back onto the target: keeping the perturbed ratio, random search over the log-ratio with twice the gradient's forward budget, and twenty Adam steps on the log-ratio through the differentiable rollout. With the areal weight parameterized as $w=\bar w\,e^{z}$ and $\bm{x}^{\star}(z)$ the equilibrium state under it, the loss is the mean vertex distance to the target, $L_m(z)=\frac{1}{|\mathcal{V}|}\sum_i\lVert\bm{x}^{\star}_i(z)-\bm{x}^{\mathrm{tgt}}_i\rVert$, and $\mathrm{d}L_m/\mathrm{d}z$ comes from the same adjoint sweep as the placement gradient, through $\partial\bm{R}_n/\partial z$ in the mass term. The ratio is applied to the areal weight so that the integrator's step is the same in every condition.

Table~\ref{tab:supp-material} reports, over \matCases{} cases on \matScenes{} supports, the parameter error $|\log_4(w/w^\star)|$, for which 1 is the starting error, and the drape error. A case costs about \matSec{} minutes on one A5000. The landscape of this problem is smooth, unlike that of the placement. The gradient recovers the ratio to a median of \matOursMedPct\%, with all \matCases{} cases within $10\%$, in 20 rollouts, and the search, 40 draws from a normal of width 0.8 in the natural log around the wrong start, reaches a median of \matSearchMedPct\%. The search's draws are seeded identically in every case, so its best sample lies at the same two distances from the true value, 11\% and 20\%, throughout. However, a different seed or a one-dimensional line search would remove that gap, so the table is a check that the material gradient is correct and efficient, not evidence that search cannot solve a one-parameter problem. The value of the adjoint is that the same backward sweep returns the gradient of every parameter of the scene at once.

The material library's fabrics themselves differ little in drape at this scale. The rocking-chair blanket re-solved as silk, cotton, fleece and wool, which span a membrane stiffness of $5\times10^{5}$ to $10^{6}$\,Pa, a bending stiffness of $0.2$ to $2.0$ and an areal weight of $0.07$ to $0.42$\,kg/m$^{2}$, equilibrates to nearly the same shape. The sag between the chair's arms and the hang of the hem are the only visible differences. The sag is governed by the weight-to-stiffness ratio, which is what the recovery above measures.

\begin{table}[tb]
  \centering\scriptsize
  \setlength{\tabcolsep}{3pt}
  \begin{tabular*}{\linewidth}{@{\extracolsep{\fill}}l rrrr@{}}
\toprule
 & med.\ $|\log_4(w/w^\star)|$ & $\leq 10\%$ (\%) & med.\ dist.\ (cm) & $\leq 1$\,mm (\%) \\
\midrule
naive & 1.00 & 0 & 0.25 & 23 \\
search & 0.11 & 0 & 0.02 & 69 \\
ours & \textbf{0.02} & \textbf{100} & \textbf{0.00} & \textbf{73} \\
\bottomrule
\end{tabular*}

  \caption{\textbf{Material recovery} over \matCases{} cases: the weight-to-stiffness ratio, perturbed by a factor of four, recovered from the equilibrated drape.}
  \label{tab:supp-material}
\end{table}

\subsection{Placement benchmark}
\label{supp:release}

The language model specifies where a deformable object should come to rest, and the quantity under our control is the pose from which it is released. Each target is a point on the landing surface of a support. Its difficulty is defined geometrically as the fraction of the sheet's own ground projection $\mathcal{F}_s$ whose downward ray does not reach that surface $\mathcal{T}$, $\delta=\frac{1}{|\mathcal{F}_s|}\,|\{\bm{q}\in\mathcal{F}_s:\mathrm{ray}(\bm{q},-\hat{\bm{y}})\cap\mathcal{T}=\emptyset\}|$, because a backrest or a rim obstructs the path or the surface is smaller than the cloth. When $\delta=1/3$ is exceeded, the target is classified as obstructed. Three conditions use one evaluator and a matched forward budget: a straight drop above the target, random search over the two placement coordinates with twice the forward rollouts of the gradient run, and Adam on the gradient of the placement objective of Sec.~3.6 of the main paper through the equilibrium state. The evaluator weights cloth vertices by contact with the support, so it is well-defined whether the cloth lies flat or drapes over the side, and a cloth touching nothing counts as a miss rather than as a distance.

\begin{table*}[t]
  \centering\scriptsize
  \setlength{\tabcolsep}{2pt}
  \begin{minipage}[t]{0.62\textwidth}\centering\begin{tabular*}{\linewidth}[t]{@{\extracolsep{\fill}}l r c ccc c ccc@{}}
\toprule
& & & \multicolumn{3}{c}{bodies landed (\%)} & & \multicolumn{3}{c}{mean error (cm)} \\
\cmidrule(lr){4-6}\cmidrule(lr){8-10}
targets & $N$ & runs & naive & search & ours & & naive & search & ours \\
\midrule
obstructed, lr\,0.04 & 1 & 4 & 50 & 75 & \textbf{100} & & 2.2 & 9.1 & 12.6 \\
 & 2 & 4 & 50 & 62 & \textbf{100} & & 2.0 & 2.0 & 13.5 \\
 & 4 & 4 & 50 & 69 & \textbf{100} & & 2.3 & 2.3 & 13.8 \\
\midrule
obstructed, lr\,0.01 & 4 & 4 & 50 & 69 & 50 & & 2.3 & 8.5 & \textbf{1.2} \\
\midrule
obstructed, two-phase & 2 & 4 & 50 & 75 & \textbf{100} & & 2.0 & 9.7 & 12.9 \\
 & 4 & 4 & 50 & 69 & \textbf{100} & & 2.3 & 8.5 & 12.5 \\
\bottomrule
\end{tabular*}
\end{minipage}
  \caption{\textbf{Joint placement optimization of $N$ coupled sheets} on the obstructed layout, at the optimiser steps discussed in the text.
  ``Landed'' is the solver's own flag, contact with the support. Error
  is the contact patch's distance from the requested point over the
  sheets \emph{that method} landed, so it is comparable across columns
  only where landing rates agree. Targets no
  method ever reached are excluded. ``Runs'' counts the runs that enter
  the landing rate.}
  \label{tab:supp-release}\label{tab:supp-multi}
\end{table*}

\subsection{Joint optimization of several bodies}
\label{supp:multi}

\begin{figure*}[!t]
  \centering
  \setlength{\tabcolsep}{0pt}%
\noindent\begin{tabular}{@{}c@{\hspace{0.03in}}c@{\hspace{0.03in}}c@{\hspace{0.03in}}c@{\hspace{0.03in}}c@{}}
\includegraphics[width=1.351in]{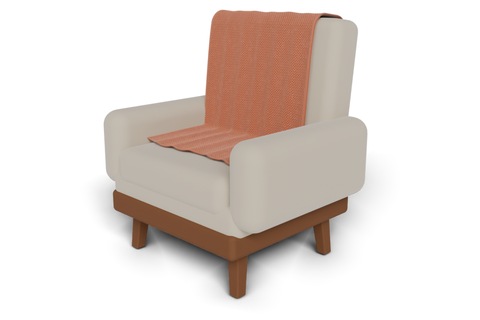} & \includegraphics[width=1.351in]{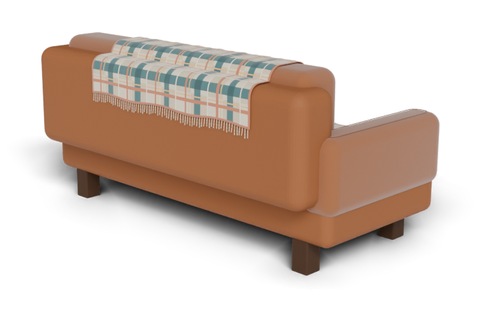} & \includegraphics[width=1.351in]{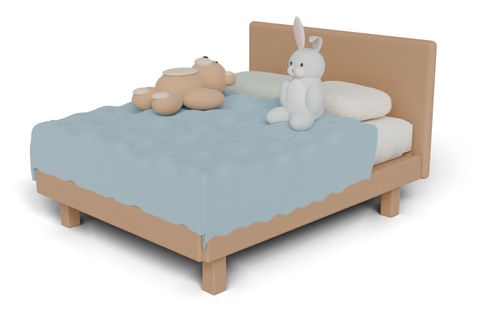} & \includegraphics[width=1.351in]{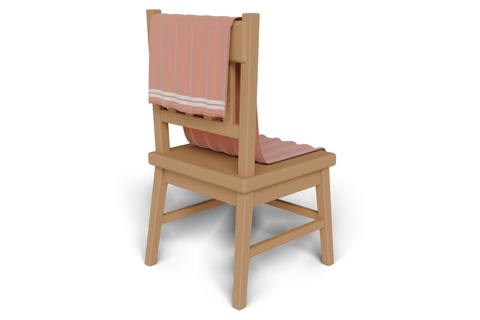} & \includegraphics[width=1.351in]{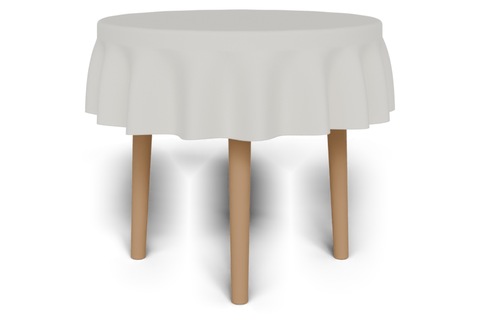} \\[1pt]
\figcap{blanket, armchair}& \figcap{throw, sofa}& \figcap{messy bed}& \figcap{towel, chair}& \figcap{tablecloth}\\[4pt]
\includegraphics[width=1.351in]{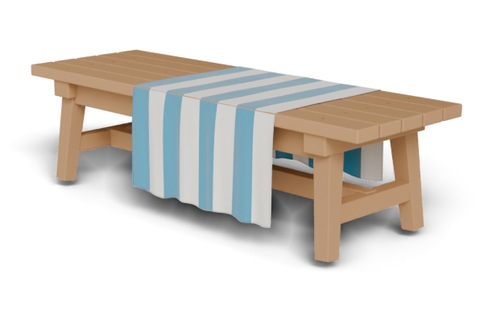} & \includegraphics[width=1.351in]{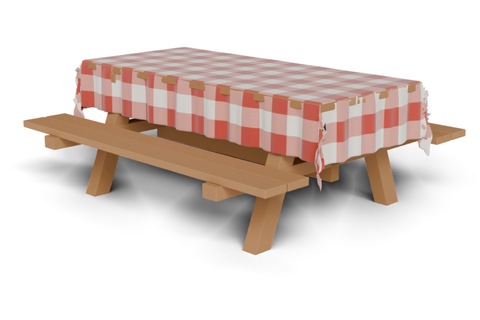} & \includegraphics[width=1.351in]{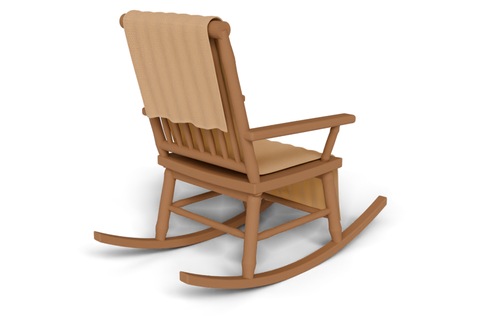} & \includegraphics[width=1.351in]{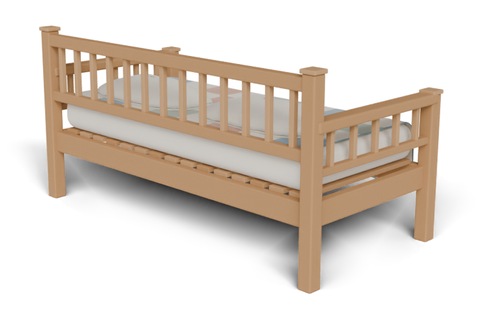} & \includegraphics[width=1.351in]{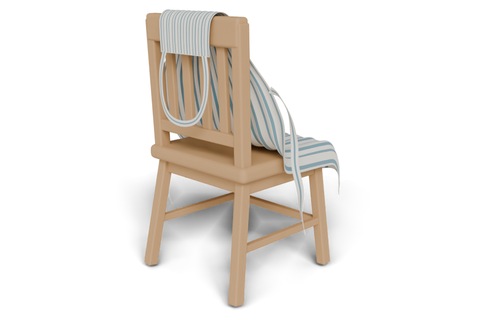} \\[1pt]
\figcap{beach towel, bench}& \figcap{picnic}& \figcap{rocking chair}& \figcap{daybed}& \figcap{apron, chair}
\end{tabular}

  \caption{\textbf{GPT-6 Astra's scenes, rendered.} Ten of the
  thirty-two scenes the model wrote as Blender programs,
  rendered from its exported meshes under our studio camera and
  lighting with the diffuse colours its scripts assign. The fabric is
convincing in a rendered view, with a throw with fringe, a tablecloth with a
hem, and a striped beach towel. Each is a surface posed by the script
rather than equilibrated by a solver. Table~\ref{tab:supp-astra} and
Fig.~\ref{fig:supp-slices} measure where those surfaces pass through
the furniture.}
  \label{fig:supp-astra}
\end{figure*}

A single placement has two parameters, which sampling covers (Sec.~\ref{supp:release}). The argument of Sec.~3.6 of the main paper concerns a scene, that is, $N$ bodies whose placements are corrected jointly in one differentiated simulation, with $2N$ parameters coupled through contact. Table~\ref{tab:supp-multi} equilibrates $N$ sheets on one support, each aimed at its own target, with the sheets coupled by a contact term so that one sheet landing on another moves it. Three conditions share one evaluator and a matched forward budget: straight drop, random search over all $2N$ coordinates at twice the forward rollouts of the gradient run, and Adam on the joint gradient. The grouping is by target difficulty, decided by geometry as before. A target that no method reaches in any run at any $N$ is unreachable from every placement we tried, and counting it adds the same failure to all three columns, so such targets are listed and excluded.

We use two supports. On the stool, every method lands every sheet, and the measure is the final error. At $N{=}4$, search finds nothing better than the straight drop in $80$ samples of eight numbers, while the gradient brings all four sheets under a centimetre. On the rocking chair, whose slatted seat the straight drop misses entirely, the measure is whether a sheet lands at all. The gradient lands every sheet at $N{=}1$, $2$ and $4$, search lands at most one of them, and straight drop lands none.

The gradient serves two tasks here: reaching a landing basin that a straight drop misses and refining within such a basin, and these two tasks require different step sizes. At $\mathrm{lr}{=}0.04$ the gradient lands every rocking-chair sheet at $N{=}1$, $2$ and $4$ (first block of Table~\ref{tab:supp-multi}); at $\mathrm{lr}{=}0.01$ with twice the iterations it brings all four stool sheets under a centimetre, $1.2$\,cm against $2.3$ for the straight drop and $8.5$ for the search (second block). The last block uses a two-phase schedule: the large step until every sheet is in contact, then a new optimiser with the small step, started from the iterate that landed them so that the momentum built up while landing is not carried into the refinement. On the stool, where every sheet lands at once, the restart occurs at the first iteration and the run is the small-step run: two sheets end within two millimetres of their targets and four within a centimetre on average. On the rocking chair the large step lands every sheet and the restart then changes nothing. The contact landscape is piecewise constant in the placement, so where the gradient improves the loss it does so within a few iterations, and the configurations it reaches are not found by sampling.

\paragraph{Grid-search baseline.} The placement grid of the single-body pipeline, applied one body at a time, serves as a further baseline. Body $j$ tries the four placements $\theta_j+f\,(\bm{a}_j-\theta_j)$, $f\in\{\tfrac14,\tfrac12,\tfrac34,1\}$, along the line toward its observed miss while the others hold their current best, the joint rollout is scored by the same rule as the random search, and the sweep repeats until the forward budget of the random search, twice the adjoint's, is spent. Table~\ref{tab:supp-multi-grid} reports it on the obstructed layout. The grid lands exactly the bodies that the naive drop lands, since every point on its line falls on the same side of the obstruction. However, the random search at the same budget lands a few more, and the adjoint lands about twice as many as the grid at every $N$, the bodies that must pass the obstruction to reach their target. The gradient carries a body past the obstruction, whereas the grid and the random search, confined to a line and to the neighbourhood of the straight drop, do not reach it. On eight supports with four seeds each, $32$ runs per $N$, where the four added supports obstruct fewer targets, the adjoint still lands the most bodies at every $N$, $72$ to $75\%$ against $56$ to $62\%$ for the grid and the naive drop and $66$ to $69\%$ for the random search (Table~\ref{tab:supp-multi-grid}, lower block).

\paragraph{Whole-scene success rate.} The obstructed rows of Table~\ref{tab:supp-multi-grid} are a re-run of the four scenes with a per-body record kept, with two seeds each, and with no target excluded. Whole-scene success, where every requested body landed, is $4/8$ runs for the adjoint at each of $N{=}1,2,4$, against $2/8$ for the naive drop and the grid, and at most $3/8$ for the random search. Over the $32$ runs per $N$ of the eight-support block, every requested body lands in $24$, $24$ and $20$ runs for the adjoint at $N{=}1,2,4$, against $20$, $20$ and $16$ for the grid and the naive drop and $22$, $20$ and $16$ for the random search. The bodies that only the adjoint lands are those that must pass the obstruction to reach their target, and they come to rest on its far side, where the nearest stable rest lies.

\begin{table}[tb]
\centering\scriptsize
\setlength{\tabcolsep}{3pt}
\begin{tabular}{l r c cccc c cccc}
\toprule
& & & \multicolumn{4}{c}{bodies landed (\%)} & & \multicolumn{4}{c}{median error (cm)} \\
\cmidrule(lr){4-7}\cmidrule(lr){9-12}
targets & $N$ & runs & naive & search & grid & ours & & naive & search & grid & ours \\
\midrule
obstructed & 1 & 8 & 25 & 38 & 25 & \textbf{50} & & 2.2 & 0.6 & 0.9 & 12.6 \\
 & 2 & 8 & 25 & 31 & 25 & \textbf{50} & & 2.0 & 2.1 & 1.0 & 13.0 \\
 & 4 & 8 & 25 & 41 & 25 & \textbf{56} & & 2.5 & 2.8 & 1.4 & 21.5 \\
\midrule
expanded & 1 & 32 & 62 & 69 & 62 & \textbf{75} & & 0.1 & 0.1 & 0.0 & 0.1 \\
 & 2 & 32 & 62 & 66 & 62 & \textbf{75} & & 0.3 & 0.5 & 0.2 & 0.4 \\
 & 4 & 32 & 56 & 66 & 56 & \textbf{72} & & 0.5 & 0.9 & 0.2 & 1.2 \\
\bottomrule
\end{tabular}

\caption{\textbf{Coordinate grid search at the budget of the random search.} The four obstructed scenes of Table~\ref{tab:supp-multi} at learning rate $0.04$, re-run with a per-body record, two seeds each, every requested body counted and no target excluded. Lower block, expanded: the same four supports and four further ones, an armchair, a bench, a bed corner and a dining table, four seeds each. Landed over all bodies; median centroid error over the bodies each method landed, which is why a method that lands more bodies can show a larger error.}
\label{tab:supp-multi-grid}
\end{table}

\subsection{Holodeck}
\begin{table*}[!t]
  \centering\scriptsize
  \setlength{\tabcolsep}{2pt}
  \begin{minipage}[t]{0.42\textwidth}\centering\begin{tabular*}{\linewidth}[t]{@{\extracolsep{\fill}}l>{\raggedright\arraybackslash}p{0.47\linewidth}@{}}
\toprule
prompt & Holodeck's outcome \\
\midrule
freestanding bathtub / towel & named in the plan, no mesh (native asset) \\
bathtub / towel & named in the plan, no mesh (native asset) \\
bed / blanket & absent from the scene \\
wooden bench / towel & rigid retrieved mesh, on nothing it named \\
sofa / blanket & absent from the scene \\
leather couch / pillow & rigid retrieved mesh, on the requested support \\
wooden daybed / quilt & absent from the scene \\
dining table / tablecloth & absent from the scene \\
picnic table / tablecloth & absent from the scene \\
wooden armchair / plump pillow & rigid retrieved mesh, on sofa-0 \\
wooden armchair / pillow & absent from the scene \\
ironing board / skirt & rigid retrieved mesh, on nothing it named \\
bamboo steamer / dumpling & rigid retrieved mesh, on kitchen counter-0 \\
cardboard box / cloth & absent from the scene \\
wooden table / plate & absent from the scene \\
leather sofa / blanket & absent from the scene \\
stool / cloth & absent from the scene \\
bar stool / towel & rigid retrieved mesh, on nothing it named \\
wooden chair / towel & named in the plan, no mesh (native asset) \\
wooden bench / towel & absent from the scene \\
park bench / diamond kite & absent from the scene \\
armchair / blanket & absent from the scene \\
rocking chair / blanket & absent from the scene \\
kitchen chair / apron & absent from the scene \\
coffee table / napkin & absent from the scene \\
wooden bench / cushion & absent from the scene \\
wooden table / cloth & absent from the scene \\
ladder / towel & rigid retrieved mesh, on towel rack-0 \\
bare wooden armchair / plump pillow & absent from the scene \\
wooden stool / plush toy & absent from the scene \\
table / tissue box & rigid retrieved mesh, on the requested support \\
trash bin / crumpled paper & absent from the scene \\
\bottomrule
\end{tabular*}
\end{minipage}\hfill
  \begin{minipage}[t]{0.57\textwidth}\centering\begin{tabular*}{\linewidth}[t]{@{\extracolsep{\fill}}l ccccc@{}}
\toprule
prompt & bodies & cross/1k & pen. (mm) & contact (\%) & floating \\
\midrule
freestanding bathtub / towel & 1 & 0.0 & 0.0 & 0.7 & 0 \\
bathtub / towel & 1 & 4.0 & 0.6 & 17.6 & 0 \\
bed / blanket & 4 & 11.4 & 0.0 & 23.5 & 0 \\
wooden bench / towel & 1 & 0.0 & 0.0 & 59.0 & 0 \\
sofa / blanket & 13 & 67.3 & 21.1 & 21.8 & 1 \\
leather couch / pillow & 9 & 28.0 & 26.2 & 7.0 & 0 \\
wooden daybed / quilt & 8 & 121.1 & 1.9 & 70.8 & 0 \\
dining table / tablecloth & 1 & 0.0 & 0.0 & 44.6 & 0 \\
picnic table / tablecloth & 1 & 19.8 & 0.0 & 38.1 & 0 \\
wooden armchair / plump pillow & 2 & 79.0 & 1.1 & 63.5 & 0 \\
wooden armchair / pillow & 1 & 9.8 & 0.0 & 33.6 & 0 \\
ironing board / skirt & 7 & 91.4 & 0.5 & 81.9 & 0 \\
bamboo steamer / dumpling & 8 & 13.1 & 0.1 & 32.5 & 0 \\
cardboard box / cloth & \multicolumn{5}{l}{no scene produced} \\
wooden table / plate & 1 & 0.0 & 0.0 & 94.2 & 0 \\
leather sofa / blanket & 1 & 0.0 & 0.0 & 0.1 & 0 \\
stool / cloth & 1 & 0.0 & 0.0 & 1.5 & 0 \\
bar stool / towel & 1 & 0.0 & 0.0 & 46.0 & 0 \\
wooden chair / towel & 2 & 0.9 & 0.0 & 36.5 & 0 \\
wooden bench / towel & 1 & 0.0 & 0.0 & 32.5 & 0 \\
park bench / diamond kite & 20 & 124.5 & 2.0 & 40.9 & 0 \\
armchair / blanket & 6 & 71.2 & 14.4 & 6.9 & 0 \\
rocking chair / blanket & 1 & 0.0 & 0.0 & 0.0 & 0 \\
kitchen chair / apron & 4 & 4.4 & 0.0 & 4.5 & 0 \\
coffee table / napkin & 1 & 0.0 & 0.0 & 100.0 & 0 \\
wooden bench / cushion & 1 & 15.0 & 0.0 & 14.1 & 0 \\
wooden table / cloth & 1 & 0.0 & 0.0 & 0.2 & 0 \\
ladder / towel & 1 & 0.0 & 0.0 & 17.6 & 0 \\
bare wooden armchair / plump pillow & 1 & 0.0 & 0.0 & 31.2 & 0 \\
wooden stool / plush toy & \multicolumn{5}{l}{no scene produced} \\
table / tissue box & 1 & 0.0 & 0.0 & 19.8 & 0 \\
trash bin / crumpled paper & \multicolumn{5}{l}{no scene produced} \\
\bottomrule
\end{tabular*}
\end{minipage}
  \caption{\textbf{Holodeck and GPT-6 Astra, per prompt.} The two systems that build their own scenes.
  Left, Holodeck on all thirty-two prompts: where it placed a mesh,
  the surface named is its own label for the object it chose. Right,
  GPT-6 Astra per prompt, scored by the same rule as every
  other system in Table~\ref{tab:supp-compare-summary}. Astra tags its
  own upholstery as soft, so the cushions, arms and welts it authors for a
sofa are treated here as furniture and scored as colliders rather than
as failed cloth. What remains is the cloth that the prompt asked for.
Its count is still not comparable across systems, because Astra writes
a garment as a garment plus its separate hems and fringe, which is why
the rates beside it pool vertices rather than averaging over bodies.}
  \label{tab:supp-sota}\label{tab:supp-astra}
\end{table*}
\begin{figure*}[!t]
  \centering
  \includegraphics{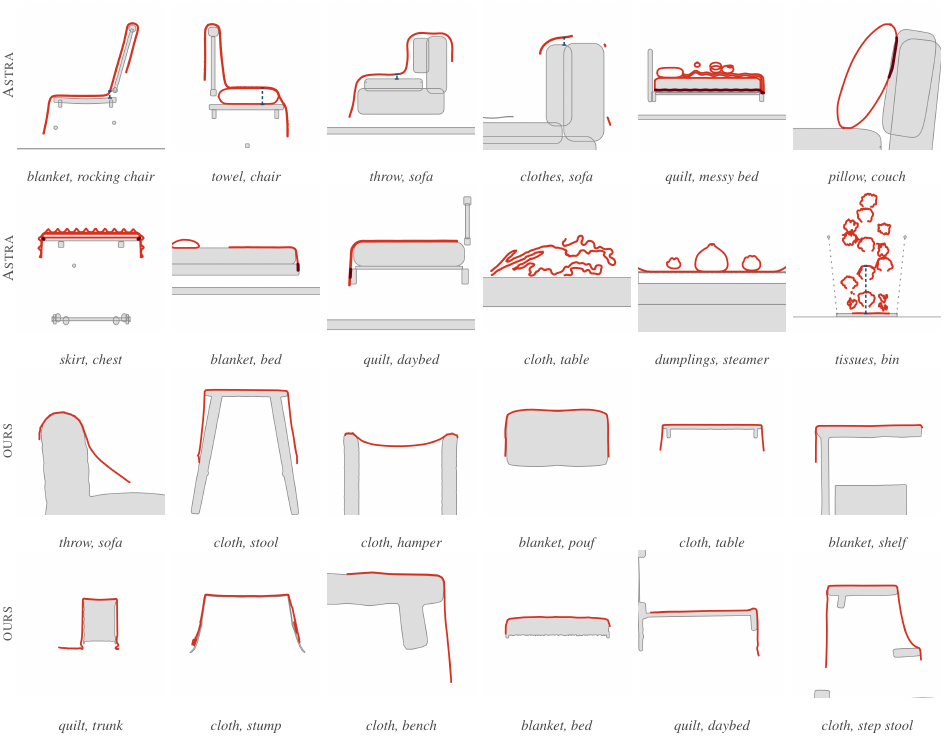}
  \caption{\textbf{Vertical sections through the scene.} Cloth in red,
  the furniture it should rest on filled in grey, cloth inside the
  furniture in dark red, and a dashed line where the cloth hangs above
  its support without touching it. A render cannot show penetration,
  because the surface that is inside another body is hidden by the body
  it is inside, whereas a section shows it directly. Top two rows:
  twelve scenes of GPT-6 Astra. The blanket, the towel, the throw and
  the clothes hang stiffly above their seats; the quilts and the pillow
  run through the mattress, the daybed and the backrest; the skirt
  ripples above the chest, the cloth on the table is a tangle of folds,
  the dumplings sit on a rippled liner, and the tissues hover inside the
  bin. Bottom two rows: twelve of our scenes, where the cloth meets the
  support and stops, wraps its edges and hangs from them under gravity.}
  \label{fig:supp-slices}
\end{figure*}

We run all thirty-two prompts through Holodeck (Yang et al., CVPR 2024), the open-source text-to-scene system closest to this setting. It requires no training. A language model names the objects, retrieval pulls meshes from Objaverse, and a solver places them. Its column in Table~\ref{tab:supp-compare-summary} is nearly empty, because it rarely produces the object that the prompt asks for. Holodeck generates a \emph{room}. Given ``a striped beach towel on a wooden bench'', it builds a striped-beach-towel room, complete with loungers, palms and paintings, but no towel. Its language model's floor-plan parser fails on three prompts at the first attempt and succeeds on a retry. The requested deformable object is absent from \hdAbsent{} of the thirty-two scenes, named in the plan but backed by no mesh in \hdNoMesh{}, and present as a rigid retrieved mesh in \hdPlaced{}, of which \hdNamed{} rest on the support that the prompt names (Table~\ref{tab:supp-sota}).

Retrieval cannot supply a draped object, because no rest shape in the database matches the support over which it would have to drape. This is the reason that our stage sews a rest shape and equilibrates it instead of retrieving one. Holodeck is designed to furnish rooms, and retrieval-and-placement has no representation for fabric that takes the shape of the object on which it lies.

\subsection{GPT-6 Astra}
\label{supp:astra}

A language model that can write and execute code authors the geometry directly, without a scene pipeline. We provide GPT-6 Astra with all thirty-two prompts and allow it to write Blender scripts, each of which exports its objects as meshes together with a manifest recording which ones it considers soft. This baseline is stronger than retrieval. \astraScored{} of the thirty-two scripts produce a scene with a body that the model itself tags as soft, and the other \astraNone{} write no geometry or tag nothing as soft. The results also appear convincing in a rendered view. The scarf over the sofa arm has fringe, the comforter has folds, and the jeans have a fly and pocket curves.

Few of these scenes are simulated. Three of the thirty-two scripts add Blender's cloth modifier, mark the furniture as a collision object and step the frames. One of those then wrote no geometry at all, and the other two are scored in Table~\ref{tab:supp-astra} like the rest. The remaining twenty-nine have no cloth modifier, no collision object, no bake and no frame stepping. The folds are analytic noise displacing a sewn surface, and one script states the design in a comment, describing the comforter as a closed thick sewn surface. The consequence is visible in cross-section (Fig.~\ref{fig:supp-slices}), where the blanket's outline runs inside the armchair it is draped over. A sculpted fold is shaped to resemble contact, and no step of the process checks for penetration. However, Table~\ref{tab:supp-compare-summary} puts the rate beside ours.

Two further artifacts arise from the same cause. First, the geometry is authored part by part. A single scarf arrives as a scarf plus thirty-four separate fringe strands, and the number of bodies in a scene reflects how the author subdivides it rather than the scene itself. We therefore pool every aggregate we report over vertices rather than average over bodies. Second, because no constraint ties the parts to one another or to the ground, the errors are not confined to cloth. In two of the first eight scenes the furniture hovers over its own floor plane, by $12$\,mm under the armchair and $31$\,mm under the bench. We therefore distinguish a body that touches nothing at all from one that touches only bodies that are themselves unsupported, and report both.

In summary, a language model produces cloth-shaped geometry with more decorative detail than our stage generates. However, sculpted geometry is not in equilibrium with what it rests on. Additional detail does not change this, because no step of the process tests whether the fabric and the furniture can occupy their positions at the same time.

\section{Scene Prompts}
\label{supp:prompts}

In this section, we list the prompt form and representative prompts. All scenes are generated end-to-end from single-sentence prompts of the form ``A \emph{[soft object]} on/over a \emph{[support]}, plain background,'' with detector-friendly simple nouns for the requested objects. Because the catalog is built from segmentation, ``side table'' silently deletes the object from the catalog, whereas ``wooden table'' detects at $0.85$. Representative prompts are: \emph{a towel
draped over a wooden chair; a tablecloth on a round side table; a
knitted blanket draped over the back of an armchair; a patterned
throw blanket draped over a leather sofa; clothes scattered on a
sofa; a laundry basket with clothes inside; a trash bin with crumpled
paper; a steamer pot with dumplings; a messy bed with a comforter,
pillows, and plush toys; a striped beach towel on a wooden bench; a
tissue box with a tissue on a table; a towel hanging on a towel
rack.}

\end{document}